\documentclass[11pt]{article}
\usepackage[T1]{fontenc}
\usepackage{lmodern}
\usepackage[margin=1in]{geometry}
\usepackage{amsmath,amssymb,amsthm,mathtools,bm}
\usepackage{microtype}
\usepackage{graphicx,booktabs,mathrsfs,placeins}
\usepackage{nicefrac}

\usepackage{needspace,aliascnt}
\usepackage{titletoc}
\usepackage[round,authoryear]{natbib}
\usepackage[colorlinks=true,citecolor=blue,linkcolor=blue,urlcolor=blue]{hyperref}
\usepackage[nameinlink,noabbrev]{cleveref}
\usepackage{color-edits}

\usepackage{tikz}
\usetikzlibrary{positioning, arrows.meta}

\crefname{section}{Section}{Sections}
\crefname{subsection}{Section}{Sections}
\crefname{theorem}{Theorem}{Theorems}
\crefname{proposition}{Proposition}{Propositions}
\crefname{corollary}{Corollary}{Corollaries}
\crefname{lemma}{Lemma}{Lemmas}
\crefname{definition}{Definition}{Definitions}
\crefname{assumption}{Assumption}{Assumptions}
\crefname{remark}{Remark}{Remarks}
\crefname{figure}{Figure}{Figures}
\crefname{table}{Table}{Tables}
\crefname{appendix}{Appendix}{Appendices}
\Crefname{appendix}{Appendix}{Appendices}
\newtheorem{theorem}{Theorem}[section]
\newaliascnt{proposition}{theorem}
\newtheorem{proposition}[proposition]{Proposition}
\aliascntresetthe{proposition}
\newaliascnt{corollary}{theorem}
\newtheorem{corollary}[corollary]{Corollary}
\aliascntresetthe{corollary}
\newaliascnt{lemma}{theorem}
\newtheorem{lemma}[lemma]{Lemma}
\aliascntresetthe{lemma}
\theoremstyle{plain}
\newaliascnt{definition}{theorem}
\newtheorem{definition}[definition]{Definition}
\aliascntresetthe{definition}
\newaliascnt{assumption}{theorem}
\newtheorem{assumption}[assumption]{Assumption}
\aliascntresetthe{assumption}
\theoremstyle{plain}
\newaliascnt{remark}{theorem}
\newtheorem{remark}[remark]{Remark}
\aliascntresetthe{remark}
\newcommand{\E}{\mathbb E}
\newcommand{\Pp}{\mathbb P}
\newcommand{\R}{\mathbb R}
\newcommand{\V}{\mathcal V}
\newcommand{\Cc}{\mathcal C}
\newcommand{\F}{\mathcal F}
\newcommand{\Hc}{\mathcal H}
\newcommand{\Dc}{\mathcal D}
\newcommand{\Mc}{\mathcal M}

\newcommand{\Pn}{\mathbb E_n}
\newcommand{\Vbb}{\mathbb V}
\newcommand{\ip}[2]{\left\langle #1,#2\right\rangle}
\newcommand{\norm}[1]{\left\lVert #1\right\rVert}
\newcommand{\Var}{\operatorname{Var}}
\newcommand{\sgn}{\operatorname{sgn}}
\newcommand{\cl}{\operatorname{cl}}

\DeclareMathOperator*{\argmin}{arg\,min}

\usepackage{bookmark}
\makeatletter
\renewcommand{\@pnumwidth}{2.5em}
\makeatother
\contentsmargin{3em}
\titlecontents{part}[0em]{\addvspace{1.2em}\large\bfseries}
  {}{}{\hfill\contentspage}
\titlecontents{section}[2.6em]{\addvspace{0.7em}\bfseries}
  {\contentslabel{2.6em}}{\hspace*{-2.6em}}{\hfill\contentspage}
\titlecontents{subsection}[6.1em]{}
  {\contentslabel{3.5em}}{}{\titlerule*[0.75pc]{.}\contentspage}
\newcommand{\appendixdivider}[2]{%
  \clearpage
  \vspace*{0.15\textheight}%
  \part*{#1}%
  \phantomsection
  \bookmarksetup{startatroot}%
  \addcontentsline{toc}{part}{#1}%
  #2\par
  \vfill
  \clearpage
}

\title{Policy Improvement with Terrible but Valid Instruments}
\author{Shiangyi Lin\\
ICME\\
Stanford University \and Vasilis Syrgkanis\\
Management Science and Engineering\\
Stanford University}
\date{October 2026}
\begin{document}
\maketitle
\begin{abstract}

We study first-order local policy improvement with an endogenous continuous
treatment and access to an instrument, when the instrumental-variable moment restrictions only partially identify the structural response. Both treatment and instrument are allowed to be multidimensional. We show that among all responses consistent with the
moments, the gradient of the one with minimum gradient $L^2$ norm (gradient energy) always yields a (weak) policy-improving direction, whereas the gradient of the typically used minimum function $L^2$ norm solution can be harmful. The minimum gradient energy, which we call the identified gradient energy, is the identifiable local improvement that the instrument can support. We provide a finite sample analog of this selection procedure in the form of a moment-constrained, minimum empirical gradient energy estimator. We show that the estimator satisfies
a lower bound on policy improvement under weak assumptions (e.g., no source condition or well-specified hypothesis class). Under further conditions, we also show that the improvement lower bound approaches the identified gradient energy as the sample size increases and the approximation and optimization errors vanish. We specialize the theory to (sparse) high-dimensional linear function spaces. Our estimator depends on a hyperparameter that requires an upper bound on the degree of misspecification and on the Rademacher complexity of the instrument test class. We also show how to select an improving direction without the need for any hyperparameter tuning, based on a simple procedure that essentially only requires estimating the covariance of the most outcome-informative instrument with a flexible set of features of the treatment. 
\end{abstract}

\clearpage
\begingroup
\small
\tableofcontents
\endgroup
\clearpage

\section{Introduction}

Many policy decisions involve changing a continuously valued treatment
using observational data in which the treatment is endogenous.
We study incremental policies that infinitesimally nudge each unit's
naturally assigned treatment, with the direction and size of the nudge
depending on the baseline treatment value. We assume that we have access to an exogenous instrument which influences the treatment assignment, but is not rich enough to identify the structural treatment response. To construct a local improvement policy, we must then decide
which of these response functions to use. Selecting the one with the smallest
norm is a common way to resolve this ambiguity. \emph{Is it safe to perturb the treatment in the direction of the gradient of the selected function, and if not, does there exist any selection criterion that guarantees local policy improvement?}

We consider the additive structural model
$Y(x)=g(x)+U$, with observed outcome $Y=g(X)+U$ and
$\E[U\mid Z]=0$, where $X\in \R^d$ is the endogenous treatment, $Z\in \R^{d_Z}$ is the instrument and $U$ is the unobserved confounder. The observed treatment $X$ is generated by a
baseline policy and can depend on the unobserved confounder $U$.
Our goal is to construct a direction $a$ such that changing treatment
to $X+\epsilon a(X)$, for some small scalar $\epsilon>0$, improves the baseline policy to first order.
Whether this direction leads to local improvement for sufficiently small $\epsilon$ is determined by the directional derivative of the value function in the direction of the perturbation. Under standard regularity conditions, the latter can be expressed as:
\[
 I(a):=\left.\frac{\partial}{\partial\epsilon}
 \E[Y(X+\epsilon a(X))]\right|_{\epsilon=0}
 =\E[\nabla g(X)^{\mathsf T}a(X)]
\] 
The data restrict $g$ through the NPIV equation
$\E[Y-g(X)\mid Z]=0$. However, in many cases, the instrument is too weak to guarantee that the NPIV equation has a unique solution. For example, a finite-valued instrument provides only finitely many conditional mean restrictions, which can leave an infinite-dimensional response class with a continuous endogenous treatment variable severely underidentified. 

If $g$ was the unique solution to the NPIV equation, then the obvious policy improving direction is $\nabla g$, which would yield a first order improvement of $I(\nabla g)=\E[\|\nabla g(X)\|^2]$, i.e., the local improvement is the \emph{gradient energy} of the response function. However, when the NPIV equation has multiple solutions, then there could exist feasible solutions $h$, such that if we act as-if they were the true response function and take a step in the direction of their gradient $\nabla h$, then we might end up with a negative first order effect $I(\nabla h)=\E[\nabla g(X)^\top \nabla h(X)]$. \emph{Our main goal is the development of selection procedures that lead to the selection of a point $h_\star$ in the feasible set, that guarantees first order policy improvement, i.e., $I(\nabla h_{\star})\geq 0$.} At the outset, it is not even clear that such a selection procedure even exists; it is plausible that for any selection procedure, there exists a true response function $g$ for which the first order effect is negative. For instance, in the face of multiple solutions, several prior works (e.g., \cite{darolles2011,bennett2023}) have focused on the minimum function norm selection criterion, i.e., choosing the feasible solution $h$ that has the minimum $L^2(P_X)$ norm $\|h\|_{L^2(P_X)}^2\triangleq \E[h(X)^2]$. A natural question is whether this selection procedure has the non-negative improvement property.

\paragraph{Selection criterion for safe improvement.} 
Our first observation is that (perhaps not surprisingly), the minimum function norm selection procedure can lead to negative improvement. Thus, the most widespread way of resolving the under-identification ambiguity can lead to harmful policy perturbations. This holds even when restricting to smooth structural response functions and informative instruments.

However (and perhaps surprisingly), we show that there exists a universally safe selector that always leads to perturbations that guarantee non-negative first-order improvement! In particular, we show that in the population limit, 
selecting the feasible response with minimum gradient norm 
\[
{\cal E}(h) := \|\nabla h\|_{L^2(P_X)}^2=\E[\|\nabla h(X)\|^2]
\]
guarantees nonnegative policy improvement. Let $\V$ be the function space over which the conditional moment equation is solved, assume that $\V$ is a convex set, and define
\[
 \kappa:=\inf_{h\in\V:\,\E[Y-h(X)\mid Z]=0}
                    \E[\|\nabla h(X)\|^2].
\] 
Whenever this minimum is attained at $h_\star$, we show that, irrespective of which feasible solution is the true response $g$, using the gradient of $h_\star$ as the policy perturbation direction gives first order improvement guarantee
\[
I(\nabla h_\star)\geq \kappa\ge0.
\]
The \emph{identified gradient energy} $\kappa$ measures the degree of local improvement that is \emph{safely} supported by our instrument and could be substantially smaller than the gradient energy $\E\|\nabla g(X)\|^2$ of the true response $g$, especially when the instrument is very weak. In that sense, the perturbation $\nabla h_\star$ can be viewed as a \emph{``best safe effort'' solution, given the cards we have been dealt.} We formally show that the improvement guaranteed by this direction is the optimal worst-case improvement guarantee among all possible perturbation directions (after unit-norm normalization), even ones that do not correspond to gradients of a selected response function. 
In the extreme case of a very weak instrument, it is plausible that the identified gradient energy is $0$, in which case the only safe improvement strategy is to maintain the status quo; the instruments cannot credibly support any local perturbation. However, we expect that in most realistic settings, a relevant instrument (with large reduced form predictive power, i.e., large $\Var(\E[Y\mid Z])$), no matter how terrible it is, will result in a strictly positive identified gradient energy $\kappa$.

Attainment of the infimum by a function $h_\star$ holds under benign regularity conditions, e.g., when $\V$ is a bounded and closed set (under the graph norm $\|h\|_{L^2(P_X)}+\|\nabla h\|_{L^2(P_X)}$), or when $\V$ corresponds to the space of functions with square integrable weak derivatives  on a bounded connected Lipschitz domain and the treatment density is bounded away from zero and infinity. More broadly, it is attained (and can be shown to be unique) under a condition known as the Poincar\'e inequality, which bounds the variance of functions in $\V$ by their gradient energy, i.e., for some constant $C_P>0$ 
\begin{equation}
 \Var(h(X))\le C_P\mathcal E(h)\qquad(h\in\V),\qquad 0<C_P<\infty.
\end{equation}
Hence, attainment should be thought of as the typical case. Moreover, under this same Poincar\'e inequality note that $\kappa 
\geq\Var(\E[Y\mid Z])/C_P>0$,\footnote{For any feasible response $h$, by the law of total variance: $\Var(h(X))\geq \Var(\E[h(X)\mid Z])=\Var(\E[Y\mid Z])$.} giving a quantitive form of the claim in the previous paragraph.
Even if the infimum is not attained, the non-negative first order improvement property holds for the limiting minimum-gradient-energy direction. The closure of the feasible gradient set
has a unique minimum-norm element $a_\star$. Every feasible minimizing
sequence converges to this direction in gradient $L^2$ and
$I(a_\star)\geq\|a_\star\|_2^2=\kappa$. 

The intuition behind why this selection procedure leads to guaranteed non-negative improvement is relatively simple. Consider the case when the infimum is achieved at a function $h_\star$. Since $\V$ is a convex set and the moment restrictions are linear in $h$, note that the feasible set is also a convex set, if $h, h'\in \V$ satisfy the moment restrictions, then any convex combination of $h, h'$ also satisfies the moment restrictions. Moreover, since $g \in \V$ any convex combination of $h_\star$ and $g$ is feasible. The first order condition of the optimization problem implies that perturbing the solution in the direction $g-h_\star$ can only increase the gradient energy. This first order condition translates to:
\begin{align*}
    0 \leq \frac{\partial}{\partial t} {\cal E}(h_\star + t (g-h_\star))\bigg|_{t=0^+} = 2\,\E[\nabla h_{\star}(X)^\top\, (\nabla g(X) - \nabla h_\star(X))]
\end{align*}
By re-arranging the first order condition we get the first order improvement guarantee:
\begin{align*}
    I(\nabla h_*) = \E[\nabla g(X)^\top \nabla h_\star(X)] \geq \E[\|\nabla h_\star(X)\|^2] = \kappa \geq 0
\end{align*}
Note that this argument remains valid whenever the feasibility set that corresponds to the partially identified structural response function is a convex set and is not specific to the NPIV moment constraints. Hence, the high-level methodology we develop here is potentially applicable beyond NPIV (partial) identification arguments. We defer such extensions to future work and fully develop the NPIV setting in this work.

\paragraph{Finite-sample policy improvement.}
The population selection criterion, naturally gives rise to a finite sample procedure. We estimate a response function by minimizing an empirical gradient-norm
criterion within a convex set of hypotheses ${\cal H}$ satisfying an empirical
analog of the moment constraints. Given that the moment constraints are conditional, we first turn them into an unconditional set of moment constraints by replacing them with $\sup_{f\in \F} \E[(Y - h(X))\, f(Z)]$ for a potentially
infinite class of instrument \emph{test} functions $\F$ and then consider their empirical analog. This is analogous to the test-function
formulation in the context of adversarial conditional moment estimation \citep{dikkala2020}. Even though the population moments need to be zero at the true response, the empirical moments can be violated due to finite sample variance (captured by the Rademacher complexity $r_{\F}$ of the test function class $\F$) and mis-specification error (denoted $\epsilon$) of the hypothesis class $\Hc$. Thus, we need to relax the empirical moment constraints to allow for a small violation $\tau_n=O(r_\F + \epsilon)$. This leads to the minimum empirical gradient energy, empirical moment constrained estimator,
\begin{align*}
    \min_{h\in \Hc} \E_n[\|\nabla h(X)\|^2], \qquad \text{ s.t. } \sup_{f\in \F} \E_n[(Y - h(X))\, f(Z)] \leq \tau_n
\end{align*}
This estimator can be viewed as a form of a Dantzig selector and its generalization, the Regularized Minimum Distance estimator \citep{belloni2018high}, with a gradient norm objective. The moment-constrained estimator is one way to turn the selection
principle into a learning procedure. We also discuss two alternatives that incorporate the gradient penalty directly into the estimation objective, adapting two widely used paradigms in the literature, namely non-parametric two-stage least squares (e.g., DeepIV \cite{hartford2017}) and adversarial estimation (e.g., minimax IV \cite{dikkala2020}).

Our analysis
applies to general bounded hypothesis and test classes, and measures
their statistical complexity through localized Rademacher complexities
and the associated critical radii. For an approximately optimal
solution $\widehat h$, with optimization error $\varepsilon_{\rm opt}$, we establish the lower bound (see \cref{thm:safety})
\begin{equation}\label{eqn:intro-guarantee-1}
 I(\nabla \widehat h)\ge
 \tfrac12\E[\|\nabla\widehat h(X)\|^2]-C(\rho_{\Dc}^2+\epsilon^2)
                  -\varepsilon_{\rm opt},
\end{equation}
with high probability. Here, $\rho_{\Dc}$ is the confidence-adjusted
critical radius of the gradient class $\{\nabla h: h \in \Hc\}$. $\epsilon$ bounds the mis-specification error of the hypothesis class $\Hc$, i.e.,
$\inf_{h\in\Hc}\norm{h-g}_{\V}$, where $\norm{f}_{\V}^2=\norm{f}_2^2+\norm{\nabla f}_2^2$ is the graph norm of the gradient operator. This norm controls approximation error of both function values and
derivatives.
We also prove an analogous bound that uses the
empirical gradient norm 
\begin{equation}\label{eqn:intro-guarantee-2}
 I(\nabla \widehat h)\ge
 \tfrac12\E_n[\|\nabla\widehat h(X)\|^2]-C(\rho_{\Dc}^2+\epsilon^2)
                  -\varepsilon_{\rm opt},
\end{equation}
Hence, the empirical gradient energy of the estimated solution provides an observable signal for the magnitude of the positive population improvement.

\paragraph{Approaching the identified gradient energy.}  The finite sample improvement guarantee presented above in Equation~\eqref{eqn:intro-guarantee-1} (or Equation~\eqref{eqn:intro-guarantee-2}) is robust in many respects: it does not require a bound on the ill-posedness of the inverse problem (e.g., a source condition), nor any richness of the test function space $\F$, nor that our estimate converges to the response $h_\star$ with respect to $L^2$ distance. Without any such requirement, we are guaranteed to converge to a solution with non-negative improvement as the sample size increases (assuming we choose richer classes for larger sample sizes, such that the approximation error also vanishes). 

However, this result only provides a policy improvement lower bound that depends on the empirical (or population) gradient energy of the \emph{estimated} solution. If the latter is large (as indicated by the observed empirical gradient energy), then this provides a certificate of substantial local improvement. An immediate question is whether we can further lower bound the first-order improvement of the estimated solution by the identified gradient energy $\kappa$. Without further assumptions, we prove that the leading term $\E[\|\nabla \hat{h}(X)\|^2]$, in the lower bound, can be replaced by the population quantity:
\begin{align*}
\kappa_{\Hc, \F}(t) = \inf\{{\cal E}(h): h\in \Hc,\, \sup_{f\in \F} \E[(Y - h(X))\, f(Z)] \leq t\} 
\end{align*}
where $t$ is set to a slightly larger value $t_n$ than the tolerance $\tau_n$ used in the estimator, i.e., $t_n=O(\tau_n +r_{\Mc})$, where $r_{\Mc}$ is the Rademacher complexity of the function space:
\begin{align*}
\Mc=\{(Y-h(X))f(Z)/2:h\in\Hc,\ f\in\F\},
\end{align*}
so as to ensure that the empirical solution $\hat{h}$ satisfies the population version of the relaxed moment constraints.

We can further show that if the safe improvement direction $a_\star$ admits the dual representation:
\begin{align*}
\forall h\in \Hc: \E[\langle a_*(X), \nabla h(X)\rangle] = \E[q(Z)\, h(X)]
\end{align*}
and test functions are rich enough such that $q/Q$ lies in the convex hull of $\F$ for some constant $Q$, then: 
\begin{align*}
\kappa_{\Hc, \F}(t) \geq \kappa - O(t + \epsilon)
\end{align*}
Under these assumptions, we prove a local improvement lower bound that approaches $\kappa$ in the large sample limit (see Theorem~\ref{thm:benchmark}):
\begin{align*}
I(\nabla \hat{h}) \geq \kappa - O(r_{\F} + r_{\Dc} + r_{\Mc} + \epsilon + \varepsilon_{\rm opt})
\end{align*}
We also discuss how the condition can be relaxed to allow for an approximate dual representation.

\paragraph{Linear hypotheses.}  When the function space $\Hc$ comprises of linear functions of the form $\theta^\top\phi(X)$ for a known feature dictionary $\phi(X)\in \R^p$, then it suffices to test functions $f$ of the form $\mu(Z)=\E[\phi(X)\mid Z]$. Instead of learning $\mu(Z)$ and testing for it, which requires solving many non-parametric regression problems,
one can simply test for a sufficiently large set of test functions $\psi(Z)\in \R^k$:
\begin{align*}
\min_{\theta\in \Theta} \theta^\top \E_n[J_\phi^\top J_\phi] \theta, \qquad \text{ s.t. } \|\E_n[(Y - \theta^\top\phi(X))\psi(Z)]\|_{\infty} \leq \tau_n 
\end{align*}
where $J_\phi=(\nabla\phi_1(X),\ldots,\nabla\phi_p(X))$ is the Jacobian of the feature map $\phi$. In this setting, the constrained optimization problem reduces to a convex quadratic optimization problem with linear constraints, which can be solved computationally efficiently. The dimension of the test functions is allowed to grow exponentially with the sample size, as the slackness factor $\tau_n$ needs to scale only logarithmically with the number of test functions $k$, i.e., $\tau_n=O(\sqrt{\log(k)/n} + \epsilon)$, where $\epsilon$ is the mis-specification error of the linear hypothesis space used to approximate the response functions. If we impose sparsity restrictions on $\theta$, i.e., $\Theta=\{\theta\in \R^p: \|\theta\|_1\leq R\}$, for some constant $R$, then the constrained estimator yields rates that also allow for the dimension $p$ to grow exponentially. Under weak conditions we prove improvement guarantees of the form: $I(\nabla \hat{h})\geq \E_n[\|\nabla \hat{h}(X)\|^2] - O(R^2\sqrt{\log(p)/n}+R\epsilon)$ (see \cref{cor:linear-improvement}).

The returned solution always maintains the non-negative improvement property in the large sample limit; albeit, if the test functions are not expressive enough, the identified direction can have a much smaller gradient energy than $\kappa$.   Without further assumptions, we can refine the lower bound to $\kappa_{\phi,\psi}- O(\sqrt{\log(p\,k)/n}+\epsilon)$, where $\kappa_{\phi,\psi}$ is the optimal value of the population optimization problem:
\begin{align*}
\min_{\theta\in \R^p} \theta^\top \E[J_\phi^\top J_\phi] \theta, \qquad \text{ s.t. } \|\E[(Y - \theta^\top\phi(X))\psi(Z)]\|_{\infty} = 0
\end{align*}
Under interpretable conditions on the richness of the test functions $\psi$, i.e., that the kernel of the matrix $\E[\psi(Z)\phi(X)^\top]$ is the same as the kernel of the matrix $\E[\mu(Z)\,\phi(X)^\top]$, we also prove lower bounds of the form $\kappa - O(\sqrt{\log(p\,k)/n}+\epsilon)$.

\paragraph{Unknown misspecification error.} The guarantee for the constrained estimator needs a moment tolerance large enough for an approximation of the true response to pass the tests. Hence, it depends on an upper bound on the misspecification error, which is hard to posit in practice. We show that if all we want to achieve is to identify a direction with non-negative improvement, then the choice of the moment tolerance can be avoided and be replaced by a much more benign assumption; namely, that the best-in-hypothesis class function violates the moment conditions less than the everywhere-zero function. We expect that for a sufficiently flexible hypothesis space, such a property to be satisfied. We develop this methodology for a hypothesis class spanned by a fixed feature
dictionary $\phi=(\phi_1,\ldots,\phi_p)^{\mathsf T}$. We also consider a finite set of test functions $\psi = (\psi_1,\ldots, \psi_k)$. Let $b_j=\E[Y\psi_j(Z)]$ be the moment of test $\psi_j$ at the zero function, and let $j_\star$ maximize
$|b_j|$, the most violated test. We take the policy direction
\[
 a(x)=J_\phi(x)\, v,\qquad
 v=\sgn(b_{j_\star})\,G^{-1}\E[\phi(X)\psi_{j_\star}(Z)],\qquad
 G=\E[J_\phi(X)^{\mathsf T}J_\phi(X)],
\]
where $J_\phi=(\nabla\phi_1,\ldots,\nabla\phi_p)$. If the gradient projection of the true response onto the
dictionary fits the tested moments strictly better than the zero
function, this direction strictly improves policy value; its
improvement is at least the amount by which the projection's largest
tested violation falls below that of the zero function. This direction is related to the initial direction of the constrained path as the tolerance first excludes zero (see \cref{sec:initial-direction} for a formal statement). The condition can hold even when the data
do not identify the response or its average derivative. Moreover, under the stated Poincar\'e and approximation assumptions and with a nonzero baseline outcome moment $b_j$, we show that the required condition will eventually hold for a sufficiently large (albeit constant) approximating sieve (in graph norm), as long as the approximation error of the sieve converges to zero. We provide a finite sample analogue of this procedure and show that it provides a direction of improvement with high probability. 

\paragraph{A simple and interpretable nudge} The chosen direction above takes a very intuitive form in the case of identity feature maps, i.e., $\phi(x)=x$ and $\psi(z)=z$, where we also assume that the instruments are centered and standardized. In this case, the gram matrix $G$ and the jacobian $J_\phi(x)$ are the identity, i.e., $G=I$ and $J_\phi(x)=I$. Thus, we are perturbing the baseline value $X$ in a constant direction $a(x)=v$, defined as follows: i) we first identify the instrument coordinate with the largest covariance with $Y$, i.e.,  
\[
j_*=\arg\max_{j} |\mathrm{Cov}(Y, Z_j)|.
\]
and we let $s_{j_*}$ be the sign of the correlation. Essentially, this is the instrument coordinate with the largest ``intention-to-treat'' effect, in absolute terms. ii) we define the perturbation direction to be:
\[
v = s_{j_*} \mathrm{Cov}(X, Z_{j^*})
\]
Essentially: \emph{i) we find the instrument with the largest absolute intent-to-treat effect, ii) we calculate the covariance vector of that instrument with all the endogenous treatments (first-stage effect), iii) we nudge the endogenous treatments in the direction of that covariance, if the intent-to-treat effect is positive, or in the opposite direction, if the intent-to-treat effect is negative.} This principle, gives a very simple procedure to operationalize and our theorems show that it yields non-negative improvement under a benign assumption: \emph{the parameter vector $\theta$ that best approximates the gradient of the response function $g$, i.e., the average derivative vector $\theta=\E[\nabla g(X)]$, violates the moment conditions $\E[(Y-\theta^\top X)Z]=0$, less than the zero parameter vector.} 

In order to maximize the covariance with $Y$, among test functions with equal $\ell_2$ norm, it suffices to test for the reduced form and centered test function $\psi_\star(Z) = \E[Y\mid Z] - \E[Y]$. This test function achieves the highest and non-negative covariance. Thus, as long as the average derivative vector $\theta=\E[\nabla g(X)]$ violates the moment condition $\E[(Y-\theta^\top X)\psi_\star(Z)]=0$, less than the zero parameter vector, then the following constant intervention leads to first-order non-negative improvement:
\begin{align}
    v=\mathrm{Cov}(X, \E[Y\mid Z])
\end{align}
In the case of a non-identity treatment response dictionary $\phi$, the direction is not constant and given by:
\[
 a=J_\phi(x)\, v,\qquad
 v=G^{-1}\mathrm{Cov}(\phi(X),\psi_{\star}(Z)),\qquad
 G=\E[J_\phi(X)^{\mathsf T}J_\phi(X)],
\]

\paragraph{Organization.} After the example and related work, \cref{sec:setup} defines the response model and policy derivative. \Cref{sec:population} gives the population selection and worst-case guarantees. \Cref{sec:statistics} develops empirical improvement without source or richness assumptions, and \cref{sec:sources} gives comparisons with the identified gradient energy. \Cref{sec:sparse-linear} specializes these results to linear dictionaries; \cref{sec:initial-direction} studies direction selection under unknown misspecification.

\section{An Illustrative Example}
\label{sec:illustration}

\begin{figure}[!htbp]
 \centering
 \includegraphics[width=0.8\linewidth]{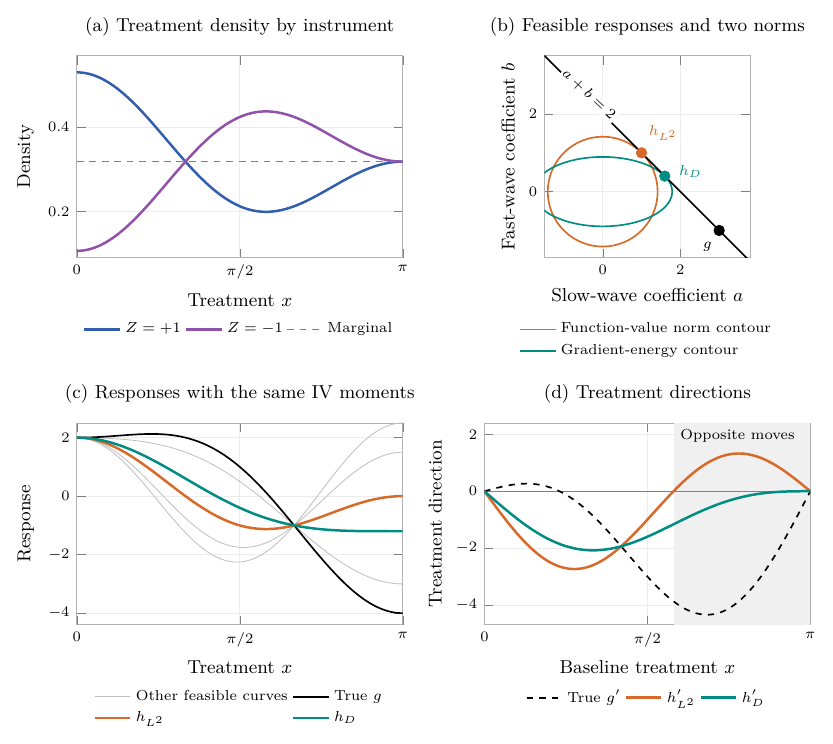}
 \caption{An instrument can be relevant without point identifying the
 response. Panel (a) shows $p(X\mid Z)$ and the uniform marginal density.
 Panel (b) includes the feasible line, $a+b=2$, drawn in black. The orange contour $a^2+b^2=2$ and teal contour $a^2+4b^2=16/5$ are the squared function value norm and gradient energy respectively. 
 Panels (c)--(d) show the feasible response curves and their treatment
 directions. In the shaded range, the two selected directions move
 treatment in opposite ways. The $L^2$-minimizing direction (orange) moves treatment right, where the true response decreases, while the gradient-energy-minimizing direction (teal) moves treatment left, where the true response rises. Hence, the gradient minimizing direction benefits such units, while the $L^2$ minimizing direction harms such units. Note though that in the region close to zero, both directions move treatment left, lowering the outcome because the true response is slightly increasing there.}
 \label{fig:illustration-geometry}
\end{figure}

In this section, we illustrate our main results through the lens of a simple example. Consider a continuously valued scalar endogenous treatment $X\in [0, \pi]$ (e.g., a drug dose). Consider an instrument $Z$ that takes value in $\{-1, 1\}$, equiprobably, (e.g., random assignment into new protocol that recommends a higher dose) and changes the distribution of treatment intensity based on the conditional density (see Figure~\ref{fig:illustration-geometry}(a)): 
\begin{align*}
 p(x\mid Z=z) ~& =\frac{1+z(\cos x+\cos2x)/3}{\pi},
 \qquad 0<x<\pi.
\end{align*}
The actual value of the treatment is determined by an unobserved confounder $U$, drawn uniformly in $[-1/2,1/2]$ independently of $Z$ (e.g., unobserved severity of patient's symptoms) through the structural equation:
\[
 X=F_Z^{-1}(U+1/2),
\]
where $F_z$ is the CDF associated with the density $p(x\mid Z=z)$. In other words, $U+1/2$ is the percentile rank of $X$ given $Z$. Note that the marginal distribution of $X$ is uniform on $(0, \pi)$. The outcome is determined by a structural response function $g(X)$ and the unobserved confounder $U$ through the structural equation:
\[
 Y=g(X)+U,\qquad g(x)=3\cos x-\cos2x.
\]
The outcome increases at low intensities and decreases at high intensities (see Figure~\ref{fig:illustration-geometry}(c)). Our goal is to find a local perturbation of the treatment that improves the outcome on average. The perturbation can depend on the current value of the treatment, e.g. increase the dose for low values of $X$ and decrease it for high values of $X$. 

We consider a simple two-dimensional response class spanned by the slow and fast cosine waves, $\cos x$ and $\cos2x$. The true response $g$ is in this class, but the instrument does not point identify it. The instrument only implies two moment restrictions on the response function $h$:
\[
\E[h(X)\mid Z=1] = \E[Y\mid Z=1] = \frac{1}{3},\qquad
\E[h(X)\mid Z=-1] = \E[Y\mid Z=-1] = -\frac{1}{3}.
\]
For a candidate response $h_{a,b}(x)=a\cos x+b\cos2x$, the
conditional mean $\E[h_{a,b}(X)\mid Z=z]$ is $z(a+b)/6$. Hence the moment restrictions hold exactly when $a+b=2$. Thus the feasible set of responses is given by the set:
\[
\Cc=\{h_{a,b}: a + b = 2\}
\]
The instrument reveals the sum of the two
coefficients, but cannot determine how that sum is split. We show that the minimum-function-value solution and the minimum-gradient-energy solution differ, and that the latter has a positive policy derivative while the former has a negative one.

The upper-right panel shows the feasibility constraint and the contour curves of the function-value norm and gradient energy. For a candidate response $h_{a,b}$ the squared $L^2$ norm and the gradient energy can be computed as:
\begin{align*}
\E[h_{a,b}(X)^2] &= \frac{a^2+b^2}{2}, &
\E[h_{a,b}'(X)^2] &= \frac{a^2+4b^2}{2}.
\end{align*}
The points where the smallest contours touch the constraint line are the minimizing solutions, which differ between the two norms: The minimum-function solution is $(a, b) = (1, 1)$, and the minimum-gradient-energy solution is $(a, b) =(8/5, 2/5)$. The lower panels display the response curves and the derivatives with respect to the treatment, of the true response and the two minimizing solutions.

The first order effect of an infinitesimal nudge in the direction of $h_{a,b}'$ is the $L^2$ inner product of the true response derivative and the nudge direction, which can be computed as:
\begin{equation}\label{eq:illustration-policy-calculation}
 I(h_{a,b}') = \E[g'(X)h_{a,b}'(X)]=\frac{3a-4b}{2}.
\end{equation}
At the $L^2$ minimizing solution $(a, b) = (1, 1)$, the first order effect is $-1/2$, while at the gradient-energy minimizing solution $(a, b) =(8/5, 2/5)$, the first order effect is $8/5$. Hence, the minimum-function solution has a negative first-order effect, and the minimum-gradient-energy solution has a positive first-order effect. Note even the gradient-energy minimizing solution can still be harmful in some baseline treatment regions (e.g., close to zero baseline treatment): the claims we make are about average first-order effects under the baseline distribution. 

\begin{figure}[!htbp]
 \centering
 \includegraphics[width=0.8\linewidth]{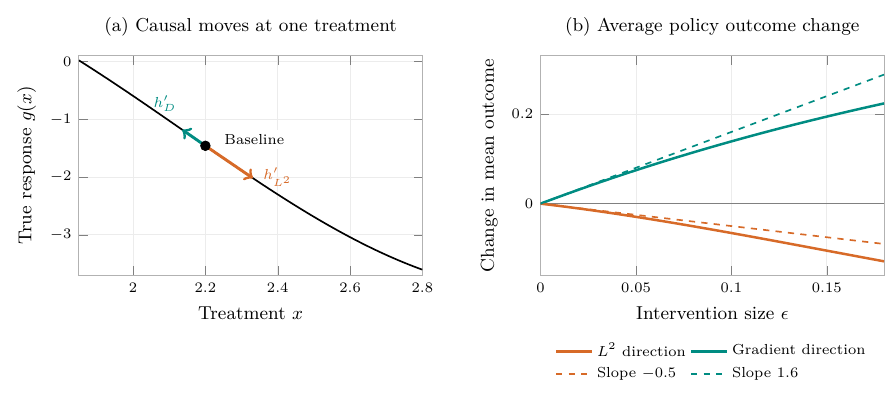}
 \caption{Policy directions have different causal consequences.
 Left: starting from treatment $x=2.2$, the orange direction moves right
 and lowers the response; the teal direction moves left and raises it
 (intervention size $0.12$). Right: average structural outcome change
 under the two interventions, computed via numerical integrals. Dashed lines show the first-order changes $\frac{3a - 4b}{2}$ at the minimizing solutions,
 $-\epsilon/2$ and $8\epsilon/5$ respectively. }
 \label{fig:illustration-policy}
\end{figure}

The left panel of \Cref{fig:illustration-policy} shows the direction of the nudge at $x=2.2$ (where they differ in sign) and the true response curve. At the displayed high treatment level, the minimum-function solution (orange) decreases the outcome and the minimum-gradient-energy solution (teal) increases it. The right panel shows the change in average outcomes, $\E[Y(X+\epsilon h_{a,b}'(X))] - \E[Y]$, as a function of step size $\epsilon$. We see that the gradient-energy minimizing solution (teal) improves the average outcome, while the $L^2$ minimizing solution (orange) decreases it across all displayed step sizes.

\begin{figure}[!htbp]
 \centering
 \includegraphics[width=0.8\linewidth]{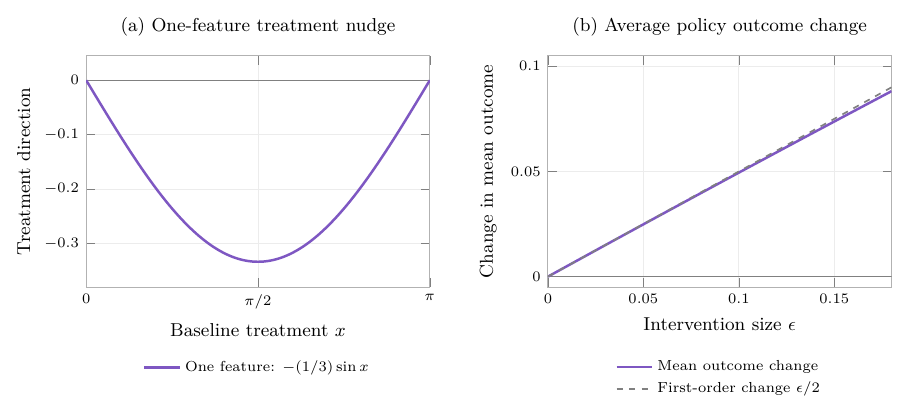}
 \caption{An improving nudge from a misspecified one-feature class.
 Left: $a_{\rm simple}(x)=-(1/3)\sin x$ decreases treatment
 in the interior and vanishes at the endpoints. Right: the
 average structural outcome change under this policy. The dashed
 line is its first-order approximation $\epsilon/2$.
}
 \label{fig:illustration-simple}
\end{figure}

\paragraph{An improving direction with a misspecified response class.} We now consider the mis-specified case when the hypothesis space of response functions we consider does not include the true response. Suppose we only use the slow feature $\phi(x)=\cos x$, i.e., we consider response functions of the form $h_a(x)=a\,\cos x$, which cannot represent $g$. We show that we can still construct a strictly improving direction using only observed covariances between the instrument and the outcome and between the instrument and the feature. Write
\[
 b=\E[ZY]=\frac13,\qquad
 B=\E[Z\phi(X)]=\frac16,\qquad
 G=\E[\phi'(X)^2]=\frac12.
\]
Among all response curves
$\alpha\,\phi(x)$, let $\alpha_\nabla\,\phi$ be the one minimizing
the discrepancy of the slopes, i.e., $\E[(g'(X)-\alpha\phi'(X))^2]$. Differentiating this quadratic objective gives the best approximate solution,
\[
 \alpha_\nabla=\frac{\E[g'(X)\phi'(X)]}{G}=3.
\]
Under a misspecified response class, the moment restrictions may not be satisfied by the minimizer. The moment violations for this approximation function and the zero function are, respectively,
\[
 |b-B\alpha_\nabla|
     =\left|\frac13-\frac16\cdot3\right|=\frac16,
 \qquad |b - 0|=\frac13.
\]
The gradient projection fits this moment better than zero does.
This simple comparison is enough to orient the policy. Observe that whenever $|b-B\alpha_\nabla|<|b|$, the projected approximate moment
$B\alpha_\nabla$ has the same sign as $b$. If we choose the nudge direction
\begin{equation}\label{eq:illustration-simple-direction}
 a_{\rm simple}(x)
   =\phi'(x)\sgn(b)\frac{B}{G}
   =-\frac13\sin x.
\end{equation}
then we get nonnegative first order average outcome improvement,
\[
 \E[g'(X)a_{\rm simple}(X)]
   =\sgn(b)B\alpha_\nabla
   \ge |b|-|b-B\alpha_\nabla|>0.
\]
\Cref{fig:illustration-simple} shows
what the resulting nudge looks like and how it changes the
average outcome.

This is a general population principle. A finite linear response family spanned by a known feature dictionary need not contain the truth. Provided that the largest absolute
moment violation at the best gradient approximation is smaller than at zero, we can construct a strictly improving direction without knowing the size of the misspecification bias. \Cref{sec:initial-direction} develops this general construction.

\FloatBarrier

\section{Related Work}
\label{sec:related}

Our work relates to three strands of literature: policy learning, and in particular incremental interventions on an existing policy; instrumental variable regression, with weak instruments, under ill-posedness, and under failure of identification; and the regularization and estimation methods that have been developed for such inverse problems. We discuss groups of closely related works within each strand, and discuss how our work differs.

\paragraph{Policy learning.}
Our work belongs to the broader literature on policy learning from observational and experimental data, which has been studied extensively in econometrics and statistics \citep{manski2004,hirano2009,qian2011,zhao2012,kitagawa2018,athey2021,zhou2023} and in computer science and machine learning \citep{beygelzimer2009,dudik2011,swaminathan2015,kallus2018continuous,chernozhukov2019,foster2023}. The majority of this literature considers \emph{global} interventions, i.e., the goal is to learn a policy that administers an optimal treatment to each individual, irrespective of the natural treatment that the individual would have received under the existing policy. Moreover, the value of the candidate policies is assumed to be identified from the data. 
Our work focuses instead on \emph{incremental} interventions, which modify the existing treatment policy based on each individual's natural treatment value. This is a more modest goal, and for that reason it does not require the value of any candidate policy to be identified: we only look for a small modification of the existing policy that is guaranteed not to decrease its value, which is possible even when the instrument is too weak to identify the response to the treatment.

\paragraph{Interventions that depend on the natural treatment value.}
Several works consider interventions that, like ours, modify the treatment that each unit would naturally receive.
\citet{stock1989} studies nonparametric evaluation of policy-induced changes
in the distribution of regressors. \citet{diaz2012} develop population
effects of stochastic treatment shifts. Interventions tied to natural
treatment values also appear in earlier work
\citep{robins2004,taubman2009}. \citet{haneuse2013}
modify each unit's naturally received dose and propose estimators for the resulting
effect under policy-specific exchangeability and support conditions that imply point identification.
\citet{young2014} and \citet{diaz2023} develop identification and estimation
for longitudinal interventions that depend on natural treatment values.
\citet{sarvet2026} review the role of natural treatment values and the
identification issues they raise.

A related use of ``incremental'' concerns stochastic changes to
the treatment assignment mechanism; see \citet{bonvini2023} for a review.
\citet{kennedy2019} multiplies binary treatment odds by a positive factor,
obtaining identifiable effects without treatment positivity, while
retaining sequential exchangeability. These interventions allow finite
changes and are different from transporting each unit's natural dose.
\citet{kim2021} extend this framework to dropout and many time points,
and \citet{mcclean2024} study conditional incremental effects and their
derivatives. \citet{ying2025} consider shifts to treatment-initiation
intensities. \citet{zhao2024} learn covariate-dependent incremental
propensity score policies without positivity, under unconfoundedness.
For continuous treatments, \citet{diaz2020mediation} consider exponential
tilting in mediation analysis, \citet{schindl2026} develop efficiency
theory and estimation rates for incremental exponential tilts, and
\citet{huang2026} extend tilting to multivariate exposures and compare
intervention directions. \citet{kallus2024natural} study efficient
evaluation of tilting and modified treatment policies in sequential
decision problems. These results clarify the benefits of staying close
to the observed assignment mechanism (e.g., by removing a positivity requirement).
Our intervention $X\mapsto X+\epsilon a(X)$ is an infinitesimal modified
treatment policy, whose direction can vary with the natural treatment.
However, we differ from this line of work in two respects. First, we consider infinitesimal nudges to a continuous treatment, as opposed to finite changes of the treatment or of its assignment mechanism. Second, and more importantly, the works above identify and estimate the effect of a \emph{given} intervention, typically under exchangeability conditions that rule out unmeasured confounding. We allow the treatment to be endogenous, and we focus on finding an intervention with guaranteed improvement, rather than on identifying the effect of a given one.

\paragraph{Infinitesimal shifts and average derivative effects.}
Several works address closely related derivative targets.
\citet{rothenhausler2019} study infinitesimal shifts of a continuous
treatment and identify their effects under local ignorability and
local overlap, even when average treatment effects are not identified. \citet{athey2021} study policy learning from observational data, and among other results, also consider instrumental variables and infinitesimal nudges of a continuous treatment (with conditional exogeneity), under conditions that identify the relevant policy value.
\citet{chernozhukov2022riesz} study inference on weighted average
directional derivatives, among other regression functionals, using
regularized Riesz representers. \citet{hines2026} develop continuous
treatment effect summaries with weighted average derivative
interpretations. Moving from evaluation to policy learning,
\citet{marmarelis2024} evaluate and optimize localized, multivariate
treatment nudges through doubly robust estimation and optimal transport,
under conditional ignorability given observed covariates. Thus,
infinitesimal shifts, treatment-dependent directions, and learning small
nudges all have direct precedents. What these works have in common is that the effect of a given direction is identified, through ignorability of the treatment. In our setting the treatment is endogenous, and the instrument might be too weak to identify the effect of any prescribed direction. The distinction we pursue concerns
which directions can be guaranteed to improve outcomes when the
available identifying restrictions leave the response unidentified. Hence, we focus on guaranteed improvement as opposed to identification of the effect of a given direction.

\paragraph{Instruments and partial identification in policy evaluation and learning.}
Instrumental variables and partial identification also have
important precedents in this literature. \citet{carneiro2010} identify
marginal policy-relevant treatment effects in a latent-index model with
endogenous binary treatment and continuous instruments.
\citet{zhou2022} relate such effects to marginal interventional effects,
with identification results both under unconfoundedness and using
instruments. \citet{mogstad2018} construct IV-based bounds on
policy-relevant treatment parameters when point identification is
unavailable. Optimal treatment rules for a binary treatment have also been learned with the help of an instrument, either under additional assumptions that identify the optimal rule \citep{cui2021,qiu2021}, or by minimizing the worst-case risk over the partially identified treatment effects \citep{pu2021}. Without instruments, \citet{kallus2018} study policy
improvement under a sensitivity model for unobserved confounding.
\citet{levis2024} study generalized policies that depend on natural
treatment and show how this dependence can yield sensitivity bounds
that collapse to the observed outcome as the policy approaches no
intervention. All of these works take the policies of interest as given, and either impose additional structure that identifies their effects or bound these effects when they are not identified. Our contribution concerns a different decision problem. Within our
additive structural model, we use valid conditional IV moments to
\emph{select} a direction with nonnegative first-order improvement for
every compatible response in the maintained function space. This
guarantee does not require the moments to identify the structural
response or the effects of arbitrary prescribed directions.

\paragraph{Weak instruments.}
Instruments that are valid but only weakly correlated with the endogenous treatment have been studied extensively in the linear instrumental variable model; see \citet{stock2002} and \citet{andrews2019} for surveys. When the instruments are weak, the two-stage least squares estimator can be badly biased and conventional inference can be misleading, even in large samples \citep{staiger1997}. The remedy is to use tests and confidence sets for the structural coefficients that remain valid irrespective of the strength of the instruments \citep{anderson1949,kleibergen2002,moreira2003}. The price of this robustness is informativeness: \citet{dufour1997} shows that a valid confidence set for a parameter that can be arbitrarily close to unidentified must be unbounded with positive probability. The question that we ask about a weak instrument is not what can be \emph{learned} about the structural response, but what can be \emph{done}. We do not construct a confidence set for the response, and we do not need the instrument to be strong enough to estimate it. We select a direction with a guarantee of nonnegative first-order improvement, and the strength of the instrument only determines the size of the guaranteed improvement, i.e., the identified gradient energy, and not the validity of the guarantee. Moreover, our response function is nonparametric, so that the instrument can fail to identify it even in the population limit.

\paragraph{NPIV as an identified but ill-posed inverse problem.}
Nonparametric instrumental variable (NPIV) regression is an important problem in the econometrics and statistics literature \citep{newey2003,ai2003,hall2005,blundell2007,carrasco2007,darolles2011,horowitz2011,chen2012,chen2018}. A major issue in the NPIV literature is that of ill-posedness: the linear inverse problem can be identified but unstable, so that small moment errors can produce large estimation errors. Several works have considered regularization techniques to address the ill-posedness of the NPIV problem; as opposed to addressing the lack of identification. Tikhonov ($L^2$ function-value) regularization has been studied by
\citet{hall2005,darolles2011}.
\citet{chen2012} study penalized sieve estimation of general conditional
moment models, relating moment fit to estimation error through a
modulus of continuity. These analyses treat NPIV as an identified but ill-posed  inverse problem: the solution is unique, and the goal is to recover it, at a rate that deteriorates with the degree of ill-posedness. We depart from both premises: we allow the moments to leave the structural function unidentified, and we do not attempt to recover it. Our target is a direction of policy improvement, and our improvement guarantee does not depend on the degree of ill-posedness.

\paragraph{NPIV without identification.}
Failure of identification (which we study) is distinct from instability of an identified inverse. In the latter case, the solution is unique but unstable to small perturbations due to finite sample noise, while in the former case, even in the population limit, the moment equations admit several structural
responses. \citet{phillips1989} studies rank failures in simultaneous
equations, and \citet{severini2006} study identification issues in
nonparametric regression with endogenous regressors.
\citet{florens2011} analyze Sobolev-penalized
estimation in nonidentified NPIV and its convergence to the solution of
minimal Sobolev norm, which is the closest prior work to our gradient-based
selection. \citet{babii2025} show that spectral regularization estimators,
such as Tikhonov regularization in $L^2$, converge in nonidentified linear inverse models to the projection of the structural function onto the orthogonal complement of the null space of the operator, its best approximation in function values.
\citet{bennett2023} develop minimax estimation of the minimum-function solution without uniqueness or closedness. These works characterize which solution a regularized estimator converges to when identification fails. Whether that solution can be used to make a decision is not addressed, and it turns out to depend on the norm. Our counterexample shows that the gradient of the minimum-function solution can decrease policy value. We instead project in gradient space, which characterizes the policy derivative and distinguishes existence of a minimizing function from existence of a
limiting minimum-gradient-energy direction. These arguments use Hilbert-space
projection \citep{bauschke2017}  and, for existence, weak compactness and Poincar\'e inequalities \citep{evans2010}. None of these prior works connect gradient-based regularization with guaranteed local policy improvement under nonidentification.

\paragraph{Identified sets and identified functionals.}
Several papers instead study the identified set and inference
under nonidentification. \citet{santos2012} develops hypothesis tests and
confidence regions in partially identified NPIV, and \citet{hong2017}
extends this approach to semiparametric conditional moment models,
including nonsmooth moments. \citet{liao2011} establish quasi-Bayesian
posterior concentration near the identified set.
\citet{freyberger2015} use shape restrictions to bound linear
functionals with discrete regressors and instruments, while
\citet{escanciano2013} characterize identified intervals under norm
constraints.

Nonidentification of the response also leaves open the
possibility that a particular functional is identified.
\citet{santos2011} develops estimators of identified continuous linear
functionals, including weighted average derivatives, using a selected
representative of a nonidentified nuisance function.
\citet{severini2012} derive efficiency bounds for linear functionals
without requiring identification of the response itself.
\citet{chen2021partial} studies slope estimation in partially linear IV
models with a partially identified nonparametric component.
\citet{bennett2026functionals} develop debiased minimax inference under a
strong identification condition on the functional, allowing nonunique
nuisance solutions. Both lines of work describe what the moments and additional restrictions reveal about the response: its identified set, or the value of a functional that happens to be identified. We use the compatible responses to make a decision instead: we select a direction with a policy improvement guarantee, as opposed to estimating a particular functional of the response or its identified set. Our main improvement result concerns the choice of
direction: minimizing gradient energy produces a lower bound on its
effect without assuming that a prespecified derivative functional is
identified or regularly estimable.

\paragraph{Choice of the regularization norm.}
Choosing a regularizer determines which aspects of a response are
penalized and, when the moments do not identify the response, which feasible
response is selected. Function-value $L^2$ penalties appear in
\citet{hall2005,darolles2011}. Sobolev penalties incorporate derivatives.
\citet{florens2011,gagliardini2012} study penalization by a Sobolev norm that includes the $L^2$ norms of the function and its (higher-order) derivatives. Derivative roughness penalties also appear in
sieve IV estimation \citep{blundell2007}, and kernel IV uses squared
reproducing-kernel Hilbert space norms \citep{singh2019kiv}.

The broader inverse-problem literature studies generalized
Tikhonov penalties $\|Lh\|_2^2$, including differential operators $L$,
and regularization on Hilbert scales \citep{engl1996,gupta2018}.
Other choices include weighted coefficient $\ell_p$ penalties for
$1\leq p\leq2$ \citep{daubechies2004}, elastic-net combinations of
$\ell_1$ and squared $\ell_2$ penalties \citep{chen2017elastic}, and
bounded-variation or total-variation penalties \citep{acar1994}.
A coefficient norm depends on the chosen representation and need not
equal a function-value $L^2$ norm. Likewise, total variation penalizes
the magnitude of a derivative measure rather than its squared $L^2$
norm. \citet{benning2018} survey these and more general convex
variational penalties. There is also a direct NPIV precedent for a gradient-only
penalty. \citet{grasmair2013} study shape-constrained instrumental
regression with penalty
$\mu\|h\|_{L^2(\Omega)}^2+\|\nabla h\|_{L^2(\Omega)}^2$,
explicitly including $\mu=0$. They establish existence, stability,
consistency, and source-based convergence guarantees in that setting.

In the works above, the penalty is a device for stabilizing an ill-posed problem, and it is assessed by the estimation rates that it yields. Our contribution concerns the policy consequences of this choice of
geometry. While some of these regularization choices are related to the gradient energy, none of these works connects the choice of regularization to guaranteed policy improvement under nonidentification. This connection
lets us construct a direction with nonnegative first-order improvement
under valid moments that need not identify the response, and develop
statistical guarantees for that direction. In the case of linear sieves, our gradient-energy penalty provides a particular form of coefficient Mahalanobis norm that should be used to guarantee policy improvement and which does not coincide with the $\ell_p$ norms of the coefficients, other than in special cases. Thus the distinction is the
policy-improvement guarantee and its analysis under nonidentification,
rather than the introduction of derivative regularization.

\paragraph{Statistical analysis of adversarial and two-stage estimators.}
\citet{dikkala2020} analyze conditional moment estimation over general
hypothesis classes for the response and test classes for the instrument,
using a minimax criterion and localized Rademacher complexities.
\citet{bennett2023source}
also analyze a Tikhonov-regularized quadratic adversarial criterion with a
function-value penalty.
\citet{hartford2017} introduce the two-stage DeepIV criterion, and
\citet{li2024} study its Tikhonov-regularized version with a
function-value penalty.
Our empirical-process argument used to prove a lower bound of first order policy improvement uses Rademacher and localized Rademacher complexities, following the analysis of
\citet{bartlett2005}, using vector contraction \citep{maurer2016} and
bounded empirical-process concentration
\citep{boucheron2013}, and is related to the
localized learning-theoretic analysis of
\citet{foster2023}.
The guarantees of the estimators above concern the recovery of the structural function. Our main moment constraint estimator adopts this adversarial view of test functions but
imposes the moments as constraints and minimizes gradient energy, and what we guarantee is a lower bound on the first-order policy improvement of the estimated direction. This guarantee requires neither a bound on the ill-posedness of the inverse problem (e.g., a source condition), nor an estimate of the response in $L^2$ distance, nor any richness of the test function class. We also discuss how adapt the
minimax and DeepIV criteria by replacing their function-value penalties
with gradient energy.

\paragraph{High-dimensional linear IV.}
Limited instruments can leave a coefficient vector
underidentified even in a finite-dimensional parametric response model.
\citet{lochner2015} study weighted-average interpretations and
exogeneity testing when instruments do not identify all treatment-level
effects. \citet{gautier2022rose} construct self-normalized moment
confidence sets that accommodate underidentified IV systems, with
optional sparsity restrictions. \citet{dikkala2020} give sparse linear applications of adversarial conditional moment estimation, including $\ell_1$ relaxations under restricted eigenvalue assumptions. \citet{gautier2011} give nonasymptotic coefficient bounds and confidence sets for sparse regression with many endogenous variables, and
\citet{belloni2012} use sparse first-stage regression to construct
optimal instruments for a low-dimensional structural
parameter. Our linear hypotheses specialization does not aim to recover the coefficient vector, which is the target of the works above (its interpretation, its estimation, or confidence sets for it). It delivers a policy improvement bound, which allows the
structural coefficient vector itself to be high dimensional and tracks
the derivative geometry needed for a policy bound. Moreover, our guarantees in the case of a sparse high-dimensional setting require no restricted strong convexity conditions \citep{bickel2009,negahban2012}, which are typical in the literature.

\section{Problem Statement and Preliminaries}
\label{sec:setup}

Let $W=(X,Z,Y)$ be an observation, with $X\in\R^d$ a treatment
generated by a baseline policy, $Z \in \R^{d_Z}$ an instrument, and $Y\in \R$ an observed outcome. We assume the additive structural model\footnote{We use $\E$ for the population expectation and $\E_n$ for the empirical average.} 
\begin{equation}\label{eq:structural}
 Y(x)=g(x)+U,\qquad Y=g(X)+U,\qquad \E[U\mid Z]=0.
\end{equation}
Independence of $U$ and $Z$ is sufficient for the conditional restriction to hold, as a nonzero mean can be absorbed into the
intercept of $g$. We assume $Y,g(X),U$ are square integrable random variables. Let $P_X$ and $P_Z$ be the marginal laws of $X$ and $Z$ respectively. The baseline treatment can depend
on $U$ and $Z$, i.e., $X$ is an endogenous treatment. 

\paragraph{Structural response space.} We further assume that the structural response $g$ lies in a known space $\V\subseteq \Vbb$, which is a convex subset of a linear space $\Vbb\subseteq L^2(P_X)$ of square-integrable functions with a well-defined linear gradient map
\begin{align}
 D~&:\Vbb\to L^2(P_X;\R^d), \qquad Dh=\nabla h.
\end{align}
We assume that $\V$ contains constant functions. The space $\V$ is equiped with the $L^2(P_X)$ function norm, $\|h\|_2^2:=\E[h(X)^2]$, the gradient seminorm defined by the $L^2(P_X;\R^d)$ norm of its derivative $\|Dh\|_2^2:=\E[\|Dh(X)\|^2]$ and the graph norm, which combines the latter two norms:\footnote{We use $\|\cdot\|_2$ as a shorthand for $\|\cdot\|_{L^2(P_X, \R^d)}$, with $P_X$ the marginal law of the input variable to the function; inner product shorthand $\ip{\cdot}{\cdot}$ is taken under the same distribution.}
\begin{align}
\|h\|_{\V}^2 = \|h\|_2^2 + \|Dh\|_2^2. \label{eqn:graph_norm}
\end{align}

\paragraph{Conditional moment constraints.} The conditional restrictions $\E[U\mid Z]=0$ imply the following conditional moment constraints that a candidate structural reponse $h\in \V$ needs to satisfy:
\begin{equation}\label{eq:moment}
 \E[Y-h(X)\mid Z]=0.
\end{equation}
For notational convenience, we define the conditional expectation operator
\begin{align}
  T~&:L^2(P_X)\to L^2(P_Z),\qquad Th=\E[h(X)\mid Z],
\end{align}
We call the conditional outcome $r = Tg=\E[Y\mid Z]$ the reduced form.
Since we are allowing for partial identification, $T$ need not be an injective map. With this notation at hand, we can write the feasible NPIV set as
\[
\Cc=\{h\in\V:Th=Tg\},
\]
and the null space (a.k.a. kernel) of the operator $T$ as 
\[
K=\ker T= \{v\in\mathbb{V}:Tv=0\}.
\]
Observe that $\Cc=(g+K)\cap \V$, and nonidentification implies $\Cc$ has more than one response. We reserve $\F$ for classes of instrument test functions.

\subsection{First order policy improvement}

Our goal is to design a local perturbation of the baseline treatment, i.e., $X \to X + \epsilon a(X)$, in some direction $a$ in the gradient space $L^2(P_X;\R^d)$, such that the outcome improves in expectation for sufficiently small values of $\epsilon$. To understand when a treatment perturbation yields policy value improvement, consider
what happens to a single unit under such a local intervention. The potential outcome under the intervention is $Y(X+\epsilon a(X))=g(X+\epsilon a(X))+U$. The baseline outcome is $Y(X)=g(X)+U$. The unobserved term $U$ is the same in both cases, and
therefore cancels when we compare the new outcome with the baseline
outcome, leaving only the change in the structural response:
\begin{align*}
Y(X+\epsilon a(X))-Y(X)=g(X+\epsilon a(X))-g(X).
\end{align*}
For small enough $\epsilon$, by a first-order Taylor expansion argument, the latter can be well approximated by the directional derivative of $g$ along $a$:
\begin{align*}
g(X+\epsilon a(X))-g(X)\approx \epsilon \nabla g(X)^{\mathsf T}a(X).
\end{align*}
Averaging
these changes and taking the limit as $\epsilon \to 0$, gives the policy derivative $I(a)$, provided differentiation can
be interchanged with expectation.
\begin{definition}[First-order effect and improving direction]
For a direction $a\in L^2(P_X;\R^d)$, define the policy derivative $I(a)$ as the inner product of the gradient of the structural response with the direction $a$:
\begin{align}
I(a):=\ip{Dg}{a}=\E[\nabla g(X)^{\mathsf T}a(X)].
\end{align}
We refer to $I(a)$ as the first-order effect of the local treatment perturbation in direction $a$. We say a direction $a$ has the first-order improving property if $I(a)\ge0$.
\end{definition} 
The following proposition supplies
conditions for this step and shows how a positive derivative can also
translate into improvement for a sufficiently small finite intervention.

\Needspace{6\baselineskip}
\begin{proposition}[Policy derivative]
\label{prop:derivative}
Let $a\in L^2(P_X;\R^d)$. Suppose $g$ is continuously differentiable along the segments
$X+\epsilon a(X)$ for $|\epsilon|\le \epsilon_0$, for some $\epsilon_0>0$. Assume these segments are admissible treatment values, and
\[
 \sup_{|\epsilon|\le \epsilon_0}|\nabla g(X+\epsilon a(X))^{\mathsf T}a(X)|\le M_a
 \quad\text{a.s.},\qquad \E M_a<\infty.
\]
Then, with $W_a(\epsilon)=\E[Y(X+\epsilon a(X))]$,
\begin{equation}\label{eq:welfare-derivative}
 W_a'(0)=I(a).
\end{equation}
In particular, the condition holds if $g$ has an $L$-Lipschitz gradient
on the relevant segments and $Dg,a\in L^2(P_X)$.
In that case, for every $0\le \epsilon\le \epsilon_0$,
\begin{equation}\label{eq:finite-step}
 W_a(t)-W_a(0)\ge \epsilon I(a)-\frac{L \epsilon^2}{2}\norm{a}_2^2.
\end{equation}
\end{proposition}

\begin{proof}
By the structural model, we have
$Y(X+\epsilon a(X))-Y(X)=g(X+\epsilon a(X))-g(X)$ pointwise. Applying the fundamental theorem of calculus along each
segment and then dominated convergence yields \eqref{eq:welfare-derivative}.
When the gradient is Lipschitz,
\[
 |\nabla g(X+\epsilon a(X))^{\mathsf T}a(X)|
 \le \norm{Dg(X)}\norm{a(X)}+L t_0\norm{a(X)}^2.
\]
The right side is integrable by Cauchy--Schwarz. For the finite-step bound, the Lipschitz-gradient remainder gives
$g(X+\epsilon a)-g(X)\ge \epsilon\nabla g(X)^{\mathsf T}a-L\epsilon^2\norm{a}^2/2$.
Taking expectations proves \eqref{eq:finite-step}.
\end{proof}

This proposition connects the inner product $I(a)$ to the change
in policy value. A strictly positive derivative leads to an average outcome increase for all sufficiently small positive steps. All subsequent guarantees concern $I(a)$; interpreting them as guarantees for an intervention requires the admissibility and differentiation conditions of \cref{prop:derivative}. %

\section{Policy Improvement via Minimum-Gradient-Energy Selection}
\label{sec:population}

If the identified set $\Cc$ contains only the true response function $g$, then the obvious direction for policy improvement is its gradient $Dg$, which would yield a first order effect of $I(a)=\|Dg\|_{2}^2$. For this reason, we will restrict attention to perturbation directions $a$ that correspond to the gradient $Dh$ of some candidate response function $h$. When $\Cc$ contains multiple functions, then there may exist functions $h\in \Cc$ in the feasible set, such that if we select them and use their gradient $Dh$ as the perturbation direction, then we might end up with a negative first-order improvement $I(Dh)=\ip{Dg}{Dh}$. We have already seen such an example in Section~\ref{sec:illustration}, where selecting the minimum function $L^2$ norm solution in the feasible set, i.e. $h_{L^2}=\arg\min_{h\in \Cc} \|h\|_2^2$, and perturbing treatment along its gradient, can yield negative first order improvement. 

Our main result, which we present in this section, is that selecting the response function that minimizes the gradient energy among feasible responses and perturbing along its gradient, guarantees first order improvement. A central concept in our exposition will be the infimum of the gradient energy among all functions in the feasible set, which we call the \emph{identified gradient energy}.

\begin{definition}[Identified gradient energy]
    The gradient energy of a function $h$, $\mathcal E(h)$, is the squared $L^2(P_X;\R^d)$ norm of its gradient:
    \begin{align}
     \mathcal E(h):=\norm{Dh}_2^2=\E\bigl[\|Dh(X)\|^2\bigr],
    \end{align}
    The identified gradient energy, $\kappa$, is the infimum of the gradient energy among all functions in the feasible NPIV set:
    \begin{align}
        \kappa:=\inf_{h\in\Cc}\mathcal E(h).
    \end{align}
\end{definition}

Even when the infimum is not attained by some function $h$ in the feasible set, we show that any sequence of feasible functions whose gradient energy approaches the infimum, will yield a sequence of perturbation directions that converge to a direction $a_\star$ that guarantees first order improvement. This direction guarantees a first order improvement at least $\|a_\star\|_2^2$ irrespective of which of the feasible response functions is the true response. Moreover, for any sub-optimal function $h\in \Cc$ the first order improvement of its gradient is bounded below by $\kappa$, minus a term that depends on the gradient energy sub-optimality of $h$ (i.e., $\mathcal E(h) - \kappa$). When $\V$ is a linear space (and not just a convex space), the selected direction coincides with the true gradient after removing its projection onto the gradients of perturbations $v$ that the instrument cannot detect, i.e., $Tv=0$.

\begin{theorem}[Minimum-gradient-energy direction and first-order improvement]
\label{thm:closure-convex}
Let $\overline{D\Cc}=\cl(D\Cc)$ in $L^2(P_X;\R^d)$ denote the closure of the set $D\Cc$ of gradients of feasible response functions. Define the \emph{minimum-gradient-energy direction} as the vector $a_\star$ that minimizes the norm over the closure of the set of feasible gradients,
\begin{align}
a_\star=\arg\min_{a\in \overline{D\Cc}}\|a\|_2^2.
\end{align}
The minimizer exists and is unique. Every feasible minimizing sequence of the \emph{identified gradient energy} program, i.e., $(h_j)\subset \Cc$ with $\mathcal E(h_j)\to\kappa$, has gradients $Dh_j$ converging strongly in $L^2(P_X;\R^d)$ to $a_\star$ and the first-order improvement of perturbation $a_\star$ is at least the identified gradient energy, namely:
\begin{align}
 \kappa&=\|a_\star\|_2^2,& I(a_\star)&\ge\kappa\ge0,
 \label{eq:canonical-convex}
\end{align}
Moreover, for any feasible response $h\in \Cc$
\begin{align}
 \mathcal E(h)&\ge\kappa+\|Dh-a_\star\|_2^2, & 
 \langle a_\star,Dh\rangle&\ge\kappa.\label{eq:canonical-functional}
\end{align}
and
\begin{equation}\label{eq:approx-bound-convex}
 I(Dh)\ge I(a_\star)
 -\sqrt{(\mathcal E(g)-\kappa)(\mathcal E(h)-\kappa)}.
\end{equation}
Consequently, when $\kappa>0$, sufficiently accurate feasible
near-minimizers strictly improve. If $\mathcal E(g)>\kappa$, strict improvement holds whenever
$\mathcal E(h)-\kappa<\kappa^2/(\mathcal E(g)-\kappa)$.
If $\mathcal E(g)=\kappa$, every feasible $h$ has $I(Dh)\ge\kappa$.

Suppose additionally that, for every $c\in \overline{D\Cc}$, there is $t_c>0$ with
\begin{equation}\label{eq:convex-two-sided}
 a_\star-t_c(c-a_\star)\in \overline{D\Cc}.
\end{equation}
This holds if $a_\star$ is in the relative algebraic interior of $\overline{D\Cc}$
(and hence if it is in its relative interior in the norm topology),
or if $\overline{D\Cc}$ is affine, in particular when $\V$ is linear.\footnote{Relative algebraic interior means that every direction in
$\operatorname{span}(\overline{D\Cc}-\overline{D\Cc})$ admits a sufficiently short two-sided segment through $a_\star$ contained in $\overline{D\Cc}$.}
Then $I(a_\star)=\kappa$, the inequalities in
\eqref{eq:canonical-functional} hold with
equality, and every feasible $h$ satisfies
\[
 |I(Dh)-\kappa|\le
 \sqrt{(\mathcal E(g)-\kappa)(\mathcal E(h)-\kappa)}.
\]
If $\V$ is linear, let $K_\V=\V\cap\ker T$, $M=\cl(DK_\V)$, and let $\Pi_M$ be the orthogonal projection onto $M$. Then
\[a_\star=Dg-\Pi_M Dg\]
\end{theorem}

The proof of Theorem~\ref{thm:closure-convex} is defered to \Cref{sec:proof-closure-convex}.
When a minimizing function of the identified gradient energy program is 
available, the reason for improvement can
be seen without taking a closure. Starting at that function, we can move
in the direction of any feasible response and
remain feasible. The gradient energy cannot decrease under such a
move, so its first-order change must be non-negative. This gives the lower bound on first-order improvement. In the case of a linear space $\V$, the first-order optimality condition is equivalent to orthogonality of the minimizing gradient to the gradients of all feasible perturbations around the true response. The first-order improvement is then exactly the squared norm of the minimizing gradient, which is at least the identified gradient energy.

\Needspace{6\baselineskip}
\begin{corollary}[Policy improvement from a minimum gradient solution]
\label{thm:attained-convex-draft}
\label{thm:attained}
If $h_\star\in\Cc$ minimizes $\mathcal E$, then
\begin{equation}\label{eq:attained-identity-convex-draft}
 \langle Dh_\star,D(h-h_\star)\rangle\ge0\quad(h\in\Cc),
 \qquad I(Dh_\star)\ge\|Dh_\star\|_2^2=\kappa\ge0.
\end{equation}
If $h_\star-t(g-h_\star)\in\V$ for some $t>0$, then
$I(Dh_\star)=\kappa$. If the corresponding reverse step is feasible
for every $h\in\Cc$, the first inequality is an equality for every
such $h$. A sufficient condition is that $h_\star$ belongs to the
relative algebraic interior of $\V$; linear $\V$ is a special case.
\end{corollary}
\begin{proof}
For $h\in\Cc$, convexity makes $h_\star+s(h-h_\star)$ feasible
for $0\le s\le1$ and $h\in \Cc$. The energy of this feasible response function has a right derivative at zero with respect to $s$, i.e., $\lim_{s \downarrow 0} \frac{{\cal E}(h_\star + s(h-h_\star)) - {\cal E}(h_\star)}{s}$, equal to
$2\langle Dh_\star,D(h-h_\star)\rangle$, which must be nonnegative.
Taking $h=g$ proves the lower bound. A reverse feasible step has the
same moments and gives the opposite sign in the first-order
inequality, proving the equality assertions. Relative algebraic
interiority permits such a reverse step toward every point of $\V$.
\end{proof}

\Cref{thm:attained-convex-draft} requires a feasible response that achieves the smallest gradient energy. In familiar smooth response models, there is a simple reason to expect such a response to exist. Consider a sequence of feasible functions whose energies approach $\kappa$. We need to keep their average squared values bounded as their gradients settle down to $\kappa$. The moments already fix their common mean. If the functions in $\V$ also satisfy what is known as a Poincar\'e inequality, then we would be able to bound their variation around this mean by their gradient energy. Together, these observations control both function values and gradients, the two quantities that appear in the graph norm \eqref{eqn:graph_norm}.

\begin{assumption}[Poincar\'e inequality]\label{ass:poincare}
For some $0<C_P<\infty$,
\begin{equation}\label{eq:poincare}
 \Var(h(X))\le C_P\,\mathcal E(h)\qquad(h\in\V).
\end{equation}
\end{assumption}

The remaining issue is whether the response class contains the limit that the minimization argument produces. A sufficient requirement is that convergence in function values and gradients does not take us outside the class, namely completeness under the graph norm. For a convex class, \cref{thm:existence} shows that this completeness property and the Poincar\'e inequality guarantee attainment. All minimizers share the same gradient (see \cref{lem:gradient-uniqueness}).

These requirements hold for the standard class of responses with square-integrable values and weak first derivatives on a bounded, connected region with a regular boundary, such as a box or ball, when the treatment density is bounded above and away from zero (see \Cref{thm:existence}). Attainment also holds when $\V$ is complete and the convex feasible set is bounded with respect to the graph norm. Attainment should therefore be viewed as the typical case in these regular models. It can fail when $P_X$ has regions of very low density, which allow for sharp changes in the function with almost no cost on gradient energy; in such scenarios the infimum can fall outside of the class of feasible gradients $D\Cc$. Even then, \cref{thm:closure-convex} gives an improving limiting direction. Feasible near-minimizers also improve eventually when $\kappa>0$.

\begin{remark}[Gap between $\mathcal E(g)$ and $\kappa$]\label{rem:energy-gap}
The identified gradient energy $\kappa$ is the energy of the best detectable gradient, while the true gradient energy $\mathcal{E}(g)$ is the energy of the true gradient. The two can be substantially different. The gap between $\mathcal{E}(g)$ and $\kappa$ can be best understood from the projection characterization used in
\cref{thm:closure-convex} in the case when $\V$ is a linear space: it is the energy of the orthogonal projection of the true
gradient onto the undetectable gradients, $\mathcal E(g)-\kappa=\norm{\Pi_M Dg}_2^2$.
\end{remark}

\begin{remark}[Minimum function-value selection]\label{rem:l2-failure}
The minimum-function solution $h_{L^2}$, i.e., the feasible function with the
smallest $L^2$ norm, satisfies the analogous first-order condition in
function values, $\ip{h_{L^2}}{v}_{L^2(P_X)}=0$ for $v\in\V\cap\ker T$, in the case when $\V$ is linear. Writing
\[
 I(Dh_{L^2})=\norm{Dh_{L^2}}_2^2+\ip{D(g-h_{L^2})}{Dh_{L^2}},
\]
the cross term no longer vanishes: $g-h_{L^2}$ is orthogonal to $h_{L^2}$ in
function values, but its gradient need not be orthogonal to $Dh_{L^2}$, and
the cross term can outweigh the first. The example of
\cref{sec:illustration} demostrates such a case. 
\end{remark}

\subsection{Optimality of minimum-gradient-energy over all safe improvement directions}
\label{sec:worst-case-improvement}

The population improvement guarantee leaves open a further question:
could another direction offer a larger improvement uniformly over the
responses compatible with the observed moments? This comparison is
especially relevant when the response is not identified. The data do
not tell us which compatible response is true, so a direction with a
large effect under one response may perform poorly under another.
We therefore compare directions through their worst-case first-order
effect over the feasible set. To make this comparison meaningful,
we must also fix the size of the nudge: multiplying a direction
multiplies its first-order effect by the same factor. We use an
$L^2$ norm budget, which controls the root mean squared treatment
perturbation, and allow all square-integrable directions.
The result below shows that the normalized minimum-gradient-energy
direction $a_\star / \|a_\star\|$ achieves the largest possible uniform guarantee under this budget. It thus gives a decision interpretation of the identified gradient energy, beyond its role as a lower bound for one selected
direction.

For a feasible response $h\in\Cc$ and direction
$a\in L^2(P_X;\R^d)$, write
\begin{equation}\label{eq:worst-case-effect}
 I_h(a):=\langle Dh,a\rangle,\qquad
 \underline{I}(a):=\inf_{h\in\Cc}I_h(a).
\end{equation}
Here $I_g(a)=I(a)$, and $\underline{I}(a)$ is the first-order
improvement guaranteed uniformly over compatible responses.
The infimum is allowed to equal $-\infty$.

\Needspace{6\baselineskip}
\begin{theorem}[Optimality of minimum-gradient-energy direction]
\label{thm:worst-case-improvement}
Suppose $\V$ is convex and $\Cc$ is
nonempty. Let $a_\star$ be the minimum-norm element of
$\overline{D\Cc}$ in $L^2(P_X;\R^d)$, and let
$\kappa=\inf_{h\in\Cc}\|Dh\|_2^2$. Then
\begin{align}
 \underline{I}(a_\star)&=\kappa, &
 \max_{\substack{a\in L^2(P_X;\R^d)\\ \|a\|_2\le1}}
       \underline{I}(a)&=\sqrt{\kappa}.
\label{eq:worst-case-selected}
\end{align}
If $\kappa>0$, the unique optimizer of $\underline{I}(a)$ over the unit ball, as an element of $L^2(P_X;\R^d)$, is 
\[
\bar{a}_\star :=a_\star/\|a_\star\|_2. 
\]
If $\kappa=0$, the maximum over the unit ball
is zero and is attained by $a=0$.
\end{theorem}

\begin{proof}
Set $A=\overline{D\Cc}$. The closed convex projection argument of
\cref{thm:closure-convex} gives
\[
 \|a_\star\|_2^2=\kappa,\qquad
 \langle a_\star,c\rangle\ge\kappa\quad(c\in A).
\]
For each fixed $a$, the functional $c\mapsto\langle c,a\rangle$
is continuous. Density of $D\Cc$ in $A$ therefore implies
\[
 \underline{I}(a)=\inf_{c\in A}\langle c,a\rangle.
\]
The projection inequality gives
$\underline{I}(a_\star)\ge\kappa$, while $a_\star\in A$ gives
the reverse inequality. This proves $\underline{I}(a_\star)=\kappa$.
For any $a\in L^2(P_X;\R^d)$, inclusion of $a_\star \in A$ and
Cauchy--Schwarz give
\begin{equation}\label{eq:worst-case-upper-bound}
 \underline{I}(a)=\inf_{c\in A} \ip{c}{a} \le\langle a_\star,a\rangle
 \le\sqrt{\kappa}\,\|a\|_2.
\end{equation}
Thus the worst-case effect of any direction in the unit ball
is at most $\sqrt{\kappa}$. This proves \eqref{eq:worst-case-selected}. 
When $\kappa>0$, positive homogeneity
of the infimum yields
\[
 \underline{I}\!\left(\bar{a}_\star\right) = \underline{I}\!\left(\frac{a_\star}{\sqrt{\kappa}}\right)
 =\frac{\underline{I}(a_\star)}{\sqrt{\kappa}}
 =\sqrt{\kappa}.
\]
This direction has unit norm, so it attains the upper bound. If another direction in the unit ball
attains this value, both inequalities in
\eqref{eq:worst-case-upper-bound} and $\|a\|_2\le1$ must be
equalities. Since $a_\star\ne0$, equality in Cauchy--Schwarz
then forces $a=a_\star/\sqrt{\kappa}$, proving uniqueness.
Finally, when $\kappa=0$, \eqref{eq:worst-case-upper-bound}
bounds every worst-case effect by zero, and
$\underline{I}(0)=0$ attains that bound.
\end{proof}

The theorem identifies $\sqrt{\kappa}$ as the best worst-case first-order effect per unit $L^2$ size of the direction. When $\kappa>0$, the normalized direction $\bar a_\star=a_\star/\|a_\star\|_2$ attains this bound. Its first-order effect under the actual response can be strictly larger, as is possible for a convex response class, but no larger lower bound holds uniformly over $\Cc$. To see this, suppose $\kappa>0$. For every candidate direction $a\in L^2(P_X;\R^d)$ satisfying
$\|a\|_2\le 1$ and $a\ne \bar{a}_\star$, there exists a feasible
response $h\in\Cc$ such that
\begin{equation}\label{eq:worst-case-strict-comparison}
 I_h(a)<\sqrt{\kappa}\le I_h(\bar{a}_\star).
\end{equation}
The feasible response may depend on $a$. To see why an actual feasible
response exists even without attainment, fix $a$ and take a feasible
minimizing sequence $(h_j)$ of the gradient energy program, which, by \cref{thm:closure-convex}, satisfies that
$Dh_j\to a_\star$, so
\[
 I_{h_j}(a)\longrightarrow\langle a_\star,a\rangle
 <\sqrt{\kappa}.
\]
The strict inequality follows from Cauchy--Schwarz: equality within
the unit ball would require $a=a_\star/\sqrt{\kappa}$.
Thus sufficiently late terms in the sequence supply a response satisfying the first inequality in \eqref{eq:worst-case-strict-comparison}; the second
holds for every feasible response by \cref{thm:worst-case-improvement}.

\section{Assumption-Lean Finite-Sample Policy Improvement}
\label{sec:statistics}

We now move to estimation. Let $W_i=(X_i,Z_i,Y_i)$, $1\le i\le n$, 
be independent observations from the population distribution described in \cref{sec:setup}. Given the samples, our goal is to select a policy direction $\hat{a}$ whose first-order effect is guaranteed to be non-negative, with high-probability, for a sufficiently large sample size. 

\Cref{sec:population} shows that moving the policy along the minimum-gradient-energy direction has non-negative first-order effect. This suggests a simple empirical procedure: 
select the response functions from a candidate hypothesis set whose empirical moments are close to zero, and among them choose 
one with small empirical gradient energy. The gradient of this function is then chosen as the policy direction. Given that the moment constraints are conditional ones, we turn them into empirical moment constraints using the test-function approach, as e.g., in \citet{dikkala2020}. The conditional moments $\E[(Y - h(X))\mid Z]=0$ are equivalent to $\sup_{f} \E[(Y - h(X))\,f(Z)]=0$ for all functions $f$. A large correlation between the residual 
of a candidate function and some function of the instrument signals a violated moment. The latter unconditional moments can be replaced by their empirical analogs and the test functions can be selected from a class of functions $\F$ of the instrument to avoid overfitting, i.e., $\sup_{f\in\F} \E_n[(Y - h(X))\,f(Z)]$. For the true response $g$, the population moments are zero, so the empirical moments are small with high probability, with the size of the violation related to the sample size $n$ and the complexity of the function class $\F$. This gives rise to the following moment-constrained, minimum empirical gradient energy estimator.

\paragraph{Minimum-gradient moment constraint estimator}
Given a hypothesis class $\Hc$ of differentiable candidate responses, an instrument test class $\F$, and a tolerance $\tau_n>0$, define the empirical feasibility set as:
\begin{equation}
\Cc_n = \left\{h\in\Hc: \sup_{f\in\F} \Pn[(Y-h(X))f(Z)] \le\tau_n\right\}.
\end{equation}
The estimator solves
\begin{align}\label{eq:moment_constraint_estimator}
\widehat h &\in \argmin_{h\in\mathcal C_n} J_{n}(h), & J_{n}(h) :=~& \tfrac12\Pn\bigl[\norm{Dh(X)}^2\bigr].
\end{align}
and returns the policy direction $\hat{a} = D\widehat h$. If $\Cc_n$ is empty, we set $\widehat h=0$ and $\hat{a}=0$.

We allow the hypothesis class $\Hc$ to not contain the true response $g$ (mis-specified case), so the tolerance $\tau_n$ must absorb both sampling error and approximation error, as there might not exist any $h\in \Hc$ that satisfies the population moments. In addition, we will allow for the solver of the minimization program to return a slighly suboptimal solution due to optimization error $\varepsilon_{\rm opt}$. The rest 
of the section shows how to choose $\tau_n$ and derives a lower bound on the first-order policy improvement in terms of the gradient energy of the estimated solution and these 
three errors. The guarantee requires only mild conditions of the candidate responses $\Hc$ and the test functions $\F$ (e.g., convexity and boundedness); it imposes no richness condition on the test class and no condition limiting how ill-posed the inverse problem is (e.g., a source condition \cite{bennett2023source} or a measure of ill-posedness \cite{chen2012}).

\subsection{Sampling error, Rademacher complexities and critical radii}

The sampling error depends on the statistical complexity of the hypothesis and test classes, 
which, as in \citet{dikkala2020}, we measure through global and localized Rademacher complexities and 
the associated critical radii \citep{bartlett2005}. 
We work with bounded, star-convex classes, where these measures directly control the 
sampling error; the following assumption collects these assumptions.
\Needspace{6\baselineskip}

\begin{assumption}[Bounded convex hypothesis class and test class]
\label{ass:statistical}
The hypothesis class $\Hc\subset\V$ is deterministic and convex,
and it contains the zero function (hence is also star-convex). For every $h\in\Hc$,
\begin{equation}\label{eq:boundedness}
 |Y|\le1,\qquad |h(X)|\le1,\qquad \norm{Dh(X)}_{\R^d}\le1
 \quad\text{almost surely}.
\end{equation}
The class of instrument test functions $\F \subset L^2(P_Z)$ is
deterministic, symmetric ($f\in \F$ implies $-f\in \F$), star-convex
about zero (closed under multiplication by scalars in $[0,1]$), and
bounded by one. We refer to these bounds as a uniform envelope of one.
The classes and the suprema appearing in our analysis below are measurable (e.g., by pointwise separability with measurable representatives).
\end{assumption}

The uniform envelope of one is a normalization; rescaling changes only constants. Note that the assumption restricts the hypothesis class, 
but not the true response $g$. 
The population objects of \cref{sec:population} ($a_\star$ and $\kappa$) are unchanged.
Both the hypotheses and test classes may depend on the sample size, 
but must be chosen before
seeing the data. 

We define the localized Rademacher complexity: for a radius 
$\delta>0$ and a class $\mathcal B$ of real-valued functions of $W$, let
\[
 \mathfrak R_n(\mathcal B,\delta)=
 \E\sup_{b\in\mathcal B:\norm{b}_2\le \delta}
       \left|\frac1n\sum_{i=1}^n\sigma_i b(W_i)\right|.
\]
For a class $\mathcal B$ of vector-valued functions of $X$, similarly define
\begin{equation} \label{eqn:complexity}
     \mathfrak R_n^{\text{vec}}(\mathcal B,\delta)=
 \E\sup_{b\in\mathcal B:\norm{b}_2\le \delta}
       \left|\frac1n\sum_{i=1}^n\sum_{j=1}^d
                    \sigma_{ij} b_j(X_i)\right|.
\end{equation}
Here the signs $\sigma_{i}$ and $\sigma_{ij}$ are mutually independent Rademacher random variables, independent of the data, and $\norm{b}_2$ denotes the population $L^2$ norm under the law of the argument of $b$.
The supremum runs over a ball of radius $\delta$ in the population norm, and $\mathfrak R_n$ averages
over both the data and the signs.

The sampling error depends on the statistical complexity of three classes: the gradients, which enter
the policy objective; the test functions, which determine empirical feasibility; and the products of
outcome residuals and test functions, which are used only in \cref{sec:sources}, but which we define
here for notational convenience:
\begin{equation}\label{eq:classes}
 \Dc=\{Dh:h\in\Hc\},\qquad \F,\qquad \Mc=\{(Y-h(X))f(Z)/2:h\in\Hc,\ f\in\F\}.
\end{equation}
Each class is star-convex and has uniform envelope one; for $\Dc$ this holds because $\Hc$ is
star-convex by \cref{ass:statistical} and $D$ is a linear operator, and for $\Mc$ it follows by scaling
$f$, regardless of the value of $Y$. We use two measures of complexity. The first is the Rademacher
complexity of the whole class, which controls the uniform deviations of empirical averages over the whole class. The second is the critical radius
\citep{bartlett2005}, which measures the complexity of the class locally, around functions of small
norm. We need the critical radius only for the gradient class, which will allow us to derive faster rates against more relaxed benchmarks than the identified gradient energy.

\Needspace{6\baselineskip}
\begin{definition}[Global complexities and critical radius]\label{def:radii}\label{def:global-radii}
Let $n\ge2$ and $0<\eta<1$, and set
\[
 N_n=2+\left\lceil\tfrac12\log_2 n\right\rceil,\qquad
 \ell_n=\log(20N_n/\eta).
\]
For each $\mathcal B\in\{\Dc,\F,\Mc\}$, let
$\mathfrak R_n(\mathcal B):=\mathfrak R_n(\mathcal B,\infty)$ be the Rademacher complexity of the whole
class, with the vector interpretation for $\Dc$, and let
\[
 r_{\mathcal B}:=\mathfrak R_n(\mathcal B)+\sqrt{\ell_n/n}
\]
be its confidence-adjusted global complexity. For the gradient class, assume moreover that there is a
deterministic $\delta_{\Dc}\in[n^{-1/2},1]$ such that
\[
 \mathfrak R_n^{\text{vec}}(\Dc,\delta_{\Dc})\le\delta_{\Dc}^2 .
\]
The smallest such solution is the critical radius of the class, and
$\rho_{\Dc}:=\delta_{\Dc}+\sqrt{\ell_n/n}$ is its confidence-adjusted critical radius. The critical radius
$\delta_{\mathcal B}$ and its confidence-adjusted version $\rho_{\mathcal B}$ are defined in the same
way for any other star-convex class $\mathcal B$.
\end{definition}

Both notions have been well-studied in the literature for various hypothesis spaces
\citep{wainwright}. The global complexity is never larger than the critical radius,
$r_{\mathcal B}\le\rho_{\mathcal B}$: when the class is star-convex and has envelope one, every element of the class, scaled by $\delta_{\mathcal B}$, belongs to the class and has norm at most
$\delta_{\mathcal B}$, which gives $\mathfrak R_n(\mathcal B)\le\delta_{\mathcal B}$. The two quantities
can, however, be of different orders. For a class contained in a $k$-dimensional linear space both admit upper bounds of order $\sqrt{k/n}$, so that $\rho_{\mathcal B}^2$ has the parametric order $(k+\ell_n)/n$. On the other hand, for
the convex hull of $p$ bounded functions the global complexity is bounded at order $\sqrt{\log(2p)/n}$, while the
critical radius can in general only be bounded by the square root of this quantity (see
\cref{sec:linear-improvement}). For this reason, our bounds use the global complexities for all the error
terms that are linear in the complexity of a class, and use the critical radius $\rho_{\Dc}$ only
where it enters squared.

\subsection{Approximation error and optimization error}

The two remaining sources of error are approximation and optimization. First, because the hypothesis class may not contain $g$, the tolerance must account 
for how far the class is from $g$. We measure this misspecification as the smallest graph-norm \eqref{eqn:graph_norm} distance between a hypothesis in the class to $g$. 

\begin{definition}[Graph-norm approximation error and comparator]\label{def:graph-approximation}
We assume that there exists a hypothesis $h^\circ\in \Hc$ such that $\|h^\circ - g\|_{\V}\leq \epsilon$. We call $\epsilon$ the approximation error of hypothesis $\Hc$ and $h^\circ$ the comparator.
\end{definition}

The graph norm controls both function value and gradient error directly, giving 
$\norm{D(h^\circ-g)}_2 \leq \epsilon$ and $\norm{h^\circ-g}_2 \leq \epsilon$. 
By conditional Jensen's inequality, this also gives $\norm{T(h^\circ-g)}_2 \leq \epsilon$, 
which bounds the moment violation of the comparator. Graph-norm approximation error is a stronger requirement than function-value error alone, but it avoids imposing a separate condition to convert function-value errors into gradient errors, which might not hold (see e.g. \cref{prop:value-misspecification}).

We also allow for an optimization error.
Numerical solvers typically 
return approximately optimal solutions. We measure the degree of suboptimality by a variational gap, which captures the degree of violation of the first-order optimality condition of the empirical optimization program.

\begin{definition}[Variational gap] 
    For a feasible estimate $\widehat h$, define
\begin{equation}\label{eq:gap}
 \operatorname{gap}_n(\widehat h)=
 \sup_{h\in\mathcal C_n}\Pn\left[
 D\widehat h^{\mathsf T}(D\widehat h-Dh)\right].
\end{equation}
Let $\varepsilon_{\rm opt}\ge0$ denote the optimization error, 
in the sense that the estimate is feasible and has variational 
gap at most $\varepsilon_{\rm opt}$: $\widehat h\in\mathcal C_n$ 
and $\operatorname{gap}_n(\widehat h)\le\varepsilon_{\rm opt}$. 
\end{definition}

When $\mathcal C_n$ is nonempty, the next 
lemma shows that any $\varepsilon_{\rm opt}>0$ is achievable, even when no exact 
minimizer exists in $\Hc$, which justifies the use of the variational gap. This result follows by the convexity of the hypothesis class $\Hc$. The empirical problem sees a hypothesis only through its values and gradients at the $n$ sample points, even 
when $\Hc$ and $\F$ are infinite. These evaluations 
form a bounded convex set in finite dimensions, whose 
closure contains an minimizer, so hypotheses 
with arbitrarily small variational gap always exist by the first-order conditions of the resulting convex optimization problem (proof is deferred to \cref{app:empirical}). 

\Needspace{6\baselineskip}
\begin{lemma}[Approximate empirical optimality]
\label{lem:gap-existence}
Under \cref{ass:statistical}, for every fixed sample on which the envelopes in \eqref{eq:boundedness} hold, with $\mathcal C_n\ne\varnothing$, and every $\varepsilon_{\rm opt}>0$, there exists
$\widehat h\in\mathcal C_n$ with gap at most $\varepsilon_{\rm opt}$.
If the minimum is attained by $\hat{h}$, then $\varepsilon_{\rm opt}=0$.
\end{lemma}

\subsection{Finite-sample improvement guarantee}

We now combine the quantities defined above to construct $\tau_n$ 
and the associated improvement lower bound. 
The empirical argument has the same structure as our population improvement argument, in that it follows from the first-order conditions of the empirical optimization problem, accounting for the sampling, approximation, and optimization errors at various steps. 

First, $g$ may not lie in $\Hc$, so 
$h^\circ \in \Hc$ serves as the comparator (\cref{def:graph-approximation}), and it must pass the empirical tests to be empirically feasible, i.e. in $\Cc_n$. 
We need to choose $\tau_n$ large enough to capture the sampling noise and the approximation error. The former is controlled by the 
global complexity $r_{\F}$ of the test class, and the latter is controlled by $\epsilon$.
The next lemma shows that $\tau_n=c_0r_{\F}+\epsilon$ covers both for a large enough
universal constant $c_0$. The proof is deferred to \Cref{lem:concentration}, which proves a stronger statement, controlling several other empirical averages used throughout the analysis.  

\begin{lemma}[Comparator feasibility]
\label{lem:concentration-short}
Under \cref{ass:statistical}, with the complexities of \cref{def:radii}, and the comparator of 
\cref{def:graph-approximation}, there exists a universal constant $c_0$ 
such that for $\tau_n = c_0 r_{\F} + \epsilon$, with probability at least $1-\eta$, $h^\circ\in\mathcal C_n$.
\end{lemma}

Second, on the event that $h^\circ$ is feasible, 
the first-order (variational gap) condition at $\widehat h$ compares $D\widehat h$ with 
$Dh^\circ$, but only in the empirical inner product and only up to the optimization error.
We can convert this empirical comparison with $Dh^\circ$ into the 
population policy derivative $I(D\widehat h)=\ip{Dg}{D\widehat h}$, accounting for the approximation error of $\Hc$ and the sampling error of empirical processes related to the gradient class. The next theorem collects the results into a lower bound on the first-order policy improvement, which limits the possible harm done by the estimate. The proof does not require recovering $a_\star$ or any further restrictions on the ill-posedness of the inverse problem.

\Needspace{6\baselineskip}
\begin{theorem}[Empirical policy improvement]\label{thm:safety}
Under \cref{ass:statistical}, with the complexities of \cref{def:radii}, and the comparator of 
\cref{def:graph-approximation}, use the tolerance
$\tau_n=c_0r_{\F}+\epsilon$ with $c_0$ from \cref{lem:concentration-short}. Let $\widehat h$ be a measurable estimate
with $\widehat h\in\mathcal C_n$ and
$\operatorname{gap}_n(\widehat h)\le\varepsilon_{\rm opt}$ whenever
$\mathcal C_n$ is nonempty. 
With probability at least $1-\eta$, the following bounds hold
simultaneously, where $e_n:=\Pn\norm{D\widehat h}^2$. The improvement is lower bounded relative to the full gradient energy of the
estimate,
\begin{align}
 I(D\widehat h)\ge~&\mathcal E(\widehat h)
 -Cr_{\Dc}-\epsilon\sqrt{\mathcal E(\widehat h)}-\varepsilon_{\rm opt},
 \label{eq:safety-global}\\
 I(D\widehat h)\ge~&e_n
 -Cr_{\Dc}-\epsilon-\varepsilon_{\rm opt},
 \label{eq:safety-global-emp}
\end{align}
and relative to half of the gradient energy of the estimate, 
\begin{align}
 I(D\widehat h)
 &\ge\tfrac12\mathcal E(\widehat h)
        -C(\rho_{\Dc}^2+\epsilon^2)-\varepsilon_{\rm opt},
       \label{eq:safety-pop}\\
 I(D\widehat h)
 &\ge\tfrac12 e_n
        -C(\rho_{\Dc}^2+\epsilon^2)-\varepsilon_{\rm opt}.
       \label{eq:safety-emp}
\end{align}
If $\varepsilon_{\rm opt}=O(\rho_{\Dc}^2)$,
negative improvement is at most $O(\delta_{\Dc}^2+\ell_n/n+\epsilon^2)$.
\end{theorem}

The proof is deferred to \cref{sec:proof-safety}. %
The two pairs of bounds are complementary. The first pair loses nothing in the gradient energy of the
estimate, at the price of an error that is linear in the global complexity of the gradient class. The
second pair gives up half of the gradient energy, and in exchange its error is quadratic in the critical
radius, limiting the possible finite sample harm at a faster rate. The proof also yields the localized bound
$I(D\widehat h)\ge\mathcal E(\widehat h)-C\rho_{\Dc}(\sqrt{\mathcal E(\widehat h)}+\rho_{\Dc})
-\epsilon\sqrt{\mathcal E(\widehat h)}-\varepsilon_{\rm opt}$ (see \eqref{eq:safety-unabsorbed} in Appendix~\ref{sec:proof-safety}), from which \eqref{eq:safety-pop} is derived. It improves on \eqref{eq:safety-global} only in
the intermediate regime where the gradient energy of the estimate is itself small (i.e., $\rho_{\Dc}\sqrt{\mathcal E(\widehat h)}+\rho_{\Dc}^2 < r_{\Dc}$). The lower bounds
\eqref{eq:safety-global-emp} and \eqref{eq:safety-emp}, stated with the empirical gradient energy $\Pn\norm{D\widehat h}^2$,
give a way to assess improvement from the
fitted direction: given a valid complexity bound and approximation error bound, a positive
right-hand side certifies positive population improvement. With fewer
data or a smaller fitted gradient, that right-hand side can be negative.

\begin{remark}[Gradient-regularized minimax estimation]\label{rem:minimax}
Instead of imposing the moments as constraints, one can 
penalize their violation in a joint loss directly:
\[
 \widehat h\in\argmin_{h\in\Hc}\;\sup_{f\in\F}\Pn\bigl[(Y-h(X))f(Z)-\tfrac12f(Z)^2\bigr]+\tfrac\lambda2\Pn\norm{Dh(X)}^2.
\]
Intuitively, the regularization level $\lambda > 0$ plays the role of the tolerance. 
This is a gradient-regularized version of the minimax estimator of \citet{dikkala2020} and can be analyzed in a similar way.
\end{remark}

\begin{remark}[Gradient-regularized DeepIV]\label{rem:deepiv}
We can also consider a two-stage alternative that first estimates the conditional law of 
$X$ given $Z$ on an independent sample, giving an estimated operator 
$\widehat Th(z)=\int h(x)\,\widehat p(dx\mid z)$, as in DeepIV 
\citep{hartford2017}. It then minimizes the squared fit to the 
outcome with a gradient-energy penalty:
\[
 \widehat h\in\argmin_{h\in\Hc}\;\tfrac12\Pn\bigl[(Y-\widehat Th(Z))^2\bigr]+\tfrac\lambda2\Pn\norm{Dh(X)}^2.
\]
This is a gradient-regularized version of DeepIV, and can be analyzed in a similar way as the Tikhonov regularized version studied in \citep{li2024}.
\end{remark}

\section{Approaching the Identified Gradient Energy}
\label{sec:sources}

In this section, we want to derive an improvement lower bound in terms of $\kappa$, that is,
we want to replace $\mathcal E(\widehat h)$ with $\kappa$ in the lower bound described in \cref{thm:safety}. We begin by considering how a candidate response function $\hat{h}$ with gradient energy lower than $\kappa$
can potentially be selected by the minimum-gradient-energy constrained estimator. Because the empirical procedure
only requires the moment condition to hold approximately, a candidate that does not satisfy the population
moment condition can still pass the empirical moment test. Specifically, the empirical moment test
relies on the test functions to approximate the conditional moment and uses only an upper bound $\tau_n$, so it is not tight. To compare $\mathcal E(\widehat h)$ with $\kappa$, we first define an intermediate population quantity, the \emph{relaxed identified gradient energy},
which we prove is a lower bound on the estimated gradient energy without any extra assumptions other than the ones presented in Section~\ref{sec:statistics}.

The main result of this section then requires connecting the relaxed identified gradient energy to $\kappa$. To achieve this we need to understand how the value of the gradient energy minimization program changes with the slackness of the constraints. The latter requires making a regularity assumption on the minimum gradient energy direction, i.e., that it admits a ``dual'' Riesz representer, which is a function that represents the minimum gradient energy direction in the instrument space (whose existence is a source condition assumption on the minimum gradient energy direction). Moreover, it requires that the test functions are rich enough to contain (or approximate) this dual representer.

\subsection{Comparison with a relaxed identified gradient energy}

First, we consider a set of response functions whose population moment condition holds approximately, in a manner analogous to that of $\mathcal{C}_n$.
Then, we can show that with high probability, $\widehat{h}$ will belong to this relaxed set. Taking the infimum of the gradient energy over this relaxed 
set then provides a lower bound on $\mathcal E(\widehat h)$. We define such gradient energy as the relaxed identified gradient energy.

\begin{definition}[Relaxed identified gradient energy]
\label{def:relaxed-margin}
For a hypothesis $h\in\Hc$, tolerance $t\ge0$ and test class $\F$,
define the relaxed population moment violation, $m_{\F}$, and the relaxed identified gradient energy,
$\kappa_{\Hc, \F}$,
\begin{align}
 m_{\F}(h)
 &:=\sup_{f\in \F}
       \E[(Y-h(X))f(Z)]
  =\sup_{f\in \F}\E[(g(X)-h(X))f(Z)],
       \label{eq:population-moment-error}\\
 \kappa_{\Hc,\F}(t)
 &:=\inf\{\mathcal E(h):h\in\Hc,\ m_{\F}(h)\le t\}.
       \label{eq:relaxed-margin}
\end{align}
The relaxed set is nonempty whenever $t\ge\epsilon$, since
$m_{\F}(h^\circ)\le\epsilon$. For smaller $t$, which may yield an empty feasible set, we use the convention $\inf\varnothing=+\infty$. 
\end{definition}

In \cref{sec:statistics}, the test tolerance for the estimation algorithm is chosen as $\tau_n = c_0 r_{\F} + \epsilon$. This level was derived by considering the sampling error of the empirical moments at a fixed hypothesis function (namely, $h^\circ$). To be able to provide a lower bound based on the relaxed identified gradient energy, we need to consider a tolerance $t$ in the relaxed energy definition that applies uniformly over all hypotheses in $\Hc$, so that the constraints are satisfied also by $\hat{h}$. For this reason, we will need to consider the relaxed identified gradient energy $\kappa_{\Hc, \F}(t_n)$ for a tolerance $t_n=\tau_n + Cr_\Mc$, where $r_\Mc$ is the confidence-adjusted global complexity of the class $\Mc$ of products of residuals and test functions defined in \eqref{eq:classes}.
To see why $\hat{h}$ satisfies the relaxed moment condition with high probability, note that for every $f\in\F$, we have
\[
 |\E[(Y-\widehat h(X))f(Z)]|
 \le\underbrace{|\Pn[(Y-\widehat h(X))f(Z)]|}_{\le\,\tau_n\ \text{(empirical feasibility)}}
   +\underbrace{|(\Pn-\E)[(Y-\widehat h(X))f(Z)]|}_{\le\,Cr_{\Mc}\ \text{(sampling error)}}
 \le\tau_n+Cr_{\Mc} .
\]
The first term is at most $\tau_n$ by empirical feasibility,
$\widehat h\in\mathcal C_n$. The second term is upper bounded by the uniform concentration of empirical processes over the function class $\Mc$. 
Taking the supremum over $f\in\F$ gives $m_{\F}(\widehat h) \leq \tau_n+Cr_{\Mc}$. This also shows
that the choice $t_n:=\tau_n+Cr_{\Mc}$ is sufficient for $\kappa_{\Hc,{\F}}(t_n)$
to act as a lower bound for $\mathcal{E}(\widehat{h})$, leading to the following improvement bound (proof in \cref{app:proofs-approaching}).

\Needspace{6\baselineskip}
\begin{theorem}[Policy improvement relative to the relaxed identified gradient energy]
\label{thm:relaxed-margin}
Under the hypotheses of \cref{thm:safety}, let
$t_n=\tau_n+Cr_{\Mc}$, where $C>0$ is a large enough universal constant. With probability at least $1-\eta$,
\begin{align}
 m_{\F}(\widehat h)&\le t_n,\qquad
 \mathcal E(\widehat h)\ge\kappa_{\Hc,{\F}}(t_n),
       \label{eq:relaxed-feasibility}\\
 I(D\widehat h)&\ge
       \kappa_{\Hc,\F}(t_n)
       -Cr_{\Dc}-\epsilon-\varepsilon_{\rm opt},
       \label{eq:relaxed-welfare-full}\\
 I(D\widehat h)&\ge
       \frac12\kappa_{\Hc,\F}(t_n)
       -C(\rho_{\Dc}^2+\epsilon^2)-\varepsilon_{\rm opt}.
       \label{eq:relaxed-welfare}
\end{align}
Consequently, with the nonnegative relaxation gap
\[
 \omega_n(t):=\max\bigl(\kappa-\kappa_{\Hc,\F}(t), 0\bigr),
\]
the same event gives
\begin{equation}\label{eq:relaxation-gap-bound}
 I(D\widehat h)\ge
       \kappa-\omega_n(t_n)
       -Cr_{\Dc}-\epsilon-\varepsilon_{\rm opt},\qquad
 I(D\widehat h)\ge
       \frac12\kappa-\frac12\omega_n(t_n)
       -C(\rho_{\Dc}^2+\epsilon^2)-\varepsilon_{\rm opt}.
\end{equation}
If the tests characterize exact population feasibility on $\Hc$,
meaning $m_{\F}(h)=0$ implies $Th=Tg$ for every $h\in\Hc$, then
\[
 \kappa_{\Hc,\F}(0)
 =\inf_{h\in\Hc\cap\Cc}\mathcal E(h)\ge\kappa.
\]
Equality holds if $\Hc\cap\Cc$ contains a feasible minimizing sequence
for $\kappa$, in particular if it contains an attained minimizer.
\end{theorem}

The gap $\omega_n(t)$ above is analogous to 
the modulus of continuity in prior works (see e.g., \citep{chen2012}): it measures how much gradient energy can be lost when 
the tested moments are violated by at most $t$. The following subsection bounds this gap under the assumption of the existence of a dual representer for the minimum gradient energy direction and  relates $\kappa_{\Hc,\F}(t_n)$ to $\kappa$.

\subsection{An exact dual representer condition}

\Cref{thm:relaxed-margin} leaves the quantity $\omega_n(t_n)$ unbounded, that is, the difference between $\kappa_{\Hc,\F}(t_n)$ and $\kappa$. 
We begin with the intuition that if the test functions approximate the conditional moment well, then the population conditional moment violation
should also be detected by the tests. In combination with a small moment tolerance, any candidate that passes the tests will also have gradient energy that cannot be too far below  $\kappa$. 

Expanding on the above, we first need to define the relevant notion of ``richness''. The key intuition is that not all conditional moments are important (binding) constraints for the minimum gradient energy program. So we only need the test functions to be rich enough so as to represent the ``binding constraints''. Note that $\kappa$ is simply the squared norm of $a_\star$,
so deriving a bound relative to $\kappa$ requires the moment test to detect violations along the direction of $a_\star$, which can be measured through the inner product between the estimate's 
gradient and $a_\star$. Hence, ideally we would like to test for $a_\star$ in the moment constraints. However, $a_\star$ is a function of the endogenous treatment and not the instrument. If this inner product can be ``represented'' by a dual function that is only a function of the instrument, then we can use this dual function as the test function and it would correspond to the single binding constraint that would suffice to control. We formalize the requirement as the \emph{dual representer condition}.

\Needspace{6\baselineskip}
\begin{definition}[Exact dual representer condition]\label{def:certificate}
Let $a_\star$ be the minimum-gradient-energy direction of \cref{thm:closure-convex}.
A dual representation for $a_\star$ is a function $q\in L^2(P_Z)$ such that
\begin{equation}\label{eq:source}
 \ip{a_\star}{Dh}=\E[q(Z)h(X)]\qquad(h\in \Hc).
\end{equation}
The condition is said to hold with
size $Q$ if the function $q$ exists and there is a constant $Q>0$ such that $q/Q\in\operatorname{conv}(\F)$, i.e., the convex hull of the test class.
\end{definition}

Since $g$ may not belong to $\Hc$, \Cref{eq:source} might not hold for $h=g$. In our analysis, we will need to relate the inner products of $a_\star$ and $g$ with the dual representation $\E[q(Z)g(X)]$.  For this reason, we define the \emph{anchor discrepancy} by comparing the dual moment with the first-order effect
of $a_\star$ under the true response:
\[
 \chi_q:=|\E[q(Z)g(X)]-\ip{a_\star}{Dg}|.
\]
As we will argue, the graph approximation property implies
$\chi_q\le(\norm q_2+\sqrt\kappa)\epsilon$.
When $I(a_\star)=\kappa$, in particular when $\V$ is linear,
this definition reduces to $|\E[q(Z)g(X)]-\kappa|$.

The dual representation extends the result \eqref{eq:canonical-functional}
in \cref{thm:closure-convex} to functions whose moments only hold approximately.
By \eqref{eq:source}, $q/Q\in\operatorname{conv}(\F)$ and the linearity of the moment violation in the test function, for
$h\in\Hc$,
\[
 \bigl|\ip{a_\star}{Dh}-\E[q(Z)g(X)]\bigr|
 =Q\,\bigl|\E[(q/Q)(Z)\,(h-g)(X)]\bigr|
 \le Q\,m_{\F}(h),
\]
Combining this bound with the definition of $\chi_q$ gives
\[
 |\ip{a_\star}{Dh}-I(a_\star)|
 \le Q\,m_{\F}(h)+\chi_q.
\]
Thus a small tested violation controls the difference between the
first-order effects of $a_\star$ under $h$ and under the true response,
up to the anchor discrepancy. The size $Q$ reflects how costly the
representation is. The bound above also suggests that if we include a scaled copy of the dual representation $q$ in the convex hull of $\F$, the moment violation gives a direct bound on $\ip{a_\star}{Dh}$ and the gap in gradient energy 
up to some anchor discrepancy. Combining this with \cref{thm:relaxed-margin} gives an improvement lower bound that
approaches $\kappa$ (proof in \cref{app:proofs-approaching}).

\Needspace{6\baselineskip}
\begin{theorem}[Policy improvement under an exact dual representation condition]
\label{thm:benchmark}
Under the hypotheses of \cref{thm:safety}, suppose an exact dual representer condition holds
with size $Q$ and anchor discrepancy $\chi_q$
(\cref{def:certificate}).
For every $t\ge0$,
\begin{equation}\label{eq:source-relaxed-margin}
 \kappa_{\Hc,{\F}}(t)\ge\kappa-2Qt-2\chi_q.
\end{equation}
With probability at least $1-\eta$, with $t_n=\tau_n+Cr_{\Mc}$ as in \cref{thm:relaxed-margin},
\begin{align}
 \mathcal E(\widehat h)&\ge
       \kappa-2Qt_n-2\chi_q,
       \label{eq:benchmark-energy}\\
 I(D\widehat h)&\ge\kappa
 -Cr_{\Dc}-2Qt_n
 -2\chi_q-\epsilon-\varepsilon_{\rm opt},
 \label{eq:benchmark-welfare}\\
 I(D\widehat h)&\ge\tfrac12\kappa
 -Qt_n-\chi_q-C(\rho_{\Dc}^2+\epsilon^2)-\varepsilon_{\rm opt}.
 \label{eq:benchmark-welfare-half}
\end{align}
Here $\tau_n=c_0r_{\F}+\epsilon$, and
$\chi_q\le(Q+\sqrt\kappa)\epsilon$. If the dual representation identity holds
at $g$, the anchor term vanishes. 
\end{theorem}

Equivalently, \eqref{eq:source-relaxed-margin} says
$\omega_n(t)\le2Qt+2\chi_q$, which answers the question we posed at the beginning of the section. For a fixed finite $Q$, the dual representation condition replaces $\mathcal E(\widehat h)$ by $\kappa$ at the cost of the
dual size $Q$, not of a slower rate. To see this, compare the bounds relative to the full energy:
\eqref{eq:safety-global} of \cref{thm:safety} gives
$I(D\widehat h)\ge\mathcal E(\widehat h)-Cr_{\Dc}-\epsilon-\varepsilon_{\rm opt}$, and
\eqref{eq:benchmark-welfare} gives
$I(D\widehat h)\ge\kappa-Cr_{\Dc}-2Q(c_0r_{\F}+Cr_{\Mc})-C(1+Q+\sqrt\kappa)\,\epsilon-\varepsilon_{\rm opt}$.
Both errors are linear in the global complexities of the classes involved, and the second differs
only through $Q$ and the anchor. For instance, when the three classes are contained in $k$-dimensional
linear spaces (\cref{lem:dimension}), both errors are of the root-$n$ order $\sqrt{(k+\ell_n)/n}$. The
faster order $(k+\ell_n)/n+\epsilon^2$ of the critical radius is available only relative to half of the
energy, as in \eqref{eq:safety-pop} and \eqref{eq:benchmark-welfare-half}, and in the latter only for
the error that comes from the gradient class.

\begin{remark}[Interpreting dual representer existence as a source condition]
The dual representer condition can be understood through the lens of
\emph{source conditions} commonly used to analyze regularized inverse
problems, including NPIV \citep{bennett2023,bennett2023source}.
Source conditions express the regularity of a target function relative
to a given linear operator; essentially imposing that the target function is primarily supported on the top part of the spectrum of the operator. A common form requires the target to lie
in the range of the operator's adjoint. To relate our condition to this literature, we need to specify what is the ``target'' function that we care about and which needs to lie in the range of the adjoint operator. 

To understand this, we need to consider which aspect (or summary statistics or linear functional) of a response function $h$ do we care about when we are interested in the identified gradient energy $\kappa$. The aspect that matters is the average derivative in direction $a_\star$, $\ip{a_\star}{Dh}$, which is the linear functional that appears in the first-order conditions of the minimum energy optimization problem.
For simplicity, consider the case where $\V$ is
a closed linear subspace of $L^2(P_X)$ and the functional
$\ell_\star(h):=\ip{a_\star}{Dh}$ is continuous in the $L^2(P_X)$ norm.
The Riesz representation theorem then gives a unique
$b_\star\in\V$ satisfying
\[
 \ip{a_\star}{Dh}
 =\ip{b_\star}{h}_{L^2(P_X)}
 =\E[b_\star(X)h(X)]
 \qquad(h\in\V).
\]
Let $T^*:L^2(P_Z)\to\V$ denote the adjoint of
$T:\V\to L^2(P_Z)$ under the usual $L^2$ inner products.\footnote{If $\V=L^2(P_X)$, then $T^*$ is simply the conditional expectation operator $T^* q=\E[q(Z)\mid X=\cdot]$.}
The Riesz representer of $h\mapsto\E[q(Z)h(X)]$ is $T^*q$.
By uniqueness of the representer, requiring \eqref{eq:source}
on all of $\V$ is therefore equivalent to requiring
\[
\exists q\in L^2(P_Z)\text{ such that }T^*q=b_\star.
\]
Thus if we wanted the dual representer to represent every $h\in \V$ and not just $h\in \Hc$, then the condition essentially asks that
$b_\star\in\operatorname{range}(T^*)
=\operatorname{range}((T^*T)^{1/2})$.
This is a $\beta=1$ source condition on $b_\star$ relative to
the conditional expectation operator $T$ (using the convention that $(u, A)$ satisfies the $\beta$ source condition if there exists $v$ such that
$u=(A^*A)^{\beta/2}v$ \citep{bennett2023source}).
\end{remark}

\begin{remark}[Interpreting dual representer as square-integrable Lagrange multiplier]
Existence of such a dual representer has appeared in the past literature on NPIV (\cite{bennett2026functionals}) to obtain point identification of a target linear functional of a partially identified NPIV response model. Though similar in nature, the dual representer condition here arises under a very different context: it is used to represent the first-order condition of an optimization problem and provide stability of its value to slackness in the constraints. Moreover, the fact that $a_\star$ is the minimizing direction, makes this assumption much more plausible than in the case of an arbitrary linear functional of the response function, which was considered in prior work (e.g., \cite{bennett2023source}). 

In particular, when the minimum gradient energy direction is attained at a feasible response function $h_\star$ (i.e., $a_\star=Dh_\star$) and $\V$ is a linear space, the dual representer is simply the Lagrange multiplier associated with the moment constraint in the minimum gradient energy problem, and the assumption requires that this multiplier be square integrable.
The Lagrangian of the minimum energy problem is
\[
\mathcal L(h,q)=\tfrac12\mathcal E(h)-\ip{q}{Th-Tg}_{L^2(P_Z)}
\]
$q(z)$ is the Lagrange multiplier associated with the moment constraint  $\E[Y-h(X)\mid Z=z]=0$. The first-order condition for a minimum is then:
every $v\in\V$
\[
 \begin{aligned}
 \left.\frac{d}{dt}\mathcal L(h_\star+tv,q)\right|_{t=0}
 &=\ip{Dh_\star}{Dv}-\ip{q}{Tv}_{L^2(P_Z)}=0
 \end{aligned}
\]
which is exactly the representer identity for $v\in \V$. Thus existence of a dual representer for all $v\in \V$ is equivalent to assuming that $q$ is a square integrable Lagrange multiplier for minimizing $\tfrac12\mathcal E(h)$ subject to $Th=Tg$.
\end{remark}

\begin{remark}[Existence of a dual representer for closed linear response spaces]
\label{rem:dual-existence}
For a closed linear response space $\V$, the fact that the inner product of $Dg$ with $a_\star$ is point identified follows from the first order condition presented in \cref{thm:closure-convex}, i.e., the fact that $\ip{a_\star}{Dg}=\ip{a_\star}{Dh}=\kappa$ for any $h\in \Cc$. This implies that 
$\ip{a_\star}{Dv}=0$ for every $v\in\ker(T|_\V)$. So if a scalar
representer $b_\star$ exists, 
then it satisfies $\ip{b_\star}{v}_{L^2(P_X)}=0$ for every $v\in\ker(T|_\V)$, which implies
$b_\star\in(\ker(T|_\V))^\perp=\cl(\operatorname{range}(T^*))$.

If the adjoint range is closed, this already gives
$b_\star=T^*q_\star$ for some $q_\star\in L^2(P_Z)$ and the dual representer property holds, for any $h\in \V$ (not just $h\in \Hc$).
The closed range property holds, for instance, for finite-dimensional response spaces or
finite-rank operators $T$. More generally, it holds when
$\norm{Th}_2\ge c\norm h_2$ for some $c>0$ on
$(\ker(T|_\V))^\perp$: e.g., the instrument may leave many directions
unidentified, but it controls all remaining directions uniformly. These conclusions follow from the Hilbert-space closed range
theorem \citep{bauschke2017}.
\end{remark}

\begin{remark}[Approximate dual source condition]
Note that for the case when $\V$ is a closed linear space, the argument in the previous remark implies that $b_\star$ is always in the closure of the range of $T^*$. This means that there exists a sequence of functions $(q_j)_{j\ge1}$ in $L^2(P_Z)$ such that $T^*q_j\to b_\star$ in $L^2(P_X)$. This implies that, if a Riesz representer $b_\star$ exists, then for any $h\in \Hc$, $\ip{a_\star}{Dh}=\ip{b_\star}{h}_{L^2(P_X)}=\lim_{j\to\infty}\E[q_j(Z)h(X)]$. Hence, even if the dual representer does not exist, we can still approximate it with a sequence of functions $(q_j)_{j\ge1}$ in $L^2(P_Z)$, which can be used to derive an approximate version of \cref{thm:benchmark}. Our analysis can be easily extended to the case when the dual representer does not exist, but can be approximated by a sequence of functions in $L^2(P_Z)$ in a quantitative manner, imposing a rate of convergence of the approximation error. 
\end{remark}

\section{Linear Hypothesis Classes}
\label{sec:sparse-linear}

We now instantiate the results of \cref{sec:statistics,sec:sources} for hypothesis classes that are
linear in a known, finite dictionary of features, with the conditional moments tested against a finite
list of test functions of the instrument. The dictionary is used only as an approximation device: we do
not assume that the true response $g$, or the response space $\V$, is linear in the dictionary, but
only that $g$ can be approximated by a linear combination of the features, with an approximation error
$\epsilon$ in the graph norm. In this setting, we show that the estimator is a finite-dimensional convex program with a quadratic objective and the tolerance needs to grow only logarithmically with the number of tests $k$ (i.e., vanishing error requires $\log(k)=o(n)$). We provide improvement guarantees for either the case when we impose $\ell_2$ or $\ell_1$ ball constraints on the parameter vector (\cref{sec:linear-improvement}). In the latter case the dependence on dictionary size $p$ is also logarithmic, allowing for the dictionary to grow exponentially with the sample size. Finally, we show that proving guarantees that approach the population identified gradient energy simplifies substantially: the relaxed
identified gradient energy is stable to the slackness of the constraints without any richness
assumption on the test functions, and the dual representer condition reduces to a finite system of
linear equations, with an easy-to-interpret sufficient condition (\cref{sec:linear-benchmarks}).

\paragraph{Dictionary and approximation error.}
Let $\phi=(\phi_1,\ldots,\phi_p)^{\mathsf T}$ be a known dictionary of functions of
the treatment with square-integrable values and gradients, and with linear span $\V_\phi$. Here $p$ is the dictionary size and need not equal the dimension of the instrument. For
$\theta\in\R^p$ we write
\[
 h_\theta(x)=\theta^{\mathsf T}\phi(x),\qquad
 J_\phi(x)=(\nabla\phi_1(x),\ldots,\nabla\phi_p(x)),\qquad
 Dh_\theta(x)=J_\phi(x)\theta ,
\]
so that the gradient energy is the quadratic form
\begin{equation}\label{eq:linear-gradient-norm}
 \mathcal E(h_\theta)=\theta^{\mathsf T}G\theta,\qquad G=\E[J_\phi(X)^{\mathsf T}J_\phi(X)] .
\end{equation}
The hypothesis class we use for estimation is $\Hc=\{h_\theta:\theta\in\Theta_\Hc\}\subseteq\V$, where $\Theta_\Hc$ is a convex set of
coefficients that contains a neighborhood of the origin. We allow for
mis-specification, i.e., we do not assume that $g$ lies in $\Hc$, or even in $\V_\phi$. Instead, we
only assume that $g$ can be approximated by functions in $\Hc$ in the graph norm, as in
\cref{def:graph-approximation}, which here means that there exists a comparator coefficient
$\theta^\circ\in\Theta_\Hc$ such that
\begin{equation}\label{eq:linear-approximation-error}
 \norm{h_{\theta^\circ}-g}_{\V}^2
 =\E\bigl[(g(X)-\phi(X)^{\mathsf T}\theta^\circ)^2\bigr]
   +\E\bigl[\norm{Dg(X)-J_\phi(X)\theta^\circ}^2\bigr]\ \le\ \epsilon^2 .
\end{equation}
We say that the dictionary is \emph{well specified} when $g=h_{\theta_0}$ for some $\theta_0\in\R^p$.

\paragraph{Tests and estimator.}
We test the conditional moments with a finite list $\psi=(\psi_1,\ldots,\psi_k)^{\mathsf T}$ of functions
of the instrument, fixed before seeing the data and normalized so that $|\psi_\ell|\le1$. The test class
is the symmetric star hull of this list,
\begin{equation}\label{eq:linear-test-class}
 \F=\bigl\{c\,\psi_\ell:\ 1\le\ell\le k,\ c\in[-1,1]\bigr\},
\end{equation}
which is the smallest symmetric and star-convex class that contains the $k$ tests, as required by
\cref{ass:statistical}. Let
\begin{equation}\label{eq:linear-test-moments}
\begin{aligned}
 B_{\ell j}&=\E[\psi_\ell(Z)\phi_j(X)],&\qquad b_\ell&=\E[\psi_\ell(Z)Y],\\
 \widehat B_{\ell j}&=\Pn[\psi_\ell(Z)\phi_j(X)],&\qquad \widehat b_\ell&=\Pn[\psi_\ell(Z)Y].
\end{aligned}
\end{equation}
Since the supremum of a tested moment over $c\in[-1,1]$ is its absolute value, the empirical and the
population moment violations of $h_\theta$ can be simply written as
\begin{equation}\label{eq:linear-moment-violation}
 \sup_{f\in\F}\Pn[(Y-h_\theta(X))f(Z)]=\norm{\widehat b-\widehat B\theta}_\infty,\qquad
 m_{\F}(h_\theta)=\norm{b-B\theta}_\infty ,
\end{equation}
and the minimum-gradient moment constraint estimator \eqref{eq:moment_constraint_estimator} is the
convex program with a quadratic objective
\begin{align}
 \widehat\theta&\in\argmin_{\theta\in\Theta_n}\tfrac12\theta^{\mathsf T}G_n\theta ,
 \qquad G_n=\Pn[J_\phi(X)^{\mathsf T}J_\phi(X)],
 \label{eq:linear-direct-estimator}\\
 \Theta_n&=\left\{\theta\in\Theta_\Hc:\
             \norm{\widehat b-\widehat B\theta}_\infty\le\tau_n\right\},
 \label{eq:linear-direct-set}
\end{align}
with $2k$ linear moment inequalities in addition to the coefficient constraint $\theta\in\Theta_\Hc$. We follow the convention that we return the zero direction if the constrained set is empty.

\subsection{Finite-sample improvement bounds}
\label{sec:linear-improvement}

We consider two choices of the coefficient set. The first is a Euclidean ($\ell_2$) ball, which is natural
when the dictionary is small relative to the sample size. The second is an $\ell_1$ ball, which allows
the dictionary to be much larger than the sample size. Such bounds on the $\ell_1$ norm of the coefficients are natural under sparsity restrictions \citep{gautier2011,belloni2012}: if an approximation uses at most $s$ features, then $\|\theta\|_1\le \sqrt{s}\|\theta\|_2$, and $\|\theta\|_1=O(s)$, assuming coordinate-wise bounded coefficients.

\begin{assumption}[Linear dictionary]\label{ass:linear-dictionary}
The outcome satisfies $|Y|\le1$, the tests satisfy $|\psi_\ell|\le1$, and $\Hc\subseteq\V$. Moreover,
one of the following two cases holds.
\begin{itemize}
 \item[(a)] \emph{$\ell_2$ ball with small dictionary:} the coefficient set is $\Theta_\Hc=\{\theta:\norm\theta_2\le R_2\}$,
 the features satisfy $\norm{\phi(x)}_2\le L_\phi$ and $\norm{J_\phi(x)}_{\rm op}\le L_\phi$ with
 $R_2L_\phi\le1$, and $p\le n$.
 \item[(b)] \emph{$\ell_1$ ball with high-dimensional dictionary:} the coefficient set is $\Theta_\Hc=\{\theta:\norm\theta_1\le R_1\}$,
 the features satisfy $|\phi_j(x)|\le1$ and $\norm{\nabla\phi_j(x)}\le1$ for every $j\le p$, and
 $2\log(2p)\le n$.
\end{itemize}
\end{assumption}

In case (a), the values and the gradients of the hypotheses are bounded by one, by the Cauchy--Schwarz
inequality, so that \cref{ass:statistical} holds. In case (b), H\"older's inequality bounds them by
$R_1$, with no dependence on $p$, and since the results of \cref{sec:statistics,sec:sources} are stated
for classes with envelope one, we apply them to the problem in which the outcome and the hypotheses are
divided by $\bar R:=\max\{R_1,1\}$; we let $\bar R:=1$ in case (a). It remains to compute the
complexities of \cref{def:radii}. The test class is the star hull of $k$ bounded functions, and the
maximal inequality for sub-Gaussian variables bounds its Rademacher complexity by
$\sqrt{2\log(2k)/n}$. In case (a), the gradient class is contained in a linear space of
dimension at most $p$, so that, by \cref{lem:dimension}, its Rademacher complexity is at most
$\sqrt{p/n}$ and $\delta_{\Dc}=\sqrt{p/n}$ is an admissible critical radius, while the class $\Mc$ is a
union of $k$ classes of dimension at most $p+1$, and \cref{lem:union-dimension} shows that taking this
union adds $\log k$ inside the square root of the global-complexity bound for $\Mc$. In case (b), H\"older's inequality reduces a supremum over the
$\ell_1$ ball of a Rademacher average that is linear in the coefficients to a maximum over finitely many
Rademacher averages of bounded functions, to which the maximal inequality applies again. Hence, with the
notation
\[
 \Delta_{n,m}:=\sqrt{\frac{2\log(2m)}{n}}+\sqrt{\frac{\ell_n}{n}}\qquad(m\ge1),
\]
the complexities of the classes of the rescaled problem satisfy
\begin{equation}\label{eq:linear-radii}
 r_{\F}\le\Delta_{n,k},\qquad r_{\Dc}\le C\,\bar r_{\Dc},\qquad r_{\Mc}\le C\,\bar r_{\Mc},
\end{equation}
where
\[
\begin{aligned}
 (\bar r_\Dc,\bar r_\Mc)
 &=\left(\sqrt{\frac{p+\ell_n}{n}},\sqrt{\frac{p+\log k+\ell_n}{n}}\right)
 &&\text{in case (a)},\\
 (\bar r_\Dc,\bar r_\Mc)&=(\Delta_{n,p},\Delta_{n,kp})
 &&\text{in case (b)}.
\end{aligned}
\]
and, in case (a), the critical radius satisfies $\rho_{\Dc}\le\sqrt2\,\bar r_{\Dc}$.

In case (b), the critical radius, however is not small: up to scaling, the gradient class is the convex
hull of the $2p$ functions $\pm\nabla\phi_j$, and without further conditions on the dictionary (e.g., restricted isometry property of the covariance matrix of the dictionary) its
critical radius is of the order $(\log(p)/n)^{1/4}$. The $\ell_1$ ball is
therefore a case where the global complexity is of smaller order than the critical radius: the bounds of
\cref{thm:safety,thm:relaxed-margin,thm:benchmark} whose errors are linear in the global complexities
attain the rate $\sqrt{\log(p)/n}$, whereas the bounds relative to half of the gradient energy, whose
errors are quadratic in the critical radius, give no faster rate. Instantiating \cref{thm:safety} with
these complexities gives the following guarantee, which we prove, together with the other results of
this section, in \cref{app:linear-proofs}.

\Needspace{6\baselineskip}
\begin{corollary}[Empirical policy improvement for a linear dictionary]
\label{cor:linear-improvement}
Under \cref{ass:linear-dictionary}, use the tolerance
\begin{equation}\label{eq:linear-tolerance}
 \tau_n=c_0\bar R\,\Delta_{n,k}+\epsilon
 =O\Bigl(\bar R\sqrt{\frac{\log k+\ell_n}{n}}+\epsilon\Bigr).
\end{equation}
Let $\widehat h=h_{\widehat\theta}$ for a minimizer $\widehat\theta$ of
\eqref{eq:linear-direct-estimator}, and let $e_n:=\widehat\theta^{\mathsf T}G_n\widehat\theta$ be its
empirical gradient energy. With probability at least $1-\eta$, the comparator is feasible,
$\theta^\circ\in\Theta_n$, and
\begin{align}
 I(D\widehat h)&\ge\mathcal E(\widehat h)-C\bar R^2\,\bar r_{\Dc}-\epsilon\sqrt{\mathcal E(\widehat h)},
 \label{eq:linear-full}\\
 I(D\widehat h)&\ge e_n-C\bar R^2\,\bar r_{\Dc}-\bar R\,\epsilon .
 \label{eq:linear-full-emp}
\end{align}
In case (a), moreover,
\begin{equation}\label{eq:linear-half-emp}
 I(D\widehat h)\ge\tfrac12e_n-C\bigl(\bar r_{\Dc}^2+\epsilon^2\bigr).
\end{equation}
\end{corollary}

In both cases, the number of tests enters the bounds only through the tolerance, and the dimension of
the dictionary only through the complexity of the gradient class. In case (a), the errors are of the
order $\sqrt{(p+\ell_n)/n}$ relative to the full gradient energy, and of the parametric order
$(p+\ell_n)/n+\epsilon^2$ relative to half of it. In case (b), the dimension enters only through
$\sqrt{\log p}$: if $g$ is approximated with error $\epsilon$ by a coefficient with at most $s$ nonzero
entries and Euclidean norm at most $r$, with $\sqrt s\,r\ge1$, then this coefficient lies in the
$\ell_1$ ball of radius $R_1=\sqrt s\,r$, since $\norm\theta_1\le\sqrt s\,\norm\theta_2$ for such a coefficient, and the errors
are of the order $s\,r^2\sqrt{(\log p+\ell_n)/n}$ and $\sqrt s\,r\,\epsilon$.

\subsection{Comparison with population benchmarks}
\label{sec:linear-benchmarks}

\Cref{cor:linear-improvement} bounds the improvement by the gradient energy of the estimate. As in
\cref{sec:sources}, we now compare this energy with the population identified gradient energy $\kappa$. By \cref{thm:relaxed-margin},
the gradient energy of the estimate is at least the relaxed identified gradient energy
$\kappa_{\Hc,\F}(t_n)$, at a tolerance $t_n$ that exceeds $\tau_n$ by the global complexity of the
class $\Mc$ (see \cref{cor:linear-benchmarks}). In the linear setting the relaxed identified gradient energy is the value of the
finite-dimensional program
\begin{equation}\label{eq:linear-relaxed-energy}
 \kappa_{\Hc,\F}(t)=\min_{\theta\in\Theta_\Hc}\ \theta^{\mathsf T}G\theta
 \qquad\text{subject to}\qquad \norm{b-B\theta}_\infty\le t .
\end{equation}
The question is how much gradient energy is lost when the tested moments are relaxed by $t$. We give
two answers. The first compares with a dictionary analogue of the identified gradient energy and
requires only feasibility of its tested moment equations; the second compares with $\kappa$ itself and requires the dual representer
condition of \cref{def:certificate}, which becomes a linear system.

\paragraph{The tested dictionary energy.}
Suppose that the $k$ tested moment equations $\E[\psi(Z)(Y-h_\theta(X))]=0$ admit a solution in $\R^p$,
i.e., that $B\theta=b$ is solvable, which holds, for instance, when the dictionary is well specified or
when $B$ has full row rank. We define the \emph{tested dictionary energy} as the smallest gradient
energy among the dictionary functions that satisfy these equations,
\begin{equation}\label{eq:tested-dictionary-energy}
 \kappa_{\phi,\psi}:=\min_{\theta\in\R^p}\ \theta^{\mathsf T}G\theta
 \qquad\text{subject to}\qquad B\theta=b .
\end{equation}
$\kappa_{\phi,\psi}$ is the minimum gradient energy under the given dictionary $\phi$ and tests $\psi$, where the coefficients are unrestricted ($\theta \in \R^p$). In \cref{sec:sources}, stability of the relaxed identified gradient energy to the slackness of the constraints
required a dual representer that is contained in the test class. Comparing $\kappa_{\Hc,\F}(t)$ to $\kappa_{\phi,\psi}$ requires no such assumption. This property follows from the fact that a quadratic program with linear constraints always admits a Lagrange multiplier $w\in\R^k$, and
the function $w^{\mathsf T}\psi$ lies in the span of the tests by construction. The existence of such a multiplier is the only thing that we need in order to relate $\kappa_{\Hc,\F}(t)$ to $\kappa_{\phi,\psi}$.

\Needspace{6\baselineskip}
\begin{proposition}[Stability of the tested dictionary energy]
\label{prop:tested-energy}
Suppose that $B\theta=b$ is solvable. Then the minimum in \eqref{eq:tested-dictionary-energy} is attained
at some $\theta_\psi$, and there exists $w\in\R^k$ with $G\theta_\psi=B^{\mathsf T}w$. Let $Q_\phi$ be the
smallest $\ell_1$ norm of such a multiplier. Then, for every $t\ge0$,
\begin{equation}\label{eq:tested-stability}
 \kappa_{\Hc,\F}(t)\ \ge\ \kappa_{\phi,\psi}-2\,Q_\phi\,t .
\end{equation}
\end{proposition}

When the tested moment equations have no solution, the same result holds with the equations relaxed to
any tolerance $t_0$ at which they are feasible, e.g., $t_0=\epsilon$, and with $t-t_0$ in place of $t$
(\cref{prop:tested-energy-anchor}).

\paragraph{The dual representer condition.}
Comparing with $\kappa$ requires a dual representer for the minimum-gradient-energy direction $a_\star$,
which is in the closure of gradients of response functions in the space $\V$ and is not necessarily the gradient of a dictionary function.\footnote{Note that \cref{rem:dual-existence} does not apply here, since it concerns a linear
response space $\V$ and a representation on all of $\V$, whereas here only the hypothesis class is
linear, and it merely approximates the response. It would have been applicable in the well-specified setting.} Nevertheless, the dual representer condition takes a simple form, because
both sides of Equation~\eqref{eq:source} are linear in the coefficients. Let
\begin{equation}\label{eq:linear-gamma-star}
 \gamma_\star:=\E\bigl[J_\phi(X)^{\mathsf T}a_\star(X)\bigr]\in\R^p ,\qquad
 \Gamma:=\E[\mu(Z)\mu(Z)^{\mathsf T}],\qquad \mu(Z):=\E[\phi(X)\mid Z],
\end{equation}
so that $\ip{a_\star}{Dh_\theta}=\gamma_\star^{\mathsf T}\theta$, and $\ker\Gamma$ consists of the
dictionary directions $v$ that the instrument cannot detect, $Th_v=0$.\footnote{Note that $(Th_v)(z)=\E[\phi(X)^\top\theta\mid Z=z]=\mu(z)^\top\theta$. Hence, $Th_v=0$ if and only if $\mu(Z)^\top\theta=0$ almost surely, which is equivalent to $\E[(\mu(Z)^\top\theta)^2]=\theta^\top\Gamma \theta=0$, i.e., $\theta\in\ker\Gamma$.} 

\Needspace{6\baselineskip}
\begin{lemma}[Dual representer condition for a linear dictionary]
\label{lem:linear-dual-dictionary}
The dual representer condition of \cref{def:certificate} holds on $\Hc$ with size $Q$ if and only if
there exists $w\in\R^k$ such that
\begin{equation}\label{eq:linear-dual-system}
 B^{\mathsf T}w=\gamma_\star,\qquad \norm w_1\le Q ,
\end{equation}
in which case $q=w^{\mathsf T}\psi$. The system $B^{\mathsf T}w=\gamma_\star$ has a solution if and only if
$\ip{a_\star}{Dh_v}=0$ for every direction $v$ that no test detects, $Bv=0$. In particular, it has a
solution when $\V$ is a linear space that contains $\V_\phi$ and $\ker B=\ker\Gamma$.
\end{lemma}

We always have $\ker\Gamma\subseteq\ker B$, since a direction that the instrument cannot detect is not
detected by any test. The condition $\ker B=\ker\Gamma$, or equivalently
$\operatorname{rank}B=\operatorname{rank}\Gamma$, asks that the tests detect every dictionary direction
that the instrument can detect, and it leaves the unidentified directions unrestricted. It requires at
least as many tests as the rank of $\Gamma$, but not as many as the dimension of the dictionary. When it
fails, there is a direction that changes the conditional moments without being detected by any test,
and moving along it can decrease the gradient energy below $\kappa$.

Combining \cref{prop:tested-energy,lem:linear-dual-dictionary} with
\cref{thm:relaxed-margin,thm:benchmark}, gives the following bounds for both coefficient sets.

\Needspace{6\baselineskip}
\begin{corollary}[Policy improvement relative to population identified gradient energy for a linear dictionary]
\label{cor:linear-benchmarks}
In the setting of \cref{cor:linear-improvement}, let $t_n=\tau_n+C\bar R\,\bar r_{\Mc}$. With probability
at least $1-\eta$, $\norm{b-B\widehat\theta}_\infty\le t_n$ and the following bounds hold. If $B\theta=b$
is solvable, then
\begin{equation}\label{eq:linear-tested}
 I(D\widehat h)\ge\kappa_{\phi,\psi}-2Q_\phi\,t_n-C\bar R^2\,\bar r_{\Dc}-\bar R\,\epsilon .
\end{equation}
If the system \eqref{eq:linear-dual-system} has a solution with $\norm w_1=Q$, then
\begin{equation}\label{eq:linear-kappa}
 I(D\widehat h)\ge\kappa-2Q\,t_n-2(Q+\sqrt\kappa)\,\epsilon-C\bar R^2\,\bar r_{\Dc}-\bar R\,\epsilon .
\end{equation}
\end{corollary}

The two bounds of \cref{cor:linear-benchmarks}, \eqref{eq:linear-tested} relative to $\kappa_{\phi,\psi}$
and \eqref{eq:linear-kappa} relative to $\kappa$, can be compared as follows. Suppose
the tests are rich relative to the dictionary,
$\ker B=\ker\Gamma$, and some dictionary function satisfies the tested moment restrictions exactly,
$B\theta=b$. The latter holds in particular when the dictionary is well specified or, we expect to hold when we have few tests compared to the dictionary size (under-specified setting). In that case the tested moment equations and the conditional moment restriction have the same solutions within the
dictionary, and therefore $\kappa_{\phi,\psi}\ge\kappa$ when $\V_\phi\subseteq\V$, with equality when
$\V=\V_\phi$ (\cref{sec:proof-linear-benchmarks}). The first bound then compares the estimate with a benchmark
that is no smaller than $\kappa$.

\section{Policy Improvement under Unknown Misspecification Error}
\label{sec:initial-direction}

The linear dictionary constraint estimator \eqref{eq:linear-direct-estimator} is simple to implement and exhibits an observable improvement certificate (\cref{cor:linear-improvement}). However, it requires a tolerance $\tau_n$ large enough to capture the misspecification error, which is unknown in practice.

Thus, we go back to our original question, which was simpler in the first place. \emph{Can we identify a direction with positive first-order policy improvement, no matter how small the improvement is?} We take a step back from the previous approach, which uses the gradient of a fitted function, and construct the direction directly from $Z$, $X$, $Y$, and the dictionary $\phi$ and test functions $\psi$, in a way that does not require choosing $\tau_n$. In exchange, we no longer compare the improvement to the (relaxed) identified gradient energy: we only ask that it is strictly positive.

Our analysis in this section utilizes the linear hypothesis space formulation and the notation of \cref{sec:sparse-linear}, i.e., hypotheses that are linear in a known, finite dictionary of features, with the conditional moments tested against a finite list of test functions of the instrument. We show that, in the population limit, if the best gradient approximator, namely the function in the span of the feature dictionary whose gradient is closest to the gradient of the response function, fits the test moments better than the everywhere-zero function, we can find such a strictly improving direction. This is only a sufficient condition, yet a very intuitive one. One way to ensure that this condition is met is to increase the size of the dictionary until its linear span is close enough to contain a good approximate solution with small test moments. The condition described above also requires a comparison against the test moments of the everywhere-zero function. We derive the single test function that maximizes this to ensure that our sufficient condition is not too conservative. Finally, we provide the empirical analogue at the end.

\subsection{The direction associated with most-violated test}

The intuition behind the construction is relatively simple: we look at what the constrained program of \cref{sec:sparse-linear} does when we push its tolerance to the point where the zero function is ``just excluded''. As a reminder, the population analogue of the program \eqref{eq:linear-direct-estimator}, with tolerance $t$ and without a constraint on the coefficients, is
\begin{equation}\label{eq:initial-population-path}
 \theta_t=\argmin_{\theta\in\R^p}\ \theta^{\mathsf T}G\theta
 \qquad\text{subject to}\qquad m_\F(h_\theta):=\norm{b-B\theta}_\infty\le t ,
\end{equation}
where $G$ is the gradient Gram matrix in \eqref{eq:linear-gradient-norm}, and $B$ and $b$ are the test moments in \eqref{eq:linear-test-moments}. The zero function is feasible if and only if $t\ge m_\F(0)=\norm b_\infty$, in which case the program returns the zero direction. Note that $m_\F(0)$ is the largest violation of a tested moment \emph{under the zero response}, and, unlike the misspecification error, it only depends on the moments of the outcome with the tests.

However, pushing the tolerance below $m_\F(0)$ is only meaningful if the solution of the program still has the improvement property. In \cref{sec:statistics}, we guaranteed this by choosing a tolerance large enough for a comparator, which approximates $g$ in the graph norm, to be feasible. Here in the population program \eqref{eq:initial-population-path}, we can use a different comparator, which only approximates the gradient of $g$. We assume $G\succ0$ for notational convenience, so that its inverse is well-defined,\footnote{If $G$ is not invertible, one may use $G^\dagger$ in place of $G^{-1}$. We omit this derivation.} and we define the \emph{best gradient approximator} as the function in the span $\V_\phi$ of the dictionary whose gradient is closest to the gradient of the response function:
\begin{equation}\label{eq:linear-projection-on-phi}
  h_\phi := h_{\theta_\phi} = \theta_\phi^{\mathsf T}\phi, \qquad \theta_\phi = \argmin_{\theta\in\R^p}\norm{Dg-J_\phi\theta}_2 =G^{-1}\E[J_\phi^{\mathsf T}Dg].
\end{equation}
This is an important distinction from the previous sections: we switch from approximation in the graph norm to approximation in the gradient seminorm alone, so that the values of $h_\phi$, and hence its tested moments, are not required to be close to those of $g$. The reason that gradient approximation suffices is that the error $Dg-Dh_\phi$ is orthogonal to the gradients of all the functions in $\V_\phi$, and therefore every direction of the form $J_\phi\theta$ has the same first-order improvement under $g$ as under $h_\phi$. Hence, whenever $h_\phi$ is feasible, i.e., $t\ge m_\F(h_\phi)$, the first-order optimality argument of \cref{thm:attained} applies with $h_\phi$ in the role of the true response, and the solution has an improvement of at least its own gradient energy, $I(J_\phi\theta_t)\ge\theta_t^{\mathsf T}G\theta_t$. 

Thus, every tolerance in the window $m_\F(h_\phi)\le t<m_\F(0)$ yields a strictly improving direction: the best gradient approximator is feasible, and the zero function is excluded, so that the gradient energy of the solution is strictly positive. We do not know where this window starts, since $m_\F(h_\phi)$ is unknown, but we know where it ends. Hence, if the window is non-empty, every tolerance just below $m_\F(0)$ falls in it. Such a tolerance does not yield the largest gradient energy, but it yields a non-zero solution that still has the improvement property. Moreover, we do not even need to choose how far below $m_\F(0)$ to go. For tolerances just below $m_\F(0)$, the only constraint that excludes zero is the one of the most-violated test $j_\star\in\operatorname*{arg\,max}_j|b_j|$, and the minimum-energy way of reducing the violation of this single moment is to move along the vector $v := \sgn(b_{j_\star})G^{-1}B_{j_\star}^{\mathsf T}$. When the most-violated test is unique, the solution $\theta_t$ in \eqref{eq:initial-population-path} is a positive multiple of this vector $v$ for every tolerance just below $m_\F(0)$ (\cref{prop:initial-path}). Since the first-order improvement is linear in the direction, we can simply use this vector itself, with no tolerance to choose. 

The next theorem shows that it is a strictly improving direction, as long as the window is non-empty. Note that while $h_\phi$ serves as an oracle comparator in the analysis, it is not needed to construct the proposed direction.

\Needspace{6\baselineskip}
\begin{theorem}[Selecting an improving direction without knowing the violation]
\label{thm:initial-single-test}
Suppose that $G\succ0$. Let $j_\star$ be a most-violated test under the zero response, i.e., $j_\star\in\operatorname*{arg\,max}_{1\le j\le k}|b_j|$ (choose any maximizing test under ties), and define the most-violated-test direction
\begin{equation} \label{eq:max-violated-test-direction}
 a_v = J_\phi v,\qquad
 v=\sgn(b_{j_\star})\,G^{-1}B_{j_\star}^{\mathsf T},
\end{equation}
where $B_j$ denotes the $j$-th row of $B$ and $\sgn(0)=0$. Then
\begin{equation}\label{eq:initial-single-bound}
 I(a_v)\ \ge\ \Delta_\phi:=m_\F(0)-m_\F(h_\phi)=\norm b_\infty-\norm{b-B\theta_\phi}_\infty .
\end{equation}
In particular, if $\Delta_\phi>0$, then $a_v$ is a non-zero direction with strictly positive first-order improvement.
\end{theorem}

Thus, a positive gap in test moments ($\Delta_\phi > 0$) is a sufficient condition for positive improvement. We refer to this condition as the \emph{positive-gap condition}; it states exactly that the window of tolerances described above is non-empty. The proof of the theorem (\cref{sec:proof-initial-single-test}) is a direct calculation based on the first-order optimality condition of the gradient approximator: the improvement of $a_v$ equals $\sgn(b_{j_\star})B_{j_\star}\theta_\phi$, i.e., the amount by which $h_\phi$ reduces the violation of the most-violated moment, and this amount is at least $\Delta_\phi$.

To see what this direction represents, we consider the simplest case where the dictionary and the tests are identity maps, $\phi(x) = x$ and $\psi(z) = z$, and $Z$ is centered and standardized. Then $J_\phi = G =I_d$, and
\begin{equation*}
\begin{aligned}
j_\star &= \operatorname*{arg\,max}_{1\leq j \leq k} |\operatorname{Cov}(Z_j,Y)|,\\
a_v&=\sgn\bigl(\E[Z_{j_\star}Y]\bigr)\,\E[XZ_{j_\star}]
=\sgn\bigl(\operatorname{Cov}(Z_{j_\star},Y)\bigr)\,\operatorname{Cov}(X,Z_{j_\star}).
\end{aligned}
\end{equation*}

The covariance $\operatorname{Cov}(Z, Y)$ is analogous to the intention-to-treat effect; it measures how much the exogenous treatment nudge affects the outcome. $j_\star$ finds the instrument with the strongest effect on the outcome. The sign identifies whether the instrument positively or negatively affects the outcome. Similarly, the covariance $\operatorname{Cov}(X,Z_{j_\star})$ measures relevance between the instrument and the treatment (a.k.a. the ``first stage'' in the classical two-stage-least-squares estimator for IV regression). Putting these together, $a_v$ is the direction of treatment relevance of the strongest instrument, with the sign corrected by the relationship between the strongest instrument and the outcome. The theorem allows us to conclude that under the positive-gap condition, perturbing $X$ in the direction of $a_v$ results in positive improvement.

\paragraph{Growing sieves.}
The positive-gap condition can eventually be satisfied as the linear dictionary grows and approximates the gradient of $g$ better. Consider a sequence of dictionaries $\phi_1,\phi_2,\ldots$ with growing linear spans, $\V_{\phi_1}\subseteq\V_{\phi_2}\subseteq\cdots\subseteq\V$, and a fixed list of tests. Let $h_{\phi,m}:=h_{\phi_m}$ be the best gradient approximator in the span of the dictionary $\phi_m$, let $a_m$ be the associated most-violated-test direction, and let
\[
 \delta_m:=\norm{Dg-Dh_{\phi,m}}_2=\min_{h\in\V_{\phi_m}}\norm{Dg-Dh}_2
\]
be the gradient approximation error of the $m$-th dictionary.
The following three conditions together are sufficient to guarantee positive improvement for every sufficiently large dictionary: 1) the upper bound on the gap, $m_\F(0)$, is positive; 2) the gradient approximation error $\delta_m$ goes to $0$; and 3) the tests are square-integrable and centered, and $\V$ satisfies a Poincar\'e inequality.
The last condition is used because the best gradient approximator only minimizes the gradient error, but to get a better approximation of $g$, we need the tested moments of the residual, determined by its function values, to go to $0$ as well. Centered tests are insensitive to the mean of the residual, and the Poincar\'e inequality bounds the variation of the residual around its mean by its gradient error.

\Needspace{6\baselineskip}
\begin{proposition}[Eventual improvement with fixed tests]
\label{prop:initial-growing-sieve}
Suppose that the dictionaries have growing spans, $\V_{\phi_1}\subseteq\V_{\phi_2}\subseteq\cdots\subseteq\V$, and positive definite gradient Gram matrices, that the tests are centered, $\E[\psi_j(Z)]=0$, and that \eqref{eq:poincare} holds with $C_P<\infty$. Let $L_\psi=\max_j\norm{\psi_j}_2$. Then
\begin{equation}\label{eq:initial-sieve-violation}
 m_\F(h_{\phi,m})\le L_\psi\sqrt{C_P}\,\delta_m\qquad(m\ge1),
\end{equation}
and therefore
\begin{equation}\label{eq:initial-sieve-improvement}
 |I(a_m)-m_\F(0)|\le m_\F(h_{\phi,m})
 \le L_\psi\sqrt{C_P}\,\delta_m.
\end{equation}
Consequently, if $m_\F(0)>0$ and $\delta_m\to0$, there is
an $m_0$ such that the positive-gap condition holds and $I(a_m)>0$ for every
$m\ge m_0$, and $I(a_m)\to m_\F(0)$.
\end{proposition}

Note that the dictionary size $m_0$, after which the positive-gap condition holds, is a property of the population: it depends on $m_\F(0)$, $L_\psi$, $C_P$ and on how fast the gradient approximation error decays, and not on the sample size (see \cref{sec:proof-initial-growing-sieve} for the proof of the proposition). On the other hand, if $m_\F(0)=0$, increasing the size of the dictionary alone cannot make the gap positive. This is the case when the selected tests detect no outcome signal and the most-violated test direction always returns $0$.

\paragraph{Centered reduced-form test.} A condition required by the growing-sieve result is that the limiting improvement $m_\F(0)$ be positive. We examine here which test maximizes $m_\F(0)$. We consider only centered ($\E[\psi] = 0$) and standardized ($\|\psi\|_2 \leq 1$) tests. The former condition is also used in the growing-sieve argument, so that the tests are insensitive to the mean of $Y$ and $h$. These do not matter to the gradient and hence should be ignored by the tests.
The latter fixes the scale among different tests so that they are at a comparable level and the maximum is well-defined. It turns out that there exists a single maximizing test. To see this, note that by the tower property, and since the test is centered, the tested outcome moment for any test function $\psi$ is
\begin{equation}\label{eq:initial-outcome-moment}
 \E[Y\psi(Z)]=\E[\E[Y\mid Z]\psi(Z)]=\E[r_0(Z)\psi(Z)]\le\norm{r_0}_2\norm{\psi}_2,
\end{equation}
where $r_0(z):=\E[Y\mid Z=z]-\E Y$ and the last step is the Cauchy--Schwarz inequality. The bound is attained by the standardized version of $r_0$, which is itself a centered test:
\begin{equation}\label{eq:initial-centered-reduced-form}
  \psi_\star(z) = r_0(z)/R, \qquad
  R=\norm{r_0}_2=\sqrt{\Var(\E[Y\mid Z])}.
\end{equation}
We call $r_0$ the centered reduced form, and write $T_0f=Tf-\E[f(X)]$, so that $T_0g=r_0$. The constant $R$ is the normalizing factor that measures the standard deviation of the conditional outcome. If $R=0$, then \eqref{eq:initial-outcome-moment} shows that $m_\F(0)=0$ for any class of centered test functions, and therefore the instrument is not sufficient to inform any non-zero improving direction.

Since the standardized reduced form attains the largest outcome moment, adding more centered standardized tests cannot increase $m_\F(0)$, and it can only increase the moment violation $m_\F(h_\phi)$ of the best gradient approximator. This implies that the best test class to use is also the singleton set containing the (standardized) centered reduced form.
The following proposition derives the most-violated-test direction and its improvement when the centered reduced form is included in the test class.

\Needspace{6\baselineskip}
\begin{proposition}[A normalized centered reduced-form test]
\label{prop:initial-reduced-form}
Suppose $G\succ0$, $R>0$, every test is centered and normalized to $\norm{\psi_j}_2\le1$,
and the list of tests contains $\psi_\star=r_0/R$. Then $\psi_\star$ maximizes $m_\F(0)$ with $m_\F(0)=R$. Choosing this
test among any ties gives
\begin{equation}\label{eq:initial-reduced-form-direction}
 v_r=\frac{1}{R}G^{-1}\E[\phi(X)r_0(Z)],
 \qquad a_r=J_\phi v_r,
 \qquad I(a_r)=\frac{\langle T_0h_\phi,r_0\rangle_{L^2(P_Z)}}{R}.
\end{equation}
The gap of \cref{thm:initial-single-test} is $\Delta_\phi=R-m_\F(h_\phi)$, and
\begin{equation}\label{eq:initial-reduced-form-bound}
 |I(a_r)-R|\le m_\F(h_\phi).
\end{equation}
In particular, $I(a_r)\ge R-m_\F(h_\phi)$, so that the positive-gap condition $m_\F(h_\phi)<R$ implies strict improvement. Moreover, the moment violation of the best gradient approximator is bounded by the differences in $g$ and $h_\phi$ as follows:
\begin{equation}\label{eq:initial-reduced-form-metrics}
 m_\F(h_\phi)
 \le\norm{T_0(g-h_\phi)}_2
 \le\norm{(g-h_\phi)-\E[g-h_\phi]}_2.
\end{equation}
If, in addition, $\V_\phi\subseteq\V$ and \eqref{eq:poincare} holds, then the right-hand side is at most $\sqrt{C_P}\norm{D(g-h_\phi)}_2$, so that a gradient approximation error smaller than $R/\sqrt{C_P}$ suffices for strict improvement.
\end{proposition}

\Cref{prop:initial-reduced-form}, which we prove in \cref{sec:proof-initial-reduced-form}, shows that the improvement of the most-violated-test direction obtained with the standardized centered reduced-form test is close to $R$, and the difference is bounded by the moment violation of the best gradient approximator.
 In particular, under the Poincar\'e inequality, if the linear dictionary model is well-specified (i.e., $Dg = J_\phi \theta$ for some $\theta$), the improvement is exactly $R$. For a misspecified but sufficiently flexible dictionary, such that the gradient approximation error is smaller than $R/\sqrt{C_P}$, the improvement remains strictly positive.

\subsection{Finite-sample improvement}

The empirical most-violated test direction replaces the population moment arrays $b$ and $B$ with the empirical averages defined in \eqref{eq:linear-test-moments}, and the gradient Gram matrix $G$ with an estimate $\widehat G$. A natural choice is $\widehat G=G_n$, defined in \eqref{eq:linear-direct-estimator}, but any measurable symmetric estimate of $G$ suffices, and it can use the same observations as those used to compute $\widehat B$ and $\widehat b$.
The empirical direction is then
\begin{equation}\label{eq:initial-sample-rule}
\widehat j\in\operatorname*{arg\,max}_{1\le j\le k}|\widehat b_j|, \qquad
 \widehat v=\sgn(\widehat b_{\widehat j})
                 \widehat G^{-1}\widehat B_{\widehat j}^{\mathsf T},
\end{equation}
whenever $\widehat G \succ0$ and $\widehat b\not=0$; otherwise, let $\widehat v=0$.

Three sampling errors enter the analysis: the error in the outcome moments $\widehat b$, which select the test; the error in the tested moments of the best gradient approximator, which determine the empirical gap; and the error in the Gram matrix, which turns the selected test into a direction. The next theorem shows that the improvement of this empirical direction is at least the moment violation gap, up to an additional sampling error adjustment, when the sampling errors are bounded with high probability.

\Needspace{6\baselineskip}
\begin{theorem}[Finite-sample improvement under a positive population gap]
\label{thm:initial-statistical}
Let $\gamma=\lambda_{\min}(G)>0$, and let $\widehat v$ be the empirical most-violated-test direction \eqref{eq:initial-sample-rule}. The estimate $\widehat G$ is symmetric.
Suppose, with probability at least $1-\eta$, the following bounds hold
jointly for deterministic nonnegative $\zeta_b,\zeta_\phi,\zeta_G$, with $\zeta_G<1$:
\begin{align}
 \norm{\widehat b-b}_\infty&\le \zeta_b,\nonumber\\
 \norm{(\widehat b-\widehat B\theta_\phi)
                      -(b-B\theta_\phi)}_\infty&\le \zeta_\phi,
                         \label{eq:initial-deviations}\\
 (1-\zeta_G)\,G\preceq\widehat G&\mathrel{\preceq}(1+\zeta_G)\,G.\nonumber
\end{align}
Write $L_{B,n}=\max_j\norm{\widehat B_j}_2$. On this event,
\begin{equation}\label{eq:initial-statistical-bound}
 I(a_{\widehat v})\ge
 \Delta_\phi-\zeta_b-\zeta_\phi
 -\frac{\zeta_G}{1-\zeta_G}\cdot\frac{\norm{Dh_\phi}_2\,L_{B,n}}{\sqrt\gamma}.
\end{equation}
Neither the population gap nor the deviation bounds are
inputs to the rule.
\end{theorem}

The first two conditions bound the sampling error of the tested moments of the zero function and of the best gradient approximator, and the third one states that $\widehat G$ estimates the Gram matrix up to a relative error $\zeta_G$. We state the last condition in relative terms, because this is the error that the matrix Chernoff inequality controls, with only a logarithmic dependence on the dimension of the dictionary. In the bound \eqref{eq:initial-statistical-bound}, which we prove in \cref{sec:proof-initial-statistical}, the quantity $L_{B,n}$ can be computed from the sample, $\norm{Dh_\phi}_2\le\norm{Dg}_2$, since $Dh_\phi$ is a projection of $Dg$, and the deviation bounds $\zeta_b,\zeta_\phi,\zeta_G$ follow from standard
concentration arguments, as in \cref{sec:sparse-linear}.
In particular, under \cref{ass:linear-dictionary}, case (a), suppose the tests are also bounded pointwise, $|\psi_j|\le1$, and the best gradient approximator lies in the coefficient ball, $\norm{\theta_\phi}_2\le R_2$, and use $\widehat G=G_n$. Then
$L_{B,n}\le L_\phi$ and $\norm{Dh_\phi}_2\le R_2L_\phi\le1$, where $L_\phi$ is the bound on the features and on their Jacobian in \cref{ass:linear-dictionary}, and, as long as the quantity $\zeta_G$ below is smaller than one, with probability at least $1-\eta$, \eqref{eq:initial-deviations} holds with
\[
 \zeta_b=\zeta_\phi=C\,r_\F\le C\Delta_{n,k},\qquad
 \zeta_G=L_\phi\sqrt{\frac{3\log(4p/\eta)}{\gamma\,n}} ,
\]
where $C$ is a universal constant, $r_\F$ is the global Rademacher complexity of the test class (\cref{def:radii}) and $\Delta_{n,k}$ is its bound in \eqref{eq:linear-radii} (see \cref{lem:initial-linear-deviations}).
If moreover $\zeta_G\le1/2$, then on the same event
\begin{equation}\label{eq:initial-linear-rates}
 I(a_{\widehat v})\ge\Delta_\phi-2C\Delta_{n,k}-\frac{4L_\phi^2}{\gamma}\sqrt{\frac{\log(4p/\eta)}{n}} .
\end{equation}
The two moment deviations are bounded by the global Rademacher complexity of the test class, which is of the order $\sqrt{(\log k+\ell_n)/n}$, so the number of tests enters only logarithmically. The deviation of the Gram matrix is bounded by a matrix Chernoff inequality (\cref{lem:initial-gram-chernoff}), and the condition $\zeta_G\le1/2$ requires $n\ge12(L_\phi^2/\gamma)\log(4p/\eta)$. Unlike the bounds of \cref{cor:linear-improvement}, the error
depends on the smallest eigenvalue $\gamma$ of the gradient Gram matrix, because the rule inverts
$\widehat G$. In special cases the Gram matrix is known and does not need to be estimated, e.g., $G=I_d$ for the identity dictionary $\phi(x)=x$. In such cases we can use $\widehat G=G$, so that $\zeta_G=0$ and the last term of the bound vanishes.

The dimension of the dictionary trades off two errors. As the dictionary grows, the gradient approximation error shrinks and the population gap $\Delta_\phi$ becomes larger, but the sampling error also increases: besides the factor $\log p$, the constant $L_\phi^2/\gamma$ grows at least linearly with the dimension of the dictionary, since $p\,\gamma\le\operatorname{tr}(G)\le d\,L_\phi^2$. However, the dictionary does not need to grow with the sample size. We only need a dictionary that is sufficiently large for the positive-gap condition to hold, and, by \cref{prop:initial-growing-sieve}, this happens at a dimension that is a constant, independent of $n$. For a dictionary of that dimension, the gap $\Delta_\phi$ is a fixed positive number, and the sampling error in \eqref{eq:initial-linear-rates} vanishes as the sample size grows. Hence, for a sufficiently large sample, the sampling error becomes smaller than the gap, and the empirical direction is strictly improving with probability at least $1-\eta$.

Finally, the centered reduced-form test is also unknown in practice and may require its own estimation procedure. Using a train-test split scheme, in which the test is estimated on one fold and the direction on another, makes the estimated test independent of the moments used for the direction, so the empirical improvement bound holds conditionally on the first fold, with the population gap of the estimated test.

\section*{Disclosure of AI Assistance}
\addcontentsline{toc}{section}{Disclosure of AI Assistance}

OpenAI Codex and Claude Code assisted with mathematical development, drafting,
editing, literature search, proof and algebra checks, and LaTeX
preparation. The authors remain responsible for the mathematical statements,
proofs, references, and final manuscript.
\clearpage
\appendix
\renewcommand{\theHsection}{appendix.\Alph{section}}
\crefalias{appendix}{section}
\crefalias{subappendix}{subsection}
\crefalias{subsubappendix}{subsubsection}
\tolerance=2000
\setlength{\emergencystretch}{3em}
\startcontents[appendices]
\section*{Appendix Contents}
\begingroup
\small
\printcontents[appendices]{}{-1}[2]{}
\endgroup

\appendixdivider{Proofs of Main Results}{%
\label{app:proofs-of-main}
}
\section{Proofs of Results in Section~\ref{sec:population}}

\subsection{Proof of Theorem~\ref{thm:closure-convex}}
\label{sec:proof-closure-convex}

\begin{proof}
Since $g\in\Cc$, the feasible set is nonempty. The linearity of $T$
and convexity of $\V$ imply that $\Cc$ is convex; linearity of $D$
then makes $\overline{D\Cc}$ a nonempty closed convex subset of
$L^2(P_X;\R^d)$. The Hilbert-space projection theorem
\citep{bauschke2017} therefore gives a unique minimum-norm vector
$a_\star$. For any $c\in\overline{D\Cc}$, convexity makes
$a_\star+t(c-a_\star)$ admissible for $0\le t\le1$. Optimality gives
\[
 0\le
 \frac{\|a_\star+t(c-a_\star)\|_2^2-\|a_\star\|_2^2}{2t}
 =\langle a_\star,c-a_\star\rangle
   +\frac t2\|c-a_\star\|_2^2
 \qquad(0<t\le1).
\]
Taking $t\downarrow0$ yields the variational inequality
\begin{equation}\label{eq:convex-projection-vi}
 \langle a_\star,c-a_\star\rangle\ge0
 \qquad(c\in\overline{D\Cc}).
\end{equation}

To relate this minimum to $\kappa$, note that
$\|a_\star\|_2^2\le\|Dh\|_2^2$ for every $h\in\Cc$.
Conversely, by the definition of the closure, we can find a sequence of feasible functions $(\widetilde h_j)$
such that $D\widetilde h_j\to a_\star$ in $L^2$. Continuity of the
squared norm gives
$\mathcal E(\widetilde h_j)\to\|a_\star\|_2^2$.
Taking infima in the first inequality and using this approximating
sequence proves $\kappa=\|a_\star\|_2^2$ without assuming that a
minimizing response exists. For any $c\in\overline{D\Cc}$, expansion
and \eqref{eq:convex-projection-vi} now give
\begin{align*}
 \|c\|_2^2
 &=\kappa+2\langle a_\star,c-a_\star\rangle
             +\|c-a_\star\|_2^2
 \ge\kappa+\|c-a_\star\|_2^2,\\
 \langle a_\star,c\rangle
 &=\kappa+\langle a_\star,c-a_\star\rangle
 \ge\kappa.
\end{align*}
Using $c=Dh$ proves \eqref{eq:canonical-functional}.
Using $c=Dg$ proves $I(a_\star)\ge\kappa$.
In particular, for any feasible minimizing sequence $(h_j)$,
\[
 0\le\|Dh_j-a_\star\|_2^2
 \le\mathcal E(h_j)-\kappa\longrightarrow0,
\]
which proves the stated strong convergence.

For the approximation bound, fix $h\in\Cc$ and write
$Dh=a_\star+u$ and $Dg=a_\star+b$. Applying the variational
inequality at $c=Dh$ shows that
$\langle a_\star,u\rangle\ge0$. Consequently,
\begin{align*}
 I(Dh)-I(a_\star)
 &=\langle a_\star+b,a_\star+u\rangle
       -\langle a_\star+b,a_\star\rangle\\
 &=\langle a_\star,u\rangle+\langle b,u\rangle
 \ge\langle b,u\rangle
 \ge-\|b\|_2\|u\|_2.
\end{align*}
The squared-norm inequality proved in the previous step, applied at $c=Dg$ and
$c=Dh$, gives
\[
 \|b\|_2^2\le\mathcal E(g)-\kappa,\qquad
 \|u\|_2^2\le\mathcal E(h)-\kappa.
\]
Substitution proves \eqref{eq:approx-bound-convex}.
Since $I(a_\star)\ge\kappa$, its right side is strictly positive
when $\kappa>0$, $\mathcal E(g)>\kappa$, and
$\mathcal E(h)-\kappa<\kappa^2/(\mathcal E(g)-\kappa)$.
Every feasible minimizing sequence eventually satisfies this condition.
If $\mathcal E(g)=\kappa$, then $b=0$ and the same calculation gives
$I(Dh)\ge I(a_\star)=\kappa$ for every feasible $h$.

Now assume \eqref{eq:convex-two-sided}. For any
$c\in\overline{D\Cc}$, applying \eqref{eq:convex-projection-vi}
to $a_\star-t_c(c-a_\star)$ gives
\[
 -t_c\langle a_\star,c-a_\star\rangle\ge0.
\]
Together with \eqref{eq:convex-projection-vi} and $t_c>0$,
this implies $\langle a_\star,c-a_\star\rangle=0$.
Thus the two squared-norm and inner-product expansions above hold
with equality. In particular,
\[
 I(a_\star)=\kappa,\qquad
 \|b\|_2^2=\mathcal E(g)-\kappa,\qquad
 \|u\|_2^2=\mathcal E(h)-\kappa.
\]
The improvement difference reduces to
\[
 I(Dh)-\kappa
 =\langle a_\star+b,a_\star+u\rangle-\kappa
 =\langle b,u\rangle.
\]
Cauchy--Schwarz therefore gives the asserted absolute approximation
bound. Relative algebraic interiority supplies the required reverse
step in every direction $c-a_\star$; relative norm interiority is
sufficient for the same reason. If $\overline{D\Cc}$ is affine,
the entire line through $a_\star$ and $c$ is feasible, so the condition
also holds.

Finally, if $\V$ is linear, let $K_\V=\V\cap\ker T$. Then $\Cc=g+K_\V$ and $\overline{D\Cc}=Dg+M$ for the closed linear subspace $M=\cl(DK_\V)$. Write $Dg=p+b$, where
$b=\Pi_M Dg$ and $p=Dg-\Pi_M Dg\in M^\perp$.
Then $Dg+M=p+M$, and for every $v\in M$,
$\|p+v\|_2^2=\|p\|_2^2+\|v\|_2^2$.
The unique minimum-norm element is therefore $p$, proving
$a_\star=Dg-\Pi_M Dg$.
\end{proof}

\subsection{Attainment of infimum}
\label{app:existence}

 The projection argument gave us the minimum-gradient-energy direction without requiring
a minimizing function. To obtain the function itself, we need to keep
a minimizing sequence from escaping the chosen function space. The
graph norm is useful because it tracks both function values and
gradients. A Poincar\'e inequality lets the gradient criterion control
the nonconstant part of a function, while the moment restrictions fix
its mean. Together with closedness of the constraints, these bounds
allow the direct method argument, in the spirit of \citep[Section~8.2]{evans2010}, to produce a feasible limit attaining the
minimum. Below, we use $H^1$ to denote the first-order Sobolev space (square-integrable functions with square-integrable weak gradients).

\Needspace{6\baselineskip}
\begin{theorem}[Existence under a Poincar\'e inequality]
\label{thm:existence}
Suppose $\V$ is convex and complete for the distance induced by the
graph norm \eqref{eqn:graph_norm}, and the Poincar\'e inequality
\eqref{eq:poincare} holds on $\V$ for some finite $C_P>0$.
Then the minimum squared gradient norm on $\Cc$ is attained. Alternatively, attainment holds without Poincar\'e if $\V$ is complete and $\Cc$ is bounded in graph norm.
If a Poincar\'e inequality also holds for some finite constant $C_p'$ on the differences $\V - \V = \{h_1 - h_2: h_1, h_2 \in \V \}$, the minimizing response is unique.
In particular, attainment and uniqueness hold for any closed convex
subset $\V\subseteq H^1(\Omega)$ containing $g$, when $X$ takes values in a bounded connected Lipschitz domain $\Omega\subseteq\mathbb R^d$ and
$0<c\le p_X(x)\le C<\infty$ almost everywhere on $\Omega$.
Here closedness is taken in the graph norm, which is equivalent to the
usual $H^1$ norm under these density bounds.
\end{theorem}

\begin{proof}
Every feasible $h$ has the same mean $m=\E g(X)$, since integrating
$Th=Tg$ preserves means. If $(h_j)$ is a minimizing sequence, then
\[
 \norm{h_j}_{\V}^2\le m^2+(C_P+1)\mathcal E(h_j)
\]
is bounded. Under the alternative graph-norm-bounded hypothesis, the same boundedness holds directly. To apply weak compactness without treating the convex class
as a linear space, let $\widehat{\mathbb V}$ be the Hilbert completion
of the ambient linear space $\mathbb V$ under the graph norm.
Completeness of $\V$ makes it a closed subset of this completion;
convexity then makes it weakly closed.
The bounds
\[
 \|Th\|_2\le\|h\|_2\le\|h\|_{\V},\qquad
 \|Dh\|_2\le\|h\|_{\V}
 \qquad(h\in\mathbb V)
\]
allow $T$ and $D$ to extend to bounded linear maps on
$\widehat{\mathbb V}$. Hence
$\Cc=\V\cap\{h:Th=Tg\}$ is also weakly closed in this Hilbert
space. A bounded minimizing sequence has a weakly convergent
subsequence whose limit belongs to $\Cc$. Weak lower semicontinuity
of the squared gradient norm makes that limit a minimizer.

For uniqueness under the additional difference inequality, let
$h_1,h_2$ be two minimizers. By \cref{lem:gradient-uniqueness},
$D(h_1-h_2)=0$. Their means coincide, so the difference inequality
gives
\[
 \|h_1-h_2\|_2^2
 =\Var(h_1(X)-h_2(X))
 \le C'_P\|D(h_1-h_2)\|_2^2=0.
\]
Thus the responses agree $P_X$-almost surely.

For the Sobolev sufficient conditions, bounds on $p_X$ make the graph
norm equivalent to the ordinary $H^1$ norm. Consequently, every closed
convex subclass is complete. For any $h\in H^1(\Omega)$, let $\bar h$ be the
Lebesgue average of $h$ on $\Omega$. The usual Poincar\'e--Wirtinger
inequality gives, for a domain constant $C_\Omega$,
\[
 \Var_{P_X}(h)\le C\int_\Omega|h-\bar h|^2
 \le CC_\Omega\int_\Omega\norm{\nabla h}^2
 \le (CC_\Omega/c)\mathcal E(h).
\]
The inequality holds on the entire ambient $H^1$ space, hence on
both $\V$ and $\V-\V$. This verifies all the sufficient conditions.
\end{proof}

\Needspace{6\baselineskip}
\begin{lemma}[Uniqueness of the minimizing gradient]
\label{lem:gradient-uniqueness}
Suppose $\V$ is convex. Then any two minimizers of the gradient energy
$\mathcal E$ over $\Cc$ have the same gradient $P_X$-almost surely. In
particular, if the minimum is attained at some $h_\star\in\Cc$, then
$Dh=Dh_\star=a_\star$ for every minimizer $h$.
\end{lemma}

\begin{proof}
Let $h_1,h_2\in\Cc$ be two minimizers, so that
$\mathcal E(h_1)=\mathcal E(h_2)=\kappa$. Since $\V$ is convex and the
constraint $Th=Tg$ is linear, the midpoint $(h_1+h_2)/2$ belongs to
$\Cc$. By the parallelogram law,
\[
 \mathcal E\bigl((h_1+h_2)/2\bigr)
 =\tfrac12\mathcal E(h_1)+\tfrac12\mathcal E(h_2)-\tfrac14\norm{Dh_1-Dh_2}_2^2
 =\kappa-\tfrac14\norm{Dh_1-Dh_2}_2^2 .
\]
The midpoint is feasible, so its gradient energy is at least $\kappa$.
Hence $\norm{Dh_1-Dh_2}_2=0$, that is, $Dh_1=Dh_2$ $P_X$-almost surely.
For the last claim, a minimizer $h_\star$ satisfies
$Dh_\star\in\overline{D\Cc}$ and $\norm{Dh_\star}_2^2=\kappa$, so
$Dh_\star$ is the minimum-norm element $a_\star$ of $\overline{D\Cc}$,
which is unique by \cref{thm:closure-convex}.
\end{proof}

\clearpage
\bookmarksetup{startatroot}
\section{Proofs of Results in Section~\ref{sec:statistics}}
\label{app:proofs-constrained}

\subsection{Empirical Process Lemmas}
\label{app:empirical}

This section contains the proofs of the localization lemmas stated in
the main text, together with the finite-dimensional complexity bound
used for parametric rates. The arguments follow the standard pattern
of symmetrization, contraction, bounded empirical-process
concentration, and localization
\citep{bartlett2005,maurer2016,boucheron2013} .

\begin{proof}[Proof of \cref{lem:gap-existence}]
Fix a sample on which the envelopes in \eqref{eq:boundedness} hold, and
suppose that $\Cc_n$ is nonempty. If $h_0,h_1\in\Cc_n$ and $s\in[0,1]$,
then $h_s=(1-s)h_0+sh_1$ belongs to $\Hc$ by convexity. For every
$f\in\F$,
\[
 \Pn[(Y-h_s(X))f(Z)]
 =(1-s)\Pn[(Y-h_0(X))f(Z)]
   +s\Pn[(Y-h_1(X))f(Z)]\le\tau_n.
\]
Thus $h_s\in\Cc_n$, proving that $\Cc_n$ is convex.

Let $u(h)=(h(X_i),Dh(X_i))_{i=1}^n\in\R^{n(d+1)}$, and let
$A=\{u(h):h\in\Cc_n\}$. The evaluation map is linear, so $A$ is
nonempty and convex. Moreover,
\[
 \|u(h)\|^2=\sum_{i=1}^n\bigl(|h(X_i)|^2+\|Dh(X_i)\|^2\bigr)\le2n.
\]
Consequently $\overline A$ is a nonempty compact convex set. Write
$u=(v_i,z_i)_{i=1}^n$, with $v_i\in\R$ and $z_i\in\R^d$, and define
\[
 j(u)=\frac1{2n}\sum_{i=1}^n\|z_i\|^2,\qquad
 \nabla j(u)=(0,z_i/n)_{i=1}^n.
\]
Then $j(u(h))=J_n(h)$ and $\nabla j$ is $1/n$-Lipschitz in Euclidean
norm. Continuity and compactness give a minimizer $u_\star\in\overline A$.
For any $v\in\overline A$ and $s\in(0,1]$, convexity and optimality imply
\[
 0\le\frac{j(u_\star+s(v-u_\star))-j(u_\star)}s.
\]
Letting $s\downarrow0$ gives
$\langle\nabla j(u_\star),u_\star-v\rangle\le0$.

By the definition of closure, choose $u_k\in A$ with $u_k\to u_\star$,
and choose $h_k\in\Cc_n$ with $u(h_k)=u_k$. Let
$L_A=\sup_{u,v\in A}\|u-v\|<\infty$. For every $v\in A$,
\begin{align*}
 \langle\nabla j(u_k),u_k-v\rangle
 &=\langle\nabla j(u_k)-\nabla j(u_\star),u_k-v\rangle
   +\langle\nabla j(u_\star),u_k-u_\star\rangle
   +\langle\nabla j(u_\star),u_\star-v\rangle\\
 &\le\left(L_A/n+\|\nabla j(u_\star)\|\right)\|u_k-u_\star\|.
\end{align*}
The bound is uniform in $v$ and tends to zero. On the other hand,
\[
 \sup_{v\in A}\langle\nabla j(u_k),u_k-v\rangle
 =\sup_{h\in\Cc_n}\Pn[Dh_k^{\mathsf T}(Dh_k-Dh)]
 =\operatorname{gap}_n(h_k)\ge0,
\]
where nonnegativity follows by taking $h=h_k$. Therefore the gaps
converge to zero. For every $\varepsilon_{\rm opt}>0$, a sufficiently
large $k$ supplies the asserted feasible estimate.

Finally, if $\widehat h$ attains the minimum on $\Cc_n$, then
$u(\widehat h)$ minimizes $j$ on $A$. Applying the same directional
derivative argument to the segment from $u(\widehat h)$ to any
$u(h)\in A$ gives
$\Pn[D\widehat h^{\mathsf T}(D\widehat h-Dh)]\le0$.
Taking the supremum and using $h=\widehat h$ proves that the gap is zero.
\end{proof}

We now prove a stronger version of \cref{lem:concentration-short}. Throughout, the localized and global complexities are those defined in \cref{sec:statistics} (see \eqref{eqn:complexity} and \cref{def:radii}), the comparator $h^\circ$ and the approximation error $\epsilon$ are those of \cref{def:graph-approximation}, and the tolerance is $\tau_n=c_0r_\F+\epsilon$, as in \cref{lem:concentration-short}.

\Needspace{6\baselineskip}
\begin{lemma}[Localized deviations (Full)]
\label{lem:concentration}
Under \cref{ass:statistical}, with \cref{def:radii}, and
 \cref{def:graph-approximation}, there is a universal $C$ and an event
of probability at least $1-\eta$ on which, simultaneously,
\begin{align}
 |(\Pn-\E)[(Dh^\circ)^{\mathsf T}Dh]|
 &\le C\rho_{\Dc}(\norm{Dh}_2+\rho_{\Dc}) &&(h\in\Hc),\label{eq:cross-conc}\\
 |(\Pn-\E)[\norm{Dh}^2]|
 &\le C\rho_{\Dc}(\norm{Dh}_2+\rho_{\Dc}) &&(h\in\Hc),\label{eq:energy-conc}\\
\sup_{h\in\Hc}|(\Pn-\E)[(Dh^\circ)^{\mathsf T}Dh]|
 &\le Cr_{\Dc},\label{eq:cross-conc-global}\\
 \sup_{h\in\Hc}|(\Pn-\E)[\norm{Dh}^2]|
 &\le Cr_{\Dc},\label{eq:energy-conc-global}\\
 \sup_{f\in\F}|(\Pn-\E)[(Y-h^\circ(X))f(Z)]|
 &\le Cr_{\F},\label{eq:noise-conc}\\
 \sup_{h\in\Hc,f\in\F}|(\Pn-\E)[(Y-h(X))f(Z)]|
 &\le Cr_{\Mc}.\label{eq:moment-conc}
\end{align}
Since $|\E[(Y-h^\circ)f]|\le\epsilon$ for every $f\in\F$,
we have $h^\circ\in\mathcal C_n$ for sufficiently large $c_0$.
\end{lemma}

\begin{proof}[Proof of \cref{lem:concentration}]
The proof below establishes the inequalities on one common event of \cref{lem:concentration-short}.
Let $\delta=\delta_{\Dc}$, and
let $\rho=\rho_{\Dc}=\delta+\sqrt{\ell_n/n}$. All classes and the
comparator $h^\circ$ are deterministic.

\textbf{Rescaling the localized class.}
For $r\ge\delta$ and $a\in\Dc$ with $\|a\|_2\le r$, star-convexity
implies that $(\delta/r)a\in\Dc$ and
$\|(\delta/r)a\|_2\le\delta$. Pointwise in the data and signs, it follows
that the supremum at radius $r$ is at most $r/\delta$ times the
supremum at radius $\delta$. Taking expectations gives
\begin{equation}\label{eq:rescale}
 \mathfrak R_n^{\rm vec}(\Dc,r)
 \le(r/\delta)\mathfrak R_n^{\rm vec}(\Dc,\delta)
 \le r\delta\qquad(r\ge\delta).
\end{equation}
This argument also applies when $r>1$. Since every gradient has
population norm at most one, taking $r=1$ gives
$\mathfrak R_n^{\rm vec}(\Dc)\le\delta$.

\textbf{Deviation on one gradient ball.}
For $z$ in the Euclidean unit ball, define
\[
 \psi_1(x,z)=Dh^\circ(x)^{\mathsf T}z,\qquad
 \psi_2(x,z)=\|z\|^2,\qquad
 S_{j,r}=\sup_{\substack{a\in\Dc\\\|a\|_2\le r}}
       |(\Pn-\E)\psi_j(X,a(X))|.
\]
Both losses vanish at zero, have absolute envelope one, and satisfy
$|\psi_j(x,z)|\le\|z\|$. Their Lipschitz constants on the unit ball
are $L_1=1$ and $L_2=2$: the first follows from Cauchy--Schwarz and
$\|Dh^\circ(x)\|\le1$, and the second follows from
\[
 |\|z\|^2-\|\widetilde z\|^2|
 =|\langle z+\widetilde z,z-\widetilde z\rangle|
 \le2\|z-\widetilde z\|.
\]
The first loss is already globally one-Lipschitz in $z$.
For the second, $z\mapsto\min\{\|z\|,1\}^2$ is a globally
two-Lipschitz extension of its restriction to the unit ball.

To see the symmetrization step explicitly, let $(X'_i)$ be an
independent sample. Conditional Jensen, random swapping of $X_i,X'_i$,
and the triangle inequality give
\begin{align*}
 \E S_{j,r}
 &\le\E\sup_{\|a\|_2\le r}
    \left|\frac1n\sum_i
       \{\psi_j(X_i,a(X_i))-\psi_j(X'_i,a(X'_i))\}\right|\\
 &\le2\E\sup_{\|a\|_2\le r}
    \left|\frac1n\sum_i\sigma_i\psi_j(X_i,a(X_i))\right|.
\end{align*}
Here and below the suprema over $a$ also require $a\in\Dc$.
Vector contraction \citep[Corollary 4]{maurer2016}, applied separately
to the loss and its negative, bounds the last expression by
$4\sqrt2 L_j\mathfrak R_n^{\rm vec}(\Dc,r)$.
The separate applications are legitimate because zero belongs to
the localized class, so each one-sided supremum is nonnegative and
the absolute supremum is at most their sum. Thus, for a universal
$C_1$ and every $r\ge\delta$,
$\E S_{j,r}\le C_1r\delta$.

For each $a$ on this ball, the variance of $\psi_j(X,a(X))$ is at
most $\E\|a(X)\|^2\le r^2$. The centered functions
$\pm\{\psi_j(X,a(X))-\E[\psi_j(X,a(X))]\}$ have envelope two.
Bousquet's bounded empirical-process inequality
\citep[Chapter 12]{boucheron2013}, with this envelope and variance,
therefore gives, for $u>0$, with probability at least $1-e^{-u}$,
\[
 S_{j,r}\le\E S_{j,r}
   +\sqrt{\frac{2u}{n}\{r^2+4\E S_{j,r}\}}+\frac{2u}{3n}.
\]
Since $\delta\le r$, we have $\E S_{j,r}\le C_1r^2$.
Consequently, with
$C_G=\max\{C_1,\sqrt{2(1+4C_1)},1\}$ and $u=\ell_n$,
\begin{equation}\label{eq:fixedball}
 S_{j,r}\le C_G\left(r\delta+r\sqrt{\ell_n/n}+\ell_n/n\right)
 \qquad(r\ge\delta)
\end{equation}
except on an event of probability at most $e^{-\ell_n}$.

\textbf{Simultaneous localization.}
If $\rho<1$, let $J=\lceil\log_2(1/\rho)\rceil$ and
$r_j=2^j\rho$, $0\le j\le J$. The largest ball contains $\Dc$,
because $r_J\ge1$, and
$J+1\le1+\lceil\tfrac12\log_2 n\rceil\le N_n$, since
$\rho\ge n^{-1/2}$. If $\rho\ge1$, use just the ball $r_0=\rho$,
which again contains $\Dc$. Apply \eqref{eq:fixedball} to both
losses on every one of these deterministic balls.
On the smallest ball its right side is at most $2C_G\rho^2$,
because $\delta+\sqrt{\ell_n/n}=\rho$ and $\ell_n/n\le\rho^2$.
If $s=\|a\|_2>\rho$, the first ball containing $a$ has radius
$r_j<2s$, so its right side is at most
$C_G(2s\rho+\rho^2)$. In either case,
\[
 |(\Pn-\E)\psi_j(X,a(X))|
 \le2C_G\rho(\|a\|_2+\rho)\qquad(a\in\Dc,\ j=1,2).
\]
Substituting $a=Dh$ proves \eqref{eq:cross-conc} and
\eqref{eq:energy-conc}, with total failure probability at most
$2N_ne^{-\ell_n}$ for these two bounds.

\textbf{The four global deviations.}
Consider in turn the two loss classes above, the class
$\{(Y-h^\circ(X))f(Z):f\in\F\}$, and
$\{(Y-h(X))f(Z):h\in\Hc,f\in\F\}$.
Their absolute envelopes are respectively $1,1,2,2$.
For any of these classes $\mathcal L$, let
$S_{\mathcal L}=\sup_{\ell\in\mathcal L}|(\Pn-\E)\ell|$.
Replacing one observation changes this supremum by at most $4/n$:
for each $\ell$, the empirical mean changes by at most $4/n$,
and taking an absolute value and a supremum cannot increase this bound.
The bounded differences inequality hence yields
\[
 S_{\mathcal L}\le\E S_{\mathcal L}+2\sqrt{2\ell_n/n}
\]
with failure probability at most $e^{-\ell_n}$.

The preceding symmetrization argument gives
$\E S_{\mathcal L}\le2\mathfrak R_n(\mathcal L)$.
For the first two classes, vector contraction, now without a norm
restriction, bounds this by
$C\mathfrak R_n^{\rm vec}(\Dc)$.
For the third, condition on the observations and let
$R_i^\circ=Y_i-h^\circ(X_i)$, so that $|R_i^\circ|\le2$.
Scalar contraction for the maps $t\mapsto R_i^\circ t$ gives
$\mathfrak R_n(\mathcal L)\le C\mathfrak R_n(\F)$.
For the fourth class, $\mathcal L=2\Mc$, so homogeneity gives
$\mathfrak R_n(\mathcal L)=2\mathfrak R_n(\Mc)$.
Together with the bounded differences bound and the definition
$r_{\mathcal B}=\mathfrak R_n(\mathcal B)+\sqrt{\ell_n/n}$,
these four expectation estimates prove
\eqref{eq:cross-conc-global}--\eqref{eq:moment-conc}.
No critical radius of $\F$ or $\Mc$ enters these global bounds.
Let $C$ be the maximum of the finitely many universal constants
obtained in the localized and global bounds.

There are at most $2N_n$ localized events and four global events.
Their union has failure probability at most
\[
 (2N_n+4)e^{-\ell_n}
 =\frac{2N_n+4}{20N_n}\eta<\eta.
\]
Thus all six inequalities hold on a single event $\Omega$ of
probability at least $1-\eta$.

\textbf{Feasibility of the comparator.}
For every $f\in\F$, the conditional moment restriction,
conditional expectation, and Cauchy--Schwarz give
\begin{align*}
 |\E[(Y-h^\circ(X))f(Z)]|
 &=|\E[(g(X)-h^\circ(X))f(Z)]|\\
 &=|\E[T(g-h^\circ)(Z)f(Z)]|\\
 &\le\|T(g-h^\circ)\|_2\|f\|_2
 \le\|g-h^\circ\|_2\le\epsilon .
\end{align*}
Here conditional Jensen gives the penultimate inequality and
$\|f\|_2\le1$. On $\Omega$, \eqref{eq:noise-conc} therefore implies
\[
 \sup_{f\in\F}|\Pn[(Y-h^\circ(X))f(Z)]|
 \le\epsilon+Cr_{\F}\le\epsilon+c_0r_{\F}=\tau_n
\]
when $c_0\ge C$. Since $h^\circ\in\Hc$, this is precisely the
empirical feasibility requirement.
\end{proof}

For a concrete way to verify the complexity assumption, consider a
class lying in a finite-dimensional linear span. Localization restricts
the coefficient vectors to a small population-norm ball, yielding the
radius used in \cref{sec:linear-improvement}. The same calculation works for
scalar losses and for the coordinate-wise Rademacher process associated
with vector gradients. As in \cref{sec:statistics}, $\mathfrak R_n(\mathcal B,r)$ and $\mathfrak R_n^{\rm vec}(\mathcal B,r)$ denote the localized Rademacher complexities of a scalar and a vector class at radius $r>0$ (see \eqref{eqn:complexity}).

\Needspace{6\baselineskip}
\begin{lemma}[Localized complexity of a finite-dimensional class]
\label{lem:dimension}
If a scalar class or a vector class is contained in a $k$-dimensional
linear subspace of its $L^2$ space, then its corresponding localized
Rademacher complexity is at most $r\sqrt{k/n}$.
We use fixed measurable versions: the span representations hold
outside one null set common to the class. This convention is automatic
for a dictionary of fixed functions and their linear combinations.
\end{lemma}

\begin{proof}
Let $m\le k$ be the dimension of the containing subspace. If $m=0$,
every function is zero outside the common null set, so every
Rademacher average is zero almost surely and the conclusion follows.
Suppose $m\ge1$, and choose an orthonormal basis
$e_1,\ldots,e_m$ in the relevant population $L^2$ inner product.
Each $b$ in the localized class has a representation
$b=\sum_{\ell=1}^m\theta_\ell e_\ell$, with
\[
 \|b\|_2^2=\sum_{\ell=1}^m\theta_\ell^2\le r^2.
\]
The fixed-version convention in the statement ensures that these
representations hold simultaneously at all sample points with
probability one.

In the scalar case, let
$\xi_\ell=n^{-1}\sum_i\sigma_i e_\ell(W_i)$.
In the vector case, with vector coordinates indexed by $c$, let
$\xi_\ell=n^{-1}\sum_{i,c}\sigma_{ic}e_{\ell,c}(X_i)$.
In either case the signed average of $b$ equals $\theta^{\mathsf T}\xi$.
Enlarging the coefficient set to the whole Euclidean ball gives
\[
 \sup_{\substack{b\in\mathcal B\\\|b\|_2\le r}}
       |\text{signed average of }b|
 \le\sup_{\|\theta\|\le r}|\theta^{\mathsf T}\xi|
 =r\|\xi\|.
\]
The last equality follows from Cauchy--Schwarz, with equality at
$\theta=r\xi/\|\xi\|$ when $\xi\ne0$; when $\xi=0$ both sides vanish.

Conditional on the sample, all products of two distinct signs have
expectation zero. Thus in the scalar case
\[
 \E_\sigma[\xi_\ell^2\mid W_1,\ldots,W_n]
   =\frac1{n^2}\sum_i e_\ell(W_i)^2,
\]
and in the vector case the corresponding expression is
$n^{-2}\sum_{i,c}e_{\ell,c}(X_i)^2$.
Taking expectation over the sample and using $\|e_\ell\|_2=1$
gives $\E\xi_\ell^2=1/n$ in both cases. Hence
$\E\|\xi\|^2=\sum_{\ell=1}^m\E\xi_\ell^2=m/n$.
Finally, Cauchy--Schwarz for expectations gives
\[
 \mathfrak R_n(\mathcal B,r)
 \le r\E\|\xi\|
 \le r\sqrt{\E\|\xi\|^2}
 =r\sqrt{m/n}\le r\sqrt{k/n},
\]
with $\mathfrak R_n^{\rm vec}$ in place of $\mathfrak R_n$ in the
vector case. No independence across the coordinates $\xi_\ell$ is needed.
\end{proof}

The next lemma extends this calculation to a finite union of finite-dimensional classes, such as the
symmetric star hull of a finite list of test functions, which is a union of one-dimensional classes.
Taking the union of $k$ classes costs only an additive $\log k$ in the squared radius.

\Needspace{6\baselineskip}
\begin{lemma}[Localized complexity of a finite union of finite-dimensional classes]
\label{lem:union-dimension}
Let $\mathcal B=\bigcup_{\ell\le k}\mathcal B_\ell$ be a class of scalar functions of $W$ with absolute
envelope one, where each $\mathcal B_\ell$ is contained in a linear subspace of $L^2$ of dimension at
most $m\ge1$. Use the same fixed-version convention as in
\cref{lem:dimension}. Then, for every $r>0$, with $s_n:=\sqrt{(m+\log k)/n}$,
\[
 \mathfrak R_n(\mathcal B,r)\le4\,r\,s_n+3\,s_n^2 .
\]
Consequently, the choice $\delta_{\mathcal B}=C_0s_n$ with $C_0=5$ satisfies
$\mathfrak R_n(\mathcal B,\delta_{\mathcal B})\le\delta_{\mathcal B}^2$, and it is an admissible radius in
\cref{def:radii} whenever $n\ge C_0^2(m+\log k)$.
\end{lemma}

\begin{proof}
It is enough to prove the upper bound after adding zero to each
$\mathcal B_\ell$. This preserves its envelope and containing
subspace and can only enlarge a supremum. We use this enlargement
so that all localized classes below are nonempty. For $\ell\le k$, let
\[
 S_\ell=\sup_{\substack{b\in\mathcal B_\ell\\\|b\|_2\le r}}
       \left|\frac1n\sum_i\sigma_i b(W_i)\right|,\qquad
 A=\max_{\ell\le k}\E S_\ell,\qquad M=\max_{\ell\le k}S_\ell.
\]
The complexity of the enlarged union is $\E M$, and
\cref{lem:dimension} gives $A\le r\sqrt{m/n}$.

Treat $(\sigma_i,W_i)$ as independent observations with their joint
product law. For each fixed $b$, the two functions
$(\sigma,w)\mapsto\pm\sigma b(w)$ are centered, have absolute
envelope one, and have variance
$\E[\sigma^2b(W)^2]=\|b\|_2^2\le r^2$.
Consequently Bousquet's inequality \citep[Chapter 12]{boucheron2013}
applies to the supremum defining $S_\ell$ and gives, for $u\ge0$,
\[
 \Pr\left\{S_\ell>\E S_\ell+
       \sqrt{\frac{2u(r^2+2\E S_\ell)}n}+\frac u{3n}\right\}
 \le e^{-u}.
\]
Use $u+\log k$ in place of $u$ and take a union bound over
$\ell\le k$. Since both terms involving $\E S_\ell$ are increasing
in that expectation, this proves
\[
 \Pr\left\{M>A+
       \sqrt{\frac{2(u+\log k)(r^2+2A)}n}
       +\frac{u+\log k}{3n}\right\}\le e^{-u}.
\]
Let $f(u)$ denote the deterministic threshold inside this probability,
and let $U$ have the standard exponential distribution.
The tail bound and the fact that $f$ is increasing imply
$\E M\le\E f(U)$. Explicitly, the tail integral gives
$\E M\le f(0)+\int_0^\infty e^{-u}f'(u)\,du
=\int_0^\infty e^{-u}f(u)\,du$; for $k=1$ the possible derivative
singularity at zero is integrable, and the same identity follows by
first integrating from a positive lower endpoint and taking its limit.
Let $L=1+\log k$. Since $\E U=1$,
$\E\sqrt U\le\sqrt{\E U}=1$, and
$(1+\sqrt{\log k})^2\le2L$, we obtain
\begin{align*}
 \E M
 &\le A+\sqrt{\frac{2(r^2+2A)}n}\,
                   \E\sqrt{U+\log k}+\frac L{3n}\\
 &\le A+2\sqrt{\frac{(r^2+2A)L}n}+\frac L{3n}\\
 &\le A+2r\sqrt{L/n}+2\sqrt{2AL/n}+\frac L{3n}\\
 &\le 2A+2r\sqrt{L/n}+\frac{7L}{3n}.
\end{align*}
The third line uses $\sqrt{x+y}\le\sqrt x+\sqrt y$, and the fourth
uses $2\sqrt{2AL/n}\le A+2L/n$.
Because $m\ge1$, we have $L\le m+\log k$, so that
$A\le rs_n$, $\sqrt{L/n}\le s_n$, and $L/n\le s_n^2$.
Substitution gives
$\mathfrak R_n(\mathcal B,r)\le\E M\le4rs_n+3s_n^2$,
as asserted.

For the fixed-point claim, set $\delta_{\mathcal B}=5s_n$. The bound
just proved gives
\[
 \mathfrak R_n(\mathcal B,\delta_{\mathcal B})
 \le(4\cdot5+3)s_n^2=23s_n^2
 \le25s_n^2=\delta_{\mathcal B}^2.
\]
Also $m+\log k\ge1$ implies $\delta_{\mathcal B}\ge n^{-1/2}$,
and $n\ge25(m+\log k)$ implies $\delta_{\mathcal B}\le1$.
These verify both the fixed-point inequality and the interval
requirement in \cref{def:radii}.
\end{proof}

\subsection{Proof of Theorem~\ref{thm:safety}}\label{sec:proof-safety}

\begin{proof}
Let $\Omega$ be the simultaneous concentration event in \cref{lem:concentration-short} and \cref{lem:concentration}, so that $\Pp(\Omega)\ge1-\eta$. We prove all the stated bounds on this event. By \cref{lem:concentration}, $h^\circ\in\mathcal C_n$ on $\Omega$, so $\mathcal C_n$ is nonempty. The assumptions on the estimator therefore ensure that $\widehat h\in\mathcal C_n$ and $\operatorname{gap}_n(\widehat h)\le\varepsilon_{\rm opt}$.

Let
\[
 e_n=\Pn\norm{D\widehat h}^2,\qquad
 c_n=\Pn[(Dh^\circ)^{\mathsf T}D\widehat h].
\]
Taking $h=h^\circ$ in the supremum defining the variational gap gives
$e_n-c_n\le\varepsilon_{\rm opt}$, or equivalently
\begin{equation}
 c_n\ge e_n-\varepsilon_{\rm opt}.
 \label{eq:empirical-foc}
\end{equation}
The graph approximation assumption gives
$\norm{D(g-h^\circ)}_2\le\epsilon$. Since $D$ is linear,
\begin{align*}
 I(D\widehat h)
 &=\ip{Dg}{D\widehat h}\\
 &=\ip{Dh^\circ}{D\widehat h}
   +\ip{D(g-h^\circ)}{D\widehat h}\\
 &\ge\ip{Dh^\circ}{D\widehat h}-\epsilon \sqrt{\mathcal E(\widehat h)},
\end{align*}
where the last line is Cauchy-Schwarz.

We first prove the bounds relative to the full gradient energy. By
\eqref{eq:cross-conc-global}, $\ip{Dh^\circ}{D\widehat h}\ge c_n-Cr_{\Dc}$, and combining this with
\eqref{eq:empirical-foc} and the last display gives
\[
 I(D\widehat h)\ge e_n-Cr_{\Dc}-\epsilon \sqrt{\mathcal E(\widehat h)}-\varepsilon_{\rm opt}.
\]
Since $\sqrt{\mathcal E(\widehat h)}\le1$ by the gradient envelope, this is \eqref{eq:safety-global-emp}. Using
$e_n\ge \mathcal E(\widehat h)-Cr_{\Dc}$ from \eqref{eq:energy-conc-global} and enlarging $C$ gives
\eqref{eq:safety-global}.

We next prove the bounds relative to half of the gradient energy, which use the localized
inequalities. Choose a universal constant $C_0\ge1$ large enough for both
\eqref{eq:cross-conc} and \eqref{eq:energy-conc}. On $\Omega$ these
bounds hold for every $h\in\Hc$, so they apply to the data-dependent
choice $\widehat h$ and give
\[
 \left|c_n-\ip{Dh^\circ}{D\widehat h}\right|
 \le C_0\rho_{\Dc}(\sqrt{\mathcal E(\widehat h)}+\rho_{\Dc}),\qquad
 |e_n-\mathcal E(\widehat h)|\le C_0\rho_{\Dc}(\sqrt{\mathcal E(\widehat h)}+\rho_{\Dc}).
\]
Combining the first inequality with \eqref{eq:empirical-foc} yields
\begin{equation}
 I(D\widehat h)\ge
 e_n-C_0\rho_{\Dc}(\sqrt{\mathcal E(\widehat h)}+\rho_{\Dc})-\epsilon \sqrt{\mathcal E(\widehat h)}-\varepsilon_{\rm opt}.
 \label{eq:safety-proof-empirical-comparison}
\end{equation}
Using the lower bound
$e_n\ge \mathcal E(\widehat h)-C_0\rho_{\Dc}(\sqrt{\mathcal E(\widehat h)}+\rho_{\Dc})$ in this inequality gives
\begin{equation}\label{eq:safety-unabsorbed}
 I(D\widehat h)\ge
\mathcal E(\widehat h)-2C_0\rho_{\Dc}(\sqrt{\mathcal E(\widehat h)}+\rho_{\Dc})-\epsilon \sqrt{\mathcal E(\widehat h)}-\varepsilon_{\rm opt}.
\end{equation}
This is the localized bound relative to the full gradient energy that is discussed in the paragraph after the
theorem.
To obtain \eqref{eq:safety-pop}, apply Young's inequality:
\[
 (2C_0\rho_{\Dc}+\epsilon)\sqrt{\mathcal E(\widehat h)}
 \le\tfrac12\mathcal E(\widehat h)+\tfrac12(2C_0\rho_{\Dc}+\epsilon)^2
 \le\tfrac12\mathcal E(\widehat h)+4C_0^2\rho_{\Dc}^2+\epsilon^2.
\]
Substitution gives the explicit bound
\[
 I(D\widehat h)\ge
 \tfrac12\mathcal E(\widehat h)-(2C_0+4C_0^2)\rho_{\Dc}^2-\epsilon^2
 -\varepsilon_{\rm opt},
\]
which proves \eqref{eq:safety-pop} with a universal $C$.

For the empirical bounds, we first bound $\sqrt{\mathcal E(\widehat h)}$ in terms of $e_n$.
The upper bound on $\mathcal E(\widehat h)-e_n$ gives
\[
\mathcal E(\widehat h)-C_0\rho_{\Dc}\sqrt{\mathcal E(\widehat h)}\le e_n+C_0\rho_{\Dc}^2.
\]
Completing the square and then taking square roots gives
\begin{align*}
 \left(\sqrt{\mathcal E(\widehat h)}-\tfrac{C_0}{2}\rho_{\Dc}\right)^2
 &\le e_n+\left(C_0+\tfrac{C_0^2}{4}\right)\rho_{\Dc}^2,\\
\sqrt{\mathcal E(\widehat h)}&\le\tfrac{C_0}{2}\rho_{\Dc}
      +\sqrt{e_n+\left(C_0+\tfrac{C_0^2}{4}\right)\rho_{\Dc}^2}\\
 &\le\sqrt{e_n}+C_2\rho_{\Dc},\qquad
 C_2:=\tfrac{C_0}{2}+\sqrt{C_0+\tfrac{C_0^2}{4}},
\end{align*}
where the last line uses $\sqrt{u+v}\le\sqrt u+\sqrt v$ for
$u,v\ge0$. Substituting this upper bound for $\sqrt{\mathcal E(\widehat h)}$ in
\eqref{eq:safety-proof-empirical-comparison} we obtain
\begin{align*}
 I(D\widehat h)
 &\ge e_n-(C_0\rho_{\Dc}+\epsilon)\sqrt{e_n}
       -C_3\rho_{\Dc}^2-C_2\epsilon\rho_{\Dc}
       -\varepsilon_{\rm opt},\\
 C_3&:=C_0(C_2+1).
\end{align*}

For the simpler empirical bound, Young's inequality gives
\[
 (C_0\rho_{\Dc}+\epsilon)\sqrt{e_n}
 \le\tfrac12e_n+C_0^2\rho_{\Dc}^2+\epsilon^2,
 \qquad
 \epsilon\rho_{\Dc}\le\tfrac12(\epsilon^2+\rho_{\Dc}^2).
\]
Consequently,
\begin{align*}
 I(D\widehat h)
 &\ge\tfrac12e_n
   -\left(C_0^2+C_3+\tfrac{C_2}{2}\right)\rho_{\Dc}^2
   -\left(1+\tfrac{C_2}{2}\right)\epsilon^2
   -\varepsilon_{\rm opt}\\
 &\ge\tfrac12e_n-C(\rho_{\Dc}^2+\epsilon^2)
   -\varepsilon_{\rm opt}
\end{align*}
for a sufficiently large universal $C$. This proves
\eqref{eq:safety-emp}. A single universal $C$ can be chosen large
enough for all four bounds, so they hold simultaneously on $\Omega$.

Finally, define the negative part $[u]_-:=\max\{-u,0\}$.
Since $\mathcal E(\widehat h)\ge0$, \eqref{eq:safety-pop} gives
\[
 [I(D\widehat h)]_-
 \le C(\rho_{\Dc}^2+\epsilon^2)+\varepsilon_{\rm opt}.
\]
If $\varepsilon_{\rm opt}\le C_{\rm opt}\rho_{\Dc}^2$ for a fixed
constant $C_{\rm opt}$, then
\[
 \rho_{\Dc}^2
 =\left(\delta_{\Dc}+\sqrt{\ell_n/n}\right)^2
 \le2\delta_{\Dc}^2+2\ell_n/n
\]
proves the stated order
$O(\delta_{\Dc}^2+\ell_n/n+\epsilon^2)$ for the negative improvement.

\end{proof}

\subsection{Justification for graph-norm based approximation}
\label{sec:value-approximation}

All the guarantees in the paper measure the misspecification of the hypothesis class through the
graph-norm distance between the true response and its comparator (\cref{def:graph-approximation}), which
controls both function values and gradients. A natural question is whether this is necessary, i.e.,
whether a hypothesis class that approximates only the function values of the true response would
suffice. The following example shows that this is not the case. In the example, the instrument detects a
low-frequency component of the response, and the hypothesis class approximates the true response by a
function in which the sign of a high-frequency component is flipped. The function values of this
component are small, but its derivatives are not. Hence, the value error of the class tends to zero, but the direction selected from the class harms the
policy by a fixed amount. The example uses smooth bounded functions and a one-dimensional convex hypothesis class.

\Needspace{6\baselineskip}
\begin{proposition}[Vanishing value error with negative policy improvement]
\label{prop:value-misspecification}
For each integer $k\ge3$, let $Z$ be Rademacher and let
$X\mid Z=z$ have density $(1+z\cos x/2)/(2\pi)$ on $(0,2\pi)$.
Let $U=0$, $Y=g_k(X)$, and
\[
 g_k(x)=\tfrac14\cos x+\tfrac1{2k}\cos(kx),\qquad
 h_k(x)=\tfrac14\cos x-\tfrac1{2k}\cos(kx),\qquad
 \Hc_k=\{t h_k:0\le t\le1\}.
\]
Take the ambient response space to be $H^1_{\rm per}(0,2\pi)$. The marginal distribution of $X$ is uniform, $\Hc_k$ satisfies the
boundedness and convexity requirements of \cref{ass:statistical}, and
\[
 \inf_{h\in\Hc_k}\norm{h-g_k}_2\le\frac1{\sqrt2 k}\longrightarrow0.
\]
Nevertheless the only exactly NPIV-feasible member of $\Hc_k$ is $h_k$,
so its constrained minimum gradient selection has
\[
 I(Dh_k)=-\frac3{32},\qquad \mathcal E(h_k)=\frac5{32}.
\]
Thus no uniform extension with an approximation term tending to zero
as a function only of value error can retain either the empirical policy improvement
lower bound or nonnegative population improvement over these
models. In particular, a derivative modulus tending to zero cannot hold
uniformly for their value-accurate comparators.
\end{proposition}

\begin{proof}
The density is positive and integrates to one. Averaging over $Z$ gives the uniform marginal law. Orthogonality of trigonometric modes yields $T\cos x=z/4$ and $T\cos(kx)=0$, hence
\[
 Tg_k=Th_k=z/16 .
\]
Feasibility of $t h_k$ therefore forces $t=1$. The envelopes follow from
\[
 |g_k|,\ |h_k|\le\tfrac14+\tfrac1{2k}<1,\qquad |Dh_k|\le\tfrac14+\tfrac12<1 .
\]
Also $g_k-h_k=\cos(kx)/k$, whose squared value norm is $1/(2k^2)$. The derivatives are
\[
 Dg_k=-\tfrac14\sin x-\tfrac12\sin(kx),\qquad
 Dh_k=-\tfrac14\sin x+\tfrac12\sin(kx),
\]
and orthogonality gives
\[
 I(Dh_k)=\tfrac12\Bigl(\tfrac1{16}-\tfrac14\Bigr)=-\tfrac3{32},\qquad
 \mathcal E(h_k)=\tfrac12\Bigl(\tfrac1{16}+\tfrac14\Bigr)=\tfrac5{32}.
\]

If an empirical policy improvement inequality with coefficient $1/2$ on this energy and an approximation
remainder tending to zero were valid, its population limit would assert $-3/32\ge5/64$, a
contradiction.

More directly, if a sequence of comparators $t_k h_k$ has value error tending to zero, its
low-frequency error forces $t_k\to1$, whereas $\norm{D(t_kh_k-g_k)}_2^2$ tends to $1/2$. This rules out
the stated uniform derivative modulus.
\end{proof}

\section{Proofs of Results in Section~\ref{sec:sources}}
\label{app:proofs-approaching}

\subsection{Proof of Theorem ~\ref{thm:relaxed-margin}}

\begin{proof}
Let $\Omega$ be the event in \cref{lem:concentration}, so that
$\Pp(\Omega)\ge1-\eta$. On this event, the comparator $h^\circ$
belongs to $\mathcal C_n$. Hence $\mathcal C_n$ is nonempty, and the
assumptions on the estimator give
$\widehat h\in\mathcal C_n$ and
$\operatorname{gap}_n(\widehat h)\le\varepsilon_{\rm opt}$.
All probabilistic conclusions below are established on this single event.

First fix $f\in\F$. Since $\F$ is symmetric, empirical feasibility
applied to both $f$ and $-f$ gives
\[
 \left|\Pn[(Y-\widehat h(X))f(Z)]\right|\le\tau_n.
\]
The uniform bound \eqref{eq:moment-conc} applies simultaneously to every
$h\in\Hc$ and $f\in\F$, and therefore also to the data-dependent choice
$h=\widehat h$. The triangle inequality then gives
\begin{align*}
 \left|\E[(Y-\widehat h(X))f(Z)]\right|
 &\le
 \left|\Pn[(Y-\widehat h(X))f(Z)]\right| +
 \left|(\Pn-\E)[(Y-\widehat h(X))f(Z)]\right|\\
 &\le\tau_n+Cr_{\Mc}=t_n.
\end{align*}
Taking the supremum over $f\in\F$ yields
$m_{\F}(\widehat h)\le t_n$. Thus $\widehat h$ belongs to the set
whose energy infimum defines $\kappa_{\Hc,\F}(t_n)$, and
\[
 0\le\kappa_{\Hc,\F}(t_n)
 \le\mathcal E(\widehat h)\le1.
\]
The last inequality follows from the gradient envelope in
\cref{ass:statistical}. This proves \eqref{eq:relaxed-feasibility}
and also shows that the relaxed infimum at $t_n$ is finite.

We next establish the improvement bounds on the same event.
On the event $\Omega$, the proof of \Cref{thm:safety}
establishes \eqref{eq:safety-global} and \eqref{eq:safety-pop}, that is, for a sufficiently large
universal $C$,
\[
 I(D\widehat h)\ge \mathcal E(\widehat h)-Cr_{\Dc}-\epsilon \sqrt{\mathcal E(\widehat h)}-\varepsilon_{\rm opt},\qquad
 I(D\widehat h)\ge\tfrac12\mathcal E(\widehat h)-C(\rho_{\Dc}^2+\epsilon^2)-\varepsilon_{\rm opt}.
\]
Using $\sqrt{\mathcal E(\widehat h)}\le1$ in the first bound and the lower bound $\mathcal E(\widehat h)\ge\kappa_{\Hc,\F}(t_n)$ in both gives
\eqref{eq:relaxed-welfare-full} and \eqref{eq:relaxed-welfare}.

By the definition of $\omega_n$ and finiteness of
$\kappa_{\Hc,\F}(t_n)$ on $\Omega$,
\[
 \kappa-\omega_n(t_n)
 =\min\{\kappa,\kappa_{\Hc,\F}(t_n)\}
 \le\kappa_{\Hc,\F}(t_n).
\]
Substituting this inequality into \eqref{eq:relaxed-welfare-full} and \eqref{eq:relaxed-welfare} gives
\eqref{eq:relaxation-gap-bound}.

Finally, suppose the tests characterize exact population feasibility
on $\Hc$. Since $0\in\F$, we have $m_{\F}(h)\ge0$ for every $h\in\Hc$.
For any $h\in\Hc\cap\Cc$ and any $f\in\F$, conditional expectation and
$\E[U\mid Z]=0$ give
\[
 \E[(Y-h(X))f(Z)]
 =\E[(Tg(Z)-Th(Z))f(Z)]=0.
\]
Hence $m_{\F}(h)=0$. Conversely, if $h\in\Hc$ and
$m_{\F}(h)\le0$, nonnegativity gives $m_{\F}(h)=0$, and the
characterization assumption gives $Th=Tg$. Because $\Hc\subset\V$,
this means $h\in\Cc$. We have proved the set identity
\[
 \{h\in\Hc:m_{\F}(h)\le0\}=\Hc\cap\Cc.
\]
Taking the energy infimum and using $\Hc\cap\Cc\subseteq\Cc$ gives
\[
 \kappa_{\Hc,\F}(0)
 =\inf_{h\in\Hc\cap\Cc}\mathcal E(h)
 \ge\inf_{h\in\Cc}\mathcal E(h)=\kappa.
\]
If $\Hc\cap\Cc$ is empty, the left side is $+\infty$ by convention,
so the inequality still holds. If this intersection contains a
sequence $(h_j)_{j\ge1}$ with $\mathcal E(h_j)\to\kappa$, then
$\kappa_{\Hc,\F}(0)\le\mathcal E(h_j)$ for every $j$.
Taking the limit gives $\kappa_{\Hc,\F}(0)\le\kappa$, and therefore
equality. An attained minimizer in $\Hc\cap\Cc$ supplies such a
sequence by taking $h_j$ equal to that minimizer for every $j$.
\end{proof}

\subsection{Proof of Theorem ~\ref{thm:benchmark}}

\begin{proof}
Fix a dual representation $q$ and size $Q>0$ as in
\cref{def:certificate}, and write
$\alpha:=\ip{a_\star}{Dg}=I(a_\star)$.
By \cref{thm:closure-convex},
\[
 \norm{a_\star}_2^2=\kappa,\qquad \alpha\ge\kappa.
\]
All expectations below are finite because $q\in L^2(P_Z)$ and
$g,h\in L^2(P_X)$ for $h\in\Hc$.

We first prove the asserted bounds on the anchor discrepancy.
The identity \eqref{eq:source} holds at the comparator
$h^\circ\in\Hc$, so
\begin{align*}
 \E[qg]-\alpha
 &=\E[q(g-h^\circ)]+\E[qh^\circ]-\ip{a_\star}{Dg}\\
 &=\E[q(g-h^\circ)]+\ip{a_\star}{D(h^\circ-g)}.
\end{align*}
The triangle inequality, Cauchy--Schwarz, and the two components of
the graph approximation error give
\begin{align*}
 \chi_q
 &=|\E[qg]-\alpha|\\
 &\le\norm q_2\norm{g-h^\circ}_2
       +\norm{a_\star}_2\norm{D(h^\circ-g)}_2\\
 &\le(\norm q_2+\sqrt\kappa)\epsilon.
\end{align*}
Since $q/Q\in\operatorname{conv}(\F)$ and the test functions have envelope one, which is inherited by their convex combinations,
\[
 \norm q_2^2
 =Q^2\E[(q(Z)/Q)^2]\le Q^2.
\]
Hence $\chi_q\le(Q+\sqrt\kappa)\epsilon$.
If the source identity also holds at $g$, then
$\E[qg]=\ip{a_\star}{Dg}=\alpha$, so $\chi_q=0$.

Next fix $h\in\Hc$. The structural model and $\E[U\mid Z]=0$ give
\[
 \E[(Y-h(X))(q(Z)/Q)]
 =\E[(g(X)-h(X))(q(Z)/Q)].
\]
Since $\F$ is symmetric, both $q/Q$ and $-q/Q$ belong to $\operatorname{conv}(\F)$. Moreover, for a convex combination $f=\sum_i\lambda_if_i$ of test functions $f_i\in\F$, linearity gives $\E[(Y-h(X))f(Z)]=\sum_i\lambda_i\E[(Y-h(X))f_i(Z)]\le m_{\F}(h)$. The definition
of $m_{\F}(h)$ therefore implies
\begin{align*}
 |\E[q(h-g)]|
 &=Q\left|\E[(Y-h(X))(q(Z)/Q)]\right|\\
 &\le Q\,m_{\F}(h).
\end{align*}
In addition, the anchor discrepancy gives
$\E[qg]\ge\alpha-\chi_q$.
Expanding a nonnegative squared distance and using the source
identity at $h$ yields
\[
 0\le\norm{Dh-a_\star}_2^2
   =\mathcal E(h)-2\ip{a_\star}{Dh}+\kappa
   =\mathcal E(h)-2\E[qh]+\kappa.
\]
Consequently,
\begin{align*}
 \mathcal E(h)
 &\ge2\E[qh]-\kappa\\
 &=2\E[q(h-g)]+2\E[qg]-\kappa\\
 &\ge-2Q\,m_{\F}(h)+2\alpha-2\chi_q-\kappa\\
 &\ge\kappa-2Q\,m_{\F}(h)-2\chi_q.
\end{align*}
The last line follows from $\alpha\ge\kappa$.

For a fixed $t\ge0$, if the relaxed set
$\{h\in\Hc:m_{\F}(h)\le t\}$ is nonempty, every element of this set
satisfies
$\mathcal E(h)\ge\kappa-2Qt-2\chi_q$.
Taking its energy infimum proves \eqref{eq:source-relaxed-margin}.
If the set is empty, its infimum is $+\infty$ by convention, so
the same inequality holds.

The population bound just proved implies
$\omega_n(t)\le2Qt+2\chi_q$.
Now invoke \cref{thm:relaxed-margin} with
$t_n=\tau_n+Cr_{\Mc}$. On its single event of probability at
least $1-\eta$, \eqref{eq:relaxed-feasibility} gives
\[
 \mathcal E(\widehat h)
 \ge\kappa_{\Hc,\F}(t_n)
 \ge\kappa-2Qt_n-2\chi_q.
\]
This is \eqref{eq:benchmark-energy}.
On the same event, \eqref{eq:relaxed-welfare-full} and \eqref{eq:relaxed-welfare} give
\begin{align*}
 I(D\widehat h)
 &\ge\kappa_{\Hc,\F}(t_n)-Cr_{\Dc}-\epsilon-\varepsilon_{\rm opt}
  \ge\kappa-2Qt_n-2\chi_q-Cr_{\Dc}-\epsilon-\varepsilon_{\rm opt},\\
 I(D\widehat h)
 &\ge\tfrac12\kappa_{\Hc,\F}(t_n)-C(\rho_{\Dc}^2+\epsilon^2)-\varepsilon_{\rm opt}
  \ge\tfrac12\kappa-Qt_n-\chi_q-C(\rho_{\Dc}^2+\epsilon^2)-\varepsilon_{\rm opt},
\end{align*}
which are \eqref{eq:benchmark-welfare} and \eqref{eq:benchmark-welfare-half}.
\end{proof}%
\section{Proofs of Results in Section~\ref{sec:sparse-linear}}
\label{app:linear-proofs}

\subsection{Proof of Corollary~\ref{cor:linear-improvement}}

\begin{proof}[Proof of \cref{cor:linear-improvement}]
Let $S=\bar R$, so $S=1$ in case (a) and
$S=\max\{R_1,1\}$ in case (b).

\textbf{The rescaled statistical problem.}
Define $Y'=Y/S$, $g'=g/S$, $U'=U/S$,
$\V'=\{h/S:h\in\V\}$, and
$\Hc'=\{h_\vartheta:\vartheta\in\Theta_\Hc/S\}$.
The law of $(X,Z)$, the feature map, the derivative operator,
and the tests are unchanged. The response class $\V'$ is
convex, contains $g'$ and all constants, and the model satisfies
$Y'=g'(X)+U'$ with $\E[U'\mid Z]=0$.
The comparator coefficient $\vartheta^\circ=\theta^\circ/S$
belongs to the rescaled coefficient set. Homogeneity of the graph
norm, computed using the unchanged marginal law and derivative, gives
$\|h_{\vartheta^\circ}-g'\|_{\V}
=S^{-1}\|h_{\theta^\circ}-g\|_{\V}\le\epsilon/S=:\epsilon'$.

In case (a), for every $\theta$ in the coefficient ball,
$|h_\theta(x)|\le\|\theta\|\|\phi(x)\|\le R_2L_\phi\le1$
and $\|Dh_\theta(x)\|\le
\|J_\phi(x)\|_{\rm op}\|\theta\|\le1$.
In case (b), let $R=R_1/S\le1$. For $\|\vartheta\|_1\le R$,
\[
 |h_\vartheta(x)|\le\sum_j|\vartheta_j||\phi_j(x)|\le R\le1,
 \qquad
 \|Dh_\vartheta(x)\|
 \le\sum_j|\vartheta_j|\|\nabla\phi_j(x)\|\le R\le1.
\]
Also $|Y'|\le1$ because $S\ge1$.
The coefficient balls are convex and contain zero, and the
test class is symmetric, star-convex, and bounded by one.
Thus the rescaled problem satisfies \cref{ass:statistical}.
Let $\Dc'$ and $\Mc'$ denote its gradient and residual-test
classes, and let $r_{\Dc'},r_{\Mc'},\rho_{\Dc'}$ denote their
complexities.

\textbf{A maximal inequality used for all three classes.}
Conditional on the observations, a signed average
$\xi=n^{-1}\sum_i\sigma_i z_i$, with $|z_i|\le1$, satisfies
\[
 \E_\sigma e^{t\xi}
 =\prod_i\cosh(tz_i/n)
 \le\exp\left(\frac{t^2}{2n^2}\sum_i z_i^2\right)
 \le e^{t^2/(2n)}.
\]
If $\xi_1,\ldots,\xi_M$ have this same conditional bound
for both signs, Jensen's inequality and a sum over their
$2M$ signed versions give, for every $t>0$,
\[
 \E_\sigma\max_{j\le M}|\xi_j|
 \le\frac{\log(2M)}t+\frac{t}{2n}.
\]
Indeed, $\exp(t\max_j|\xi_j|)\le
\sum_{j,\pm}\exp(\pm t\xi_j)$, and then one takes
expectations and logarithms. Choosing
$t=\sqrt{2n\log(2M)}$ gives
$\E_\sigma\max_j|\xi_j|\le\sqrt{2\log(2M)/n}$.
Independence among the $\xi_j$ is not required. This is the
usual sub-Gaussian maximal inequality
\citep[Section 2.5]{boucheron2013}.

\textbf{Test complexity.}
For every sample and signs,
\[
 \sup_{f\in\F}\left|\frac1n\sum_i\sigma_i f(Z_i)\right|
 =\max_{\ell\le k}\left|\frac1n\sum_i\sigma_i\psi_\ell(Z_i)\right|.
\]
The maximal inequality and $|\psi_\ell|\le1$ therefore give
$\mathfrak R_n(\F)\le\sqrt{2\log(2k)/n}$ and
$r_\F\le\Delta_{n,k}$.

\textbf{Gradient and residual-test complexities in case (a).}
The gradient class $\Dc'$ lies in the span of the $p$
fixed columns of $J_\phi$. By \cref{lem:dimension},
\[
 \mathfrak R_n^{\rm vec}(\Dc',r)\le r\sqrt{p/n},
 \qquad
 \mathfrak R_n^{\rm vec}(\Dc')\le\sqrt{p/n}.
\]
The second bound uses the unit envelope and hence radius one.
Since $1\le p\le n$, $\delta_{\Dc'}=\sqrt{p/n}$ belongs
to $[n^{-1/2},1]$ and the first bound at this radius is
at most $\delta_{\Dc'}^2$. Moreover,
\[
 r_{\Dc'}\le\rho_{\Dc'}=\sqrt{p/n}+\sqrt{\ell_n/n}
 \le\sqrt2\,\sqrt{(p+\ell_n)/n}
 =\sqrt2\,\bar r_\Dc.
\]
In this case $Y'=Y$. The class $\Mc'$ is the union over
$\ell\le k$ of
$\{c(Y-h_\theta(X))\psi_\ell(Z)/2:
|c|\le1,\ \theta\in\Theta_\Hc\}$.
Each subclass has envelope one and is contained, pointwise,
in the span of the $p+1$ fixed functions
$Y\psi_\ell,\phi_1\psi_\ell,\ldots,\phi_p\psi_\ell$.
With $s_n=\sqrt{(p+1+\log k)/n}$,
\cref{lem:union-dimension} at radius one yields
$\mathfrak R_n(\Mc')\le4s_n+3s_n^2$.
When $s_n\le1$ this is at most $7s_n$; when $s_n>1$,
the unit envelope gives $\mathfrak R_n(\Mc')\le1\le7s_n$.
Since $p+1+\log k\le2(p+\log k)$,
\[
 r_{\Mc'}\le7s_n+\sqrt{\ell_n/n}
 \le(7\sqrt2+1)\sqrt{(p+\log k+\ell_n)/n}
 =(7\sqrt2+1)\bar r_\Mc .
\]

\textbf{Complexities in case (b).}
For each feature let
$\xi_j=n^{-1}\sum_{i,c}\sigma_{ic}\partial_c\phi_j(X_i)$.
For every sample and signs,
\[
 \sup_{\|\vartheta\|_1\le R}
 \left|\frac1n\sum_{i,c}\sigma_{ic}(J_\phi(X_i)\vartheta)_c\right|
 =R\max_{j\le p}|\xi_j|.
\]
The upper bound is H\"older's inequality; equality follows
by placing the whole coefficient budget on an index that
attains the maximum, with the matching sign.
The same moment-generating-function calculation as above,
now over $(i,c)$, gives variance proxy
$n^{-2}\sum_{i,c}(\partial_c\phi_j(X_i))^2\le1/n$.
Thus
$\mathfrak R_n^{\rm vec}(\Dc')\le
R\sqrt{2\log(2p)/n}\le\sqrt{2\log(2p)/n}$,
and $r_{\Dc'}\le\Delta_{n,p}$.
Let $\delta=(2\log(2p)/n)^{1/4}$. The assumption
$2\log(2p)\le n$ gives $\delta\le1$, while
$2\log(2p)\ge1$ gives $\delta\ge n^{-1/2}$.
Monotonicity under localization shows
$\mathfrak R_n^{\rm vec}(\Dc',\delta)
\le\mathfrak R_n^{\rm vec}(\Dc')\le\delta^2$.
This verifies the critical-radius requirement as well.

For the residual-test class define
$\zeta_\ell=n^{-1}\sum_i\sigma_iY'_i\psi_\ell(Z_i)$ and
$\zeta_{\ell j}=n^{-1}\sum_i\sigma_i\phi_j(X_i)\psi_\ell(Z_i)$.
The triangle inequality and H\"older's inequality give
\[
 \mathfrak R_n(\Mc')
 \le\tfrac12\E\max_{\ell\le k}|\zeta_\ell|
       +\tfrac R2\E\max_{\ell\le k,j\le p}|\zeta_{\ell j}|
 \le\tfrac12\sqrt{2\log(2k)/n}
       +\tfrac R2\sqrt{2\log(2kp)/n}
 \le\sqrt{2\log(2kp)/n}.
\]
The maximal inequality applies because both kinds of summands
are bounded by one. Thus $r_{\Mc'}\le\Delta_{n,kp}$.
These calculations establish every bound in
\eqref{eq:linear-radii}; they also show explicitly that the
constants can be chosen independently of $p,k,n,R_1,R_2$.

\textbf{Feasibility, attainment, and exact empirical optimality.}
Apply \cref{lem:concentration} to the rescaled problem and let
$\Omega$ be its event, with probability at least $1-\eta$.
Let $C_F$ be its constant in the noise bound, and choose the
universal $c_0$ at least $C_F$.
On $\Omega$, for the rescaled comparator,
\[
 \sup_{f\in\F}|\Pn[(Y'-h_{\vartheta^\circ})f]|
 \le\epsilon'+C_Fr_\F
 \le\epsilon'+c_0\Delta_{n,k}=\tau_n/S=:\tau'.
\]
For every $\vartheta=\theta/S$, the empirical moment residual
equals $(\widehat b-\widehat B\theta)/S$.
Thus $\vartheta^\circ$ is feasible for the rescaled program
exactly when $\theta^\circ$ is feasible for the original one.
In particular $\Theta_n$ is nonempty on $\Omega$.
It is the intersection of a compact coefficient ball and
finitely many closed half-spaces, hence is compact and convex.
The objective is continuous, so it attains its minimum.
For any minimizer $\widehat\theta$ and any $\theta\in\Theta_n$,
differentiate the objective along the feasible segment
$\widehat\theta+s(\theta-\widehat\theta)$ at $s=0$ to obtain
$\widehat\theta^{\mathsf T}G_n(\theta-\widehat\theta)\ge0$.
Therefore its variational gap is at most zero and is exactly
zero by taking $\theta=\widehat\theta$.

The change of coefficients $\vartheta=\theta/S$ is a bijection
between the two feasible sets and divides the objective by
$S^2$. It maps every original minimizer to a rescaled minimizer
$\widehat h'=\widehat h/S$, also with zero gap.
Since $r_\F\le\Delta_{n,k}$, the tolerance $\tau'$ is at least as large as the tolerance $c_0r_\F+\epsilon'$ of \cref{thm:safety} for the rescaled problem.
The proof of \cref{thm:safety} then applies on $\Omega$ at tolerance
$\tau'$, since it uses that tolerance only to ensure feasibility
of the comparator.

\textbf{Transferring the improvement inequalities.}
Let $I'$ denote the policy derivative for $g'$, and let
$\mathcal E'$ and $e'_n$ denote the energies of $\widehat h'$.
Linearity gives, with no change in the population or empirical
law of $X$,
\[
 I'(D\widehat h')=I(D\widehat h)/S^2,\qquad
 \mathcal E'=\mathcal E(\widehat h)/S^2,\qquad
 e'_n=e_n/S^2 .
\]
Let $C_s$ be the universal constant in \cref{thm:safety}, and let $C=2C_s$.
Its two full-energy inequalities, multiplied by $S^2$, give
\begin{align*}
 I(D\widehat h)
 &\ge\mathcal E(\widehat h)-C_sS^2r_{\Dc'}
                         -\epsilon\sqrt{\mathcal E(\widehat h)},\\
 I(D\widehat h)
 &\ge e_n-C_sS^2r_{\Dc'}-S\epsilon .
\end{align*}
Use $r_{\Dc'}\le C_D\bar r_\Dc$, with
$C_D=\sqrt2$ in case (a) and $C_D=1$ in case (b).
Since $C_D\le2$, these prove \eqref{eq:linear-full} and
\eqref{eq:linear-full-emp} with constant $C$.
In case (a), $S=1$ and $\rho_{\Dc'}^2\le2\bar r_\Dc^2$,
so the half-energy empirical inequality gives
\[
 I(D\widehat h)\ge\tfrac12e_n
                  -C_s(2\bar r_\Dc^2+\epsilon^2)
 \ge\tfrac12e_n-C(\bar r_\Dc^2+\epsilon^2).
\]
All these bounds hold on the same event $\Omega$.
\end{proof}

\subsection{Proof of Proposition~\ref{prop:tested-energy}}

The next proposition is the general form of \cref{prop:tested-energy}, in which the tested moment
equations are anchored at an arbitrary tolerance $t_0\ge0$ instead of being imposed exactly. For
$t\ge0$, let
\begin{equation}\label{eq:tested-dictionary-energy-anchor}
 \kappa_{\phi,\psi}(t):=\min_{\theta\in\R^p}\ \theta^{\mathsf T}G\theta
 \qquad\text{subject to}\qquad \norm{b-B\theta}_\infty\le t ,
\end{equation}
so that $\kappa_{\phi,\psi}(0)=\kappa_{\phi,\psi}$ and, since $\Hc\subseteq\V_\phi$,
$\kappa_{\Hc,\F}(t)\ge\kappa_{\phi,\psi}(t)$ for every $t$. The program is always feasible at
$t=\epsilon$, since the comparator satisfies $m_{\F}(h_{\theta^\circ})\le\epsilon$.

\Needspace{6\baselineskip}
\begin{proposition}[Stability of the tested dictionary energy at a general anchor]
\label{prop:tested-energy-anchor}
Let $t_0\ge0$ be such that the program \eqref{eq:tested-dictionary-energy-anchor} is feasible at $t=t_0$.
\begin{enumerate}
 \item\label{it:tested-multiplier} The minimum is attained at $t=t_0$, and for every minimizer
 $\theta_{t_0}$ there exists a multiplier $w\in\R^k$ such that
 \begin{equation}\label{eq:tested-multiplier}
  G\theta_{t_0}=B^{\mathsf T}w,\qquad
  w^{\mathsf T}(b-B\theta_{t_0})=t_0\norm w_1 .
 \end{equation}
 \item\label{it:tested-stability} Let $Q_\phi(t_0)$ be the smallest $\ell_1$ norm of a multiplier that
 satisfies \eqref{eq:tested-multiplier} for some minimizer. Then, for every $t\ge0$,
 \begin{equation}\label{eq:tested-stability-anchor}
  \kappa_{\phi,\psi}(t)\ \ge\ \kappa_{\phi,\psi}(t_0)-2\,Q_\phi(t_0)\,(t-t_0).
 \end{equation}
\end{enumerate}
\end{proposition}

\begin{proof}
Let $P_{t_0}=\{\theta:\|b-B\theta\|_\infty\le t_0\}$, which is
nonempty by assumption, and let $G^{1/2}$ be the positive semidefinite
square root of $G$.

\textbf{Attainment of the primal minimum.}
The set $P_{t_0}$ is a polyhedron, since it is defined by $2k$ affine
inequalities. Its linear image $G^{1/2}P_{t_0}$ is also a polyhedron
and therefore is closed. One way to see this standard
finite-dimensional fact is to write the image as the coordinate
projection of $\{(\theta,y):\theta\in P_{t_0},\ y=G^{1/2}\theta\}$;
eliminating the $\theta$ coordinates preserves a finite system of
linear inequalities. Choose any $y_0$ in the image. Its intersection
with the closed Euclidean ball of radius $\|y_0\|$ is nonempty and
compact, so the continuous function $y\mapsto\|y\|^2$ attains a
minimum there. Every point of the image outside this ball has
at least that minimum norm. Thus the minimum is attained on the
whole image. Choose a minimizing $y_{t_0}$ and a preimage
$\theta_{t_0}\in P_{t_0}$. Since
$\theta^{\mathsf T}G\theta=\|G^{1/2}\theta\|^2$, this preimage
attains the original minimum.

\textbf{Existence and signs of the multipliers.}
We verify the affine-constraint optimality condition directly,
including at $t_0=0$, where strict feasibility can fail.
Let $r(\theta)=b-B\theta$, $r_0=r(\theta_{t_0})$, and write the constraints as
$r_\ell(\theta)-t_0\le0$ and $-r_\ell(\theta)-t_0\le0$.
Their outward normal vectors are respectively
$n_\ell^+=-B_{\ell,\cdot}^{\mathsf T}$ and
$n_\ell^-=B_{\ell,\cdot}^{\mathsf T}$.
Let $\mathcal A$ be the set of constraints active at $\theta_{t_0}$,
and let $N$ be the cone generated by their normals. This is a
closed polyhedral cone.

If a vector $d$ satisfies $n^{\mathsf T}d\le0$ for every active
normal, then $\theta_{t_0}+sd$ is feasible for all sufficiently
small $s>0$. Indeed, each active affine constraint remains satisfied;
each inactive constraint has a strictly positive slack, and if its
normal has positive inner product with $d$, choose $s$ smaller
than its slack divided by that inner product. There are only
finitely many inactive constraints, so one positive $s$ works
for all of them. Optimality of $\theta_{t_0}$ and the directional
derivative of $\tfrac12\theta^{\mathsf T}G\theta$ then give
$(G\theta_{t_0})^{\mathsf T}d\ge0$.

If $-G\theta_{t_0}$ did not belong to $N$, separation from this
closed convex cone would give a vector $d$ such that
$n^{\mathsf T}d\le0$ for every $n\in N$ but
$(-G\theta_{t_0})^{\mathsf T}d>0$. This contradicts the last
directional inequality. Consequently $-G\theta_{t_0}\in N$.
Expressing it as a nonnegative combination of active normals,
and assigning multiplier zero to each inactive constraint, yields
$\lambda^+,\lambda^-\ge0$ such that
\begin{align*}
 G\theta_{t_0}&=B^{\mathsf T}(\lambda^+-\lambda^-),\\
 \lambda^+_\ell(r_{0,\ell}-t_0)&=0,\qquad
 \lambda^-_\ell(-r_{0,\ell}-t_0)=0\quad(\ell\le k).
\end{align*}
This is the polyhedral form of the Karush--Kuhn--Tucker condition
\citep{boyd2004}; the argument shows why it does not
require strict feasibility here.

Let $w=\lambda^+-\lambda^-$. If $t_0>0$, the two constraints
at a fixed index cannot both be active. Thus at most one of
$\lambda^+_\ell,\lambda^-_\ell$ is nonzero, and complementary
slackness gives
\[
 w_\ell r_{0,\ell}
 =t_0(\lambda^+_\ell+\lambda^-_\ell)=t_0|w_\ell|.
\]
If $t_0=0$, feasibility gives $r_0=0$, so the same identity holds
with both sides zero. Summing proves \eqref{eq:tested-multiplier}.

\textbf{Attainment of the smallest multiplier norm.}
Any two minimizing images $y_1,y_2$ must coincide: otherwise their
feasible midpoint has squared norm
$\tfrac12\|y_1\|^2+\tfrac12\|y_2\|^2-\tfrac14\|y_1-y_2\|^2$,
strictly below the minimum. Therefore all minimizing coefficients
have the same vector $G^{1/2}\theta$ and the same vector
$c_0=G\theta$. Let $\kappa_0=\kappa_{\phi,\psi}(t_0)$.
For every minimizer, $c_0^{\mathsf T}\theta=\theta^{\mathsf T}G\theta
=\kappa_0$. Hence the admissible multipliers for any such minimizer
are exactly
\[
 \mathcal W_0=\{w:B^{\mathsf T}w=c_0,\
                  w^{\mathsf T}b-\kappa_0=t_0\|w\|_1\}.
\]
This set is nonempty by the multiplier construction and closed
by continuity. Choose $w_0\in\mathcal W_0$ and intersect
$\mathcal W_0$ with $\{w:\|w\|_1\le\|w_0\|_1\}$.
The intersection is nonempty and compact, so the continuous
norm attains its minimum there. No multiplier outside this ball
can have smaller norm. Thus there exists
$w_\star\in\mathcal W_0$ with $\|w_\star\|_1=Q_\phi(t_0)$.

\textbf{Stability at any tolerance.}
Fix this $w_\star$ and any minimizing $\theta_{t_0}$. For $t\ge0$,
let $\theta$ satisfy $\|b-B\theta\|_\infty\le t$, if such a
coefficient exists. Nonnegativity of
$(\theta-\theta_{t_0})^{\mathsf T}G(\theta-\theta_{t_0})$ gives
\begin{align*}
 \theta^{\mathsf T}G\theta
 &\ge\kappa_0+2\theta_{t_0}^{\mathsf T}G(\theta-\theta_{t_0})\\
 &=\kappa_0+2w_\star^{\mathsf T}B(\theta-\theta_{t_0})\\
 &=\kappa_0+2w_\star^{\mathsf T}(b-B\theta_{t_0})
                 -2w_\star^{\mathsf T}(b-B\theta)\\
 &\ge\kappa_0+2t_0\|w_\star\|_1-2t\|w_\star\|_1\\
 &=\kappa_{\phi,\psi}(t_0)-2Q_\phi(t_0)(t-t_0).
\end{align*}
The fourth line uses complementarity at the anchor and
$|w_\star^{\mathsf T}(b-B\theta)|\le
\|w_\star\|_1\|b-B\theta\|_\infty$.
Taking the infimum over feasible $\theta$ proves
\eqref{eq:tested-stability-anchor}. If that set is empty, its
value is $+\infty$ and the same lower bound holds.
Using the attained $w_\star$ proves the inequality for both
$t<t_0$ and $t\ge t_0$, without reversing an optimization over
multiplier norms when the sign of $t-t_0$ changes.
\end{proof}

\begin{proof}[Proof of \cref{prop:tested-energy}]
Since $B\theta=b$ is solvable, the feasible set at $t_0=0$ in
\eqref{eq:tested-dictionary-energy-anchor} is nonempty.
The general-anchor proposition (\cref{prop:tested-energy-anchor}) therefore gives an attained
minimum $\theta_\psi$ and a multiplier satisfying
$G\theta_\psi=B^{\mathsf T}w$. At this anchor,
$b-B\theta_\psi=0$, so the complementarity equation is $0=0$
and places no additional restriction on $w$.

All minimizing coefficients have the same $G\theta_\psi$ by
the uniqueness of their minimizing image proved above.
Thus the quantity $Q_\phi$ in the statement is independent
of the chosen minimizer and equals the attained minimum
$Q_\phi(0)$. For any $t\ge0$, every coefficient defining
$\kappa_{\Hc,\F}(t)$ also satisfies
$\|b-B\theta\|_\infty\le t$, because
$m_\F(h_\theta)=\|b-B\theta\|_\infty$.
Its coefficient restriction $\theta\in\Theta_\Hc$ can only
raise the infimum relative to the unrestricted program.
Consequently
\[
 \kappa_{\Hc,\F}(t)
 \ge\kappa_{\phi,\psi}(t)
 \ge\kappa_{\phi,\psi}(0)-2Q_\phi(0)t
 =\kappa_{\phi,\psi}-2Q_\phi t .
\]
The first inequality includes the empty-set case with value
$+\infty$, and the second is \eqref{eq:tested-stability-anchor}
at $t_0=0$. This proves all the assertions.
\end{proof}

\subsection{Proof of Lemma~\ref{lem:linear-dual-dictionary}}

\begin{proof}[Proof of \cref{lem:linear-dual-dictionary}]
Let $Q>0$, as in \cref{def:certificate}. Each component of
$\gamma_\star$ is finite by Cauchy--Schwarz:
$|\E[\nabla\phi_j^{\mathsf T}a_\star]|
\le\|\nabla\phi_j\|_2\|a_\star\|_2$.
Linearity of the derivative and the expectation gives
\[
 \langle a_\star,Dh_\theta\rangle
 =\E[a_\star(X)^{\mathsf T}J_\phi(X)\theta]
 =\gamma_\star^{\mathsf T}\theta\qquad(\theta\in\R^p).
\]

\textbf{Convex combinations of tests.}
We first verify the exact description
$\operatorname{conv}(\F)=\{c^{\mathsf T}\psi:\|c\|_1\le1\}$.
A convex combination of functions $s_i\psi_{\ell_i}$,
with $|s_i|\le1$ and weights $\lambda_i\ge0$ summing to one,
has coefficients
$c_\ell=\sum_{i:\ell_i=\ell}\lambda_i s_i$.
The triangle inequality gives
$\sum_\ell|c_\ell|\le\sum_i\lambda_i|s_i|\le1$.
Conversely, if $\|c\|_1\le1$, then
$c^{\mathsf T}\psi$ is the convex combination of
$\operatorname{sgn}(c_\ell)\psi_\ell$ with weights $|c_\ell|$
for the nonzero coordinates, and of zero with weight
$1-\|c\|_1$. These functions all belong to $\F$.

\textbf{Equivalence with the linear system.}
If $B^{\mathsf T}w=\gamma_\star$ and $\|w\|_1\le Q$,
let $q=w^{\mathsf T}\psi$. The test envelope implies
$|q|\le\|w\|_1$, hence $q\in L^2(P_Z)$.
The convex-hull description gives
$q/Q=(w/Q)^{\mathsf T}\psi\in\operatorname{conv}(\F)$.
For every coefficient $\theta$,
\[
 \E[q(Z)h_\theta(X)]
 =\sum_{\ell,j}w_\ell\theta_j\E[\psi_\ell(Z)\phi_j(X)]
 =w^{\mathsf T}B\theta
 =\gamma_\star^{\mathsf T}\theta
 =\langle a_\star,Dh_\theta\rangle .
\]
In particular this is the required identity on $\Hc$.

Conversely, suppose a dual representation of size $Q$ holds on
$\Hc$. The convex-hull description supplies $c$ with
$q/Q=c^{\mathsf T}\psi$ and $\|c\|_1\le1$.
Let $w=Qc$. Then $\|w\|_1\le Q$, $q=w^{\mathsf T}\psi$,
and the representation gives
$(B^{\mathsf T}w-\gamma_\star)^{\mathsf T}\theta=0$
for every $\theta\in\Theta_\Hc$.
Since this set contains a neighborhood of zero, choose $s>0$
such that $s e_j\in\Theta_\Hc$ for every coordinate vector $e_j$.
Substitution gives
$s(B^{\mathsf T}w-\gamma_\star)_j=0$ for each $j$.
Thus $B^{\mathsf T}w=\gamma_\star$.

\textbf{Solvability and the conditional moment kernel.}
In finite dimensions,
$\operatorname{range}(B^{\mathsf T})=(\ker B)^\perp$:
the inclusion follows from
$\langle B^{\mathsf T}w,v\rangle=\langle w,Bv\rangle=0$
when $Bv=0$, and equality follows because both spaces have
dimension $\operatorname{rank}B$.
Therefore the system is solvable exactly when
$\gamma_\star^{\mathsf T}v=\langle a_\star,Dh_v\rangle=0$
for every $v\in\ker B$.

To prove the last sufficient condition, first observe that
conditional expectation gives $Th_v(Z)=\mu(Z)^{\mathsf T}v$ and
\begin{equation}\label{eq:linear-moment}
 \|Th_v\|_2^2
 =\E[(\mu(Z)^{\mathsf T}v)^2]
 =v^{\mathsf T}\Gamma v.
\end{equation}
The matrix $\Gamma$ is positive semidefinite; diagonalizing it
shows that $v^{\mathsf T}\Gamma v=0$ exactly when
$v\in\ker\Gamma$. Hence $Th_v=0$ exactly for those $v$.
Now assume $\V$ is linear and contains $\V_\phi$. For
$v\in\ker\Gamma$, every $g+s h_v$, $s\in\R$, belongs to $\V$
and satisfies $T(g+s h_v)=Tg$. It therefore belongs to $\Cc$.
The population projection inequality (\cref{thm:closure-convex}) implies
\[
 \langle a_\star,Dg\rangle+
      s\langle a_\star,Dh_v\rangle\ge\kappa\qquad(s\in\R).
\]
If $\langle a_\star,Dh_v\rangle$ were nonzero, choosing $s$
with the opposite sign and arbitrarily large magnitude would
contradict this finite lower bound. It must be zero.
Thus $\gamma_\star\perp\ker\Gamma$.
When $\ker B=\ker\Gamma$, the finite-dimensional range identity
just proved yields solvability.
\end{proof}

\subsection{Proof of Corollary~\ref{cor:linear-benchmarks}}\label{sec:proof-linear-benchmarks}

\begin{proof}[Proof of \cref{cor:linear-benchmarks}]
Use the rescaled problem, classes $\Dc',\Mc'$, and event
$\Omega$ from the proof of \cref{cor:linear-improvement}, so $\Pr(\Omega)\ge1-\eta$.
On $\Omega$, the estimator is feasible and both full-energy bounds hold.
Let $C_M$ be the constant in \eqref{eq:moment-conc}, applied
to the rescaled problem, and let
$C_{\rm cl}=7\sqrt2+1$, which bounds the factor in
$r_{\Mc'}\le C_{\rm cl}\bar r_\Mc$ in both cases.

\textbf{Population feasibility and the relaxed energy.}
On $\Omega$, the rescaled problem satisfies the hypotheses of \cref{thm:safety} at tolerance
$\tau_n/S$, as shown in the proof of \cref{cor:linear-improvement}. Hence
\cref{thm:relaxed-margin} applies to it with tolerance $\tau_n/S+C_Mr_{\Mc'}$, and its feasibility
bound \eqref{eq:relaxed-feasibility} gives, on $\Omega$,
\begin{align*}
 m_\F'(\widehat h')
 &=\sup_{f\in\F}|\E[(Y'-\widehat h')f]|\\
 &\le\tau_n/S+C_Mr_{\Mc'}
 \le\tau_n/S+C_MC_{\rm cl}\bar r_\Mc .
\end{align*}
Multiplication by $S$ and
\eqref{eq:linear-moment-violation} give
\[
 \|b-B\widehat\theta\|_\infty
 =m_\F(\widehat h)
 =S m_\F'(\widehat h')
 \le\tau_n+C_MC_{\rm cl}S\bar r_\Mc .
\]
Choose the constant $C$ in the corollary at least
$C_MC_{\rm cl}$ and at least the constants in the
full-energy bounds. Then the last display is at most
$t_n=\tau_n+CS\bar r_\Mc$.
Since $\widehat\theta\in\Theta_\Hc$, the response $\widehat h$
belongs to the set defining $\kappa_{\Hc,\F}(t_n)$.
The gradient envelope is $S$, so
$\mathcal E(\widehat h)\le S^2$ and
$\epsilon\sqrt{\mathcal E(\widehat h)}\le S\epsilon$.
Using \eqref{eq:linear-full} now gives
\begin{equation}\label{eq:linear-proof-relaxed-bound}
 I(D\widehat h)\ge
 \kappa_{\Hc,\F}(t_n)-CS^2\bar r_\Dc-S\epsilon .
\end{equation}
This argument is on the single event $\Omega$ and requires no
additional probability allocation.

\textbf{The tested-dictionary benchmark.}
If $B\theta=b$ is solvable,
\cref{prop:tested-energy} gives
$\kappa_{\Hc,\F}(t_n)\ge
\kappa_{\phi,\psi}-2Q_\phi t_n$.
Substituting this bound in
\eqref{eq:linear-proof-relaxed-bound}, and recalling
$S=\bar R$, proves \eqref{eq:linear-tested}.

\textbf{The identified-energy benchmark.}
Let $w$ solve $B^{\mathsf T}w=\gamma_\star$,
let $Q=\|w\|_1\ge0$, and let $q=w^{\mathsf T}\psi$.
The dictionary-dual proof (\cref{lem:linear-dual-dictionary}) establishes, directly from this
linear equation,
$\E[qh]=\langle a_\star,Dh\rangle$ for every $h\in\Hc$.
This identity is valid even when $Q=0$, without dividing by $Q$.
The test envelope gives $\|q\|_2\le Q$.
Let $\alpha=\langle a_\star,Dg\rangle\ge\kappa$ and
$\chi_q=|\E[qg]-\alpha|$.
Using the representation at $h^\circ$ gives
\begin{align*}
 \E[qg]-\alpha
 &=\E[q(g-h^\circ)]
       +\langle a_\star,D(h^\circ-g)\rangle,\\
 \chi_q
 &\le\|q\|_2\|g-h^\circ\|_2+
       \|a_\star\|_2\|D(h^\circ-g)\|_2
 \le(Q+\sqrt\kappa)\epsilon.
\end{align*}
These inequalities use the graph error in the original scale;
they do not require unit envelope for the original hypotheses.

For any $h_\theta\in\Hc$ with
$\|b-B\theta\|_\infty\le t_n$, the conditional moment
restriction gives $\E[qU]=0$, and hence
\[
 |\E[q(g-h_\theta)]|
 =|\E[q(Y-h_\theta)]|
 =|w^{\mathsf T}(b-B\theta)|\le Qt_n .
\]
Expanding a squared distance then gives
\begin{align*}
 \mathcal E(h_\theta)
 &\ge2\langle a_\star,Dh_\theta\rangle-\|a_\star\|_2^2\\
 &=2\E[qh_\theta]-\kappa\\
 &\ge2\E[qg]-2Qt_n-\kappa\\
 &\ge2\alpha-2\chi_q-2Qt_n-\kappa\\
 &\ge\kappa-2Qt_n-2(Q+\sqrt\kappa)\epsilon .
\end{align*}
This is \eqref{eq:source-relaxed-margin} of \cref{thm:benchmark} at $t=t_n$, without dividing by $Q$. The set contains $\widehat h$ on $\Omega$. Taking its energy infimum
and substituting in \eqref{eq:linear-proof-relaxed-bound} proves
\eqref{eq:linear-kappa}. The argument also covers $w=0$: then
$q=0$, $Q=0$, and the same representation and comparator identity
give $\chi_0\le\sqrt\kappa\,\epsilon$. Thus no division by $Q$ is needed.
\end{proof}

The following argument verifies the comparison made after
\cref{cor:linear-benchmarks}.
\begin{proof}
Suppose there exists $\theta_1\in\R^p$ with
$Th_{\theta_1}=Tg$. Since $\E[Y\mid Z]=Tg$, conditional expectation
and the definitions of $B$ and $\mu$ give
\[
 b=\E[\psi(Z)Y]
   =\E[\psi(Z)Tg(Z)]
   =\E[\psi(Z)\mu(Z)^{\mathsf T}]\theta_1
   =B\theta_1.
\]
Thus $B\theta=b$ holds exactly when
$B(\theta-\theta_1)=0$, and its coefficient set is
$\theta_1+\ker B$.
On the other hand, linearity and \eqref{eq:linear-moment} give
\[
 Th_\theta=Tg
 \ \Longleftrightarrow\ T h_{\theta-\theta_1}=0
 \ \Longleftrightarrow\ \theta-\theta_1\in\ker\Gamma.
\]
The conditional-moment coefficient set is therefore
$\theta_1+\ker\Gamma$.

For every $v\in\ker\Gamma$, we have $\mu(Z)^{\mathsf T}v=0$
almost surely, and so
$Bv=\E[\psi(Z)\mu(Z)^{\mathsf T}v]=0$.
This proves $\ker\Gamma\subseteq\ker B$ without any additional
test richness assumption. If the two kernels are equal, the
two coefficient sets coincide. Their images under
$\theta\mapsto h_\theta$ then coincide as well, even if this
map is not injective. Thus
\[
 \kappa_{\phi,\psi}
 =\inf_{\substack{\theta\in\R^p\\Th_\theta=Tg}}
        \mathcal E(h_\theta).
\]
When $\V_\phi\subseteq\V$, every response in this infimum is
in $\Cc$, so its infimum is at least $\kappa$.
When $\V=\V_\phi$, these are all responses in $\Cc$, so the
two infima are equal.

If $g=h_{\theta_0}$, the required compatible coefficient
$\theta_1$ can be chosen as $\theta_0$.
Moreover, the dual identity obtained from
$B^{\mathsf T}w=\gamma_\star$ holds for every dictionary
coefficient, as shown in the proof of \cref{lem:linear-dual-dictionary}. It therefore
holds at $\theta_0$, whether or not $\theta_0\in\Theta_\Hc$.
Consequently $\E[qg]=\langle a_\star,Dg\rangle$ and
$\chi_q=0$ in this well-specified case.
\end{proof}

\section{Proofs of Results in Section~\ref{sec:initial-direction}}
\label{app:misspecification-proofs}

\subsection{A generic bound for single-test directions}
\label{sec:initial-generic}

Throughout this appendix, $G\succ0$, the vector $\theta_\phi=G^{-1}\E[J_\phi^{\mathsf T}Dg]$ is the coefficient vector of the best gradient approximator \eqref{eq:linear-projection-on-phi}, and we write $a_u:=J_\phi u$ for $u\in\R^p$. Since $\theta_\phi$ minimizes the convex quadratic function $\theta\mapsto\norm{Dg-J_\phi\theta}_2^2$, its first-order optimality condition gives
\begin{equation}\label{eq:initial-projection-identity}
 \E[J_\phi^{\mathsf T}(Dg-J_\phi\theta_\phi)]=0,
 \qquad\text{and therefore}\qquad
 I(a_u)=\E[(J_\phi u)^{\mathsf T}Dg]=\theta_\phi^{\mathsf T}Gu\quad(u\in\R^p).
\end{equation}
Thus, the improvement of every direction of the form $J_\phi u$ depends on $g$ only through $\theta_\phi$. The next lemma is the generic form of \cref{thm:initial-single-test}. We state it for arbitrary moment arrays, so that it applies both to the population moments $(B,b)$ and to their empirical counterparts $(\widehat B,\widehat b)$.

\Needspace{6\baselineskip}
\begin{lemma}[Single-test direction for generic moment arrays]
\label{lem:initial-generic}
Let $M\in\R^{k\times p}$, with rows $M_j$, and $c\in\R^k$. Let $j_\star\in\operatorname*{arg\,max}_{1\le j\le k}|c_j|$, let $s=\sgn(c_{j_\star})$, with $\sgn(0)=0$, and let
\begin{equation}\label{eq:initial-generic-rule}
 v(M,c)=s\,G^{-1}M_{j_\star}^{\mathsf T}.
\end{equation}
Then
\begin{equation}\label{eq:initial-generic-bound}
 I(a_{v(M,c)})=s\,M_{j_\star}\theta_\phi
 \qquad\text{and}\qquad
 \bigl|s\,M_{j_\star}\theta_\phi-\norm c_\infty\bigr|\le\norm{c-M\theta_\phi}_\infty .
\end{equation}
In particular, $I(a_{v(M,c)})\ge\norm c_\infty-\norm{c-M\theta_\phi}_\infty$, and the vector $v(M,c)$ is non-zero whenever the right-hand side is positive.
\end{lemma}

\begin{proof}
The identity in \eqref{eq:initial-generic-bound} is \eqref{eq:initial-projection-identity} with $u=v(M,c)$, since $\theta_\phi^{\mathsf T}G\bigl(sG^{-1}M_{j_\star}^{\mathsf T}\bigr)=s\,M_{j_\star}\theta_\phi$. For the inequality, if $c=0$ then $s=0$ and the left-hand side is zero. If $c\ne0$, then $|c_{j_\star}|=\norm c_\infty>0$, so that $s\in\{-1,1\}$ and $s\,c_{j_\star}=\norm c_\infty$. Hence,
\[
 s\,M_{j_\star}\theta_\phi-\norm c_\infty
 =-s\bigl(c_{j_\star}-M_{j_\star}\theta_\phi\bigr),
\]
and the absolute value of the right-hand side is $|c_{j_\star}-M_{j_\star}\theta_\phi|\le\norm{c-M\theta_\phi}_\infty$. Finally, if $\norm c_\infty>\norm{c-M\theta_\phi}_\infty$, then $I(a_{v(M,c)})>0=I(a_0)$, so that $v(M,c)\ne0$.
\end{proof}

\subsection{Proof of Theorem ~\ref{thm:initial-single-test}}
\label{sec:proof-initial-single-test}

\begin{proof}
We apply \cref{lem:initial-generic} with $(M,c)=(B,b)$, in which case $v(B,b)$ is the vector $v$ in \eqref{eq:max-violated-test-direction}. By \eqref{eq:linear-moment-violation}, $\norm b_\infty=m_\F(0)$ and $\norm{b-B\theta_\phi}_\infty=m_\F(h_\phi)$. Hence, $I(a_v)\ge m_\F(0)-m_\F(h_\phi)=\Delta_\phi$, and if $\Delta_\phi>0$, then $v\ne0$ and $I(a_v)>0$.
\end{proof}

\subsection{The first segment of the constrained path}
\label{sec:initial-path}

We now formalize the claim of \cref{sec:initial-direction} that the most-violated-test direction is the first non-zero direction that the constrained program \eqref{eq:initial-population-path} selects, as its tolerance decreases below $m_\F(0)=\norm b_\infty$. The relevant geometry is classical in parametric quadratic programming \citep{bemporad2002}. We give a direct proof for the first segment of the path, which also covers ties among the most-violated tests.

\Needspace{6\baselineskip}
\begin{proposition}[The first segment of the constrained path]
\label{prop:initial-path}
Suppose that $G\succ0$ and that $\norm{b-Bu_0}_\infty<\norm b_\infty$ for some $u_0\in\R^p$. Let
\[
 \mathcal J_\star=\{j:|b_j|=\norm b_\infty\},\qquad
 C_j=\sgn(b_j)B_j\quad(j\in\mathcal J_\star),
\]
be the set of the most-violated tests and their signed moment rows, let $C$ be the matrix with rows $C_j$, $j\in\mathcal J_\star$, and define
\begin{equation}\label{eq:initial-ray-program}
 w=\argmin_{u:\,Cu\ge\mathbf1}\tfrac12u^{\mathsf T}Gu,
\end{equation}
where the inequality holds for every coordinate, and, with the minimum of an empty set interpreted as $+\infty$,
\begin{equation}\label{eq:initial-ray-length}
 t_{\rm loc}=\min\left\{
  \min_{j\in\mathcal J_\star}\frac{2\norm b_\infty}{1+C_jw},\quad
  \min_{j\notin\mathcal J_\star}\frac{\norm b_\infty-|b_j|}{1+|B_jw|}
                   \right\}.
\end{equation}
Then the vector $w$ exists, is unique and is non-zero, $0<t_{\rm loc}\le\norm b_\infty$, and the solution of the program \eqref{eq:initial-population-path} satisfies
\begin{equation}\label{eq:initial-ray-identity}
 \theta_{m_\F(0)-s}=s\,w\qquad\text{for every }0<s\le t_{\rm loc}.
\end{equation}
If the most-violated test is unique, i.e., $\mathcal J_\star=\{j_\star\}$, then
\begin{equation}\label{eq:initial-ray-single}
 w=\frac{v}{B_{j_\star}G^{-1}B_{j_\star}^{\mathsf T}},
 \qquad v=\sgn(b_{j_\star})\,G^{-1}B_{j_\star}^{\mathsf T},
\end{equation}
i.e., $w$ is a positive multiple of the vector $v$ of the most-violated-test direction \eqref{eq:max-violated-test-direction}. Finally, if $\Delta_\phi>0$, then the assumption on $u_0$ holds with $u_0=\theta_\phi$, and
\begin{equation}\label{eq:initial-ray-improvement}
 I(a_w)\ge\Delta_\phi\,w^{\mathsf T}Gw>0.
\end{equation}
\end{proposition}

\begin{proof}
Let $\beta=\norm b_\infty$ and $\delta_0=\beta-\norm{b-Bu_0}_\infty>0$. In particular, $\beta>0$ and $\mathcal J_\star$ is non-empty. For $j\in\mathcal J_\star$, we have $\sgn(b_j)b_j=\beta$, and therefore
\begin{equation}\label{eq:initial-ray-feasible}
 C_ju_0=\beta-\sgn(b_j)(b_j-B_ju_0)\ge \beta-\norm{b-Bu_0}_\infty=\delta_0.
\end{equation}
Hence, $u_0/\delta_0$ is feasible in \eqref{eq:initial-ray-program}. The feasible set of this program is closed and convex, and its objective is strictly convex and coercive, because $G\succ0$, so that a minimizer $w$ exists and is unique. It is non-zero, since the zero vector violates the constraint $Cu\ge\mathbf1$. Moreover, every term in \eqref{eq:initial-ray-length} is strictly positive, and since $C_jw\ge1$ for $j\in\mathcal J_\star$, the terms of the first minimum are at most $\beta$. Hence, $0<t_{\rm loc}\le \beta$.

Next, fix $0<s\le t_{\rm loc}$. We first show that $sw$ is feasible in the program \eqref{eq:initial-population-path} with tolerance $\beta-s$. For $j\in\mathcal J_\star$, the signed residual of $sw$ is $\sgn(b_j)(b_j-sB_jw)=\beta-s\,C_jw$. It is at most $\beta-s$, since $C_jw\ge1$, and at least $-(\beta-s)$, since $s(1+C_jw)\le2\beta$ by the definition of $t_{\rm loc}$. For $j\notin\mathcal J_\star$,
\[
 |b_j-sB_jw|\le|b_j|+s|B_jw|\le \beta-s,
\]
again by the definition of $t_{\rm loc}$. Hence, $\norm{b-sBw}_\infty\le \beta-s$. Conversely, every $\theta$ with $\norm{b-B\theta}_\infty\le \beta-s$ satisfies $\beta-C_j\theta=\sgn(b_j)(b_j-B_j\theta)\le \beta-s$ for every $j\in\mathcal J_\star$, i.e., $C\theta\ge s\mathbf1$. By \eqref{eq:initial-ray-program} and the homogeneity of the objective, the unique minimizer of $\theta^{\mathsf T}G\theta$ over the larger set $\{\theta:C\theta\ge s\mathbf1\}$ is $sw$. Since $sw$ is feasible in the original program, it is also its unique minimizer, which proves \eqref{eq:initial-ray-identity}.

Suppose now that $\mathcal J_\star=\{j_\star\}$, and let $q_\star=B_{j_\star}G^{-1}B_{j_\star}^{\mathsf T}$. The row $B_{j_\star}$ is non-zero, since otherwise $|b_{j_\star}-B_{j_\star}u_0|=\beta$ would contradict the assumption on $u_0$, and therefore $q_\star>0$. Let $\bar w=v/q_\star$. Then $C_{j_\star}\bar w=1$, so that $\bar w$ is feasible in \eqref{eq:initial-ray-program}, and $G\bar w=C_{j_\star}^{\mathsf T}/q_\star$. For every feasible $u$, the convexity of the objective gives
\[
 \tfrac12u^{\mathsf T}Gu-\tfrac12\bar w^{\mathsf T}G\bar w
 \ge\bar w^{\mathsf T}G(u-\bar w)=\frac{C_{j_\star}u-1}{q_\star}\ge0,
\]
so that $w=\bar w$ by uniqueness, which is \eqref{eq:initial-ray-single}.

Finally, suppose that $\Delta_\phi>0$, i.e., $\norm{b-B\theta_\phi}_\infty<\norm b_\infty$. Then the assumption of the proposition holds with $u_0=\theta_\phi$ and $\delta_0=\Delta_\phi$, and \eqref{eq:initial-ray-feasible} shows that $\theta_\phi/\Delta_\phi$ is feasible in \eqref{eq:initial-ray-program}. Since $w$ minimizes a convex function over a convex set, its first-order optimality condition gives $w^{\mathsf T}G(\theta_\phi/\Delta_\phi-w)\ge0$. By \eqref{eq:initial-projection-identity}, $I(a_w)=\theta_\phi^{\mathsf T}Gw\ge\Delta_\phi\,w^{\mathsf T}Gw$, which is strictly positive because $w\ne0$ and $G\succ0$.
\end{proof}

Note that, when the most-violated test is unique, $w^{\mathsf T}Gw=1/(B_{j_\star}G^{-1}B_{j_\star}^{\mathsf T})$, so that \eqref{eq:initial-ray-improvement} is exactly the bound $I(a_v)\ge\Delta_\phi$ of \cref{thm:initial-single-test}. With ties, the program \eqref{eq:initial-ray-program} combines the constraints of all the most-violated tests, and its solution is still strictly improving under the positive-gap condition. Moreover, every tolerance in the interval $[m_\F(0)-t_{\rm loc},m_\F(0))$ returns the same direction, which is why no numerical choice of the tolerance is needed.

\subsection{Proof of Proposition ~\ref{prop:initial-growing-sieve}}
\label{sec:proof-initial-growing-sieve}

\begin{proof}
Let $f_m=g-h_{\phi,m}$, so that $\norm{Df_m}_2=\delta_m$. We first show that the Poincar\'e inequality applies to $f_m$. Since $\V_{\phi_m}$ is a linear subspace of $\V$, we have $-h_{\phi,m}\in\V$, and the convexity of $\V$ gives $\tfrac12f_m=\tfrac12g+\tfrac12(-h_{\phi,m})\in\V$. Both sides of \eqref{eq:poincare} scale by the square of a constant factor, so that \eqref{eq:poincare} also holds for $f_m$, i.e., $\Var(f_m(X))\le C_P\,\delta_m^2$.

Next, we bound the tested moments of $h_{\phi,m}$. Fix a test $\psi_j$. Since $Y=g(X)+U$ and $\E[U\mid Z]=0$, the tower property gives $\E[U\psi_j(Z)]=0$, and therefore
\[
 \E[(Y-h_{\phi,m}(X))\psi_j(Z)]=\E[f_m(X)\psi_j(Z)]
 =\E\bigl[(f_m(X)-\E[f_m(X)])\psi_j(Z)\bigr],
\]
where the last step uses that the test is centered. By the Cauchy--Schwarz inequality, the absolute value of the right-hand side is at most $\sqrt{\Var(f_m(X))}\,\norm{\psi_j}_2\le L_\psi\sqrt{C_P}\,\delta_m$. Taking the maximum over $j$ gives \eqref{eq:initial-sieve-violation}.

The first inequality in \eqref{eq:initial-sieve-improvement} follows from \cref{lem:initial-generic}, applied to the dictionary $\phi_m$ with $(M,c)=(B_m,b)$, where $B_m=\E[\psi(Z)\phi_m(X)^{\mathsf T}]$: in this case, $\norm c_\infty=m_\F(0)$ and $\norm{c-M\theta_{\phi_m}}_\infty=m_\F(h_{\phi,m})$. The second inequality is \eqref{eq:initial-sieve-violation}.

Finally, suppose that $m_\F(0)>0$ and $\delta_m\to0$. Then there is an $m_0$ such that $L_\psi\sqrt{C_P}\,\delta_m<m_\F(0)$ for every $m\ge m_0$. For such $m$, \eqref{eq:initial-sieve-violation} gives $m_\F(h_{\phi,m})<m_\F(0)$, which is the positive-gap condition, and \eqref{eq:initial-sieve-improvement} gives $I(a_m)\ge m_\F(0)-m_\F(h_{\phi,m})>0$. Moreover, $|I(a_m)-m_\F(0)|\le L_\psi\sqrt{C_P}\,\delta_m\to0$, so that $I(a_m)\to m_\F(0)$.
\end{proof}

\subsection{Proof of Proposition ~\ref{prop:initial-reduced-form}}
\label{sec:proof-initial-reduced-form}

\begin{proof}
We first identify the maximizing test. Applying \eqref{eq:initial-outcome-moment} to $\psi_j$ and to $-\psi_j$, and using $\norm{\psi_j}_2\le1$, every test satisfies $|\E[Y\psi_j(Z)]|\le R$. The test $\psi_\star=r_0/R$ is centered because $\E[r_0(Z)]=0$, has unit norm, and has moment $\E[Y\psi_\star(Z)]=\E[r_0(Z)^2]/R=R$. It therefore attains the bound, so that it is a most-violated test under the zero response, its moment has a positive sign, and $m_\F(0)=R$.

Next, let $j_\star$ be the index of $\psi_\star$ in the list of tests, so that $b_{j_\star}=R>0$ and $B_{j_\star}=\E[\psi_\star(Z)\phi(X)^{\mathsf T}]$. The vector in \eqref{eq:max-violated-test-direction} is then $v_r=G^{-1}B_{j_\star}^{\mathsf T}=R^{-1}G^{-1}\E[\phi(X)r_0(Z)]$, and \cref{lem:initial-generic} with $(M,c)=(B,b)$ gives
\[
 I(a_r)=B_{j_\star}\theta_\phi=\frac{\E[h_\phi(X)r_0(Z)]}{R},
 \qquad
 |I(a_r)-R|\le\norm{b-B\theta_\phi}_\infty=m_\F(h_\phi).
\]
The second display is \eqref{eq:initial-reduced-form-bound}, and since $m_\F(0)=R$, the gap is $\Delta_\phi=R-m_\F(h_\phi)$. For the first one, the tower property gives $\E[h_\phi(X)r_0(Z)]=\langle Th_\phi,r_0\rangle_{L^2(P_Z)}$, and since $\E[r_0(Z)]=0$, subtracting the constant $\E[h_\phi(X)]$ from $Th_\phi$ does not change this inner product. This proves the formula for $I(a_r)$ in \eqref{eq:initial-reduced-form-direction}.

We now prove \eqref{eq:initial-reduced-form-metrics}. Let $f=g-h_\phi$ and fix a test $\psi_j$. Since $\E[U\mid Z]=0$, the tower property gives $\E[U\psi_j(Z)]=0$, so that $\E[(Y-h_\phi(X))\psi_j(Z)]=\E[f(X)\psi_j(Z)]=\langle Tf,\psi_j\rangle_{L^2(P_Z)}$. Since $\psi_j$ is centered, this equals $\langle T_0f,\psi_j\rangle_{L^2(P_Z)}$, and the Cauchy--Schwarz inequality with $\norm{\psi_j}_2\le1$ gives
\[
 |\E[(Y-h_\phi(X))\psi_j(Z)]|\le\norm{T_0f}_2 .
\]
Taking the maximum over $j$ gives $m_\F(h_\phi)\le\norm{T_0f}_2$. For the second inequality, let $\bar f=f-\E[f(X)]$. Since $T$ maps a constant to itself, $T_0f=T\bar f$. By the conditional Jensen inequality, $(T\bar f)(Z)^2\le\E[\bar f(X)^2\mid Z]$ almost surely, and taking expectations gives $\norm{T_0f}_2\le\lVert\bar f\rVert_2$.

Finally, suppose that $\V_\phi\subseteq\V$ and that \eqref{eq:poincare} holds. Then $-h_\phi\in\V$, and the convexity of $\V$ gives $\tfrac12f=\tfrac12g+\tfrac12(-h_\phi)\in\V$. Both sides of \eqref{eq:poincare} scale by the square of a constant factor, so that \eqref{eq:poincare} also holds for $f$, i.e., $\lVert\bar f\rVert_2^2=\Var(f(X))\le C_P\norm{Df}_2^2$. Hence, the right-hand side of \eqref{eq:initial-reduced-form-metrics} is at most $\sqrt{C_P}\norm{D(g-h_\phi)}_2$. If the latter is smaller than $R$, then $m_\F(h_\phi)<R$ and \eqref{eq:initial-reduced-form-bound} gives $I(a_r)\ge R-m_\F(h_\phi)>0$.
\end{proof}

\subsection{Proof of Theorem ~\ref{thm:initial-statistical}}
\label{sec:proof-initial-statistical}

\begin{proof}
We work on the event on which the three bounds in \eqref{eq:initial-deviations} hold. Let $H=G^{-1/2}\widehat GG^{-1/2}$. The third bound in \eqref{eq:initial-deviations} is equivalent to $(1-\zeta_G)I_p\preceq H\preceq(1+\zeta_G)I_p$, i.e., to $\norm{H-I_p}_{\rm op}\le\zeta_G$. Since $\zeta_G<1$, we have $\lambda_{\min}(H)\ge1-\zeta_G>0$, so that $\widehat G=G^{1/2}HG^{1/2}\succ0$ and $\norm{H^{-1}}_{\rm op}\le1/(1-\zeta_G)$. In particular, the rule returns the zero vector only if $\widehat b=0$.

Suppose first that $\widehat b\ne0$, and let $\widehat s=\sgn(\widehat b_{\widehat j})$, so that $\widehat G\widehat v=\widehat s\,\widehat B_{\widehat j}^{\mathsf T}$. Let
\[
 y=G^{1/2}\theta_\phi,\qquad z=G^{1/2}\widehat v,\qquad
 w=G^{-1/2}\widehat B_{\widehat j}^{\mathsf T},
\]
so that $Hz=\widehat s\,w$ and $\norm y^2=\theta_\phi^{\mathsf T}G\theta_\phi=\norm{Dh_\phi}_2^2$. By \eqref{eq:initial-projection-identity} with $u=\widehat v$,
\[
 I(a_{\widehat v})=\theta_\phi^{\mathsf T}G\widehat v=y^{\mathsf T}z
 =y^{\mathsf T}Hz+y^{\mathsf T}(I_p-H)z
 =\widehat s\,\widehat B_{\widehat j}\theta_\phi+y^{\mathsf T}(I_p-H)z .
\]
For the first term, the inequality in \cref{lem:initial-generic} with $(M,c)=(\widehat B,\widehat b)$, the triangle inequality, and the first two bounds in \eqref{eq:initial-deviations} give
\[
 \widehat s\,\widehat B_{\widehat j}\theta_\phi
 \ge\norm{\widehat b}_\infty-\norm{\widehat b-\widehat B\theta_\phi}_\infty
 \ge\bigl(\norm b_\infty-\zeta_b\bigr)-\bigl(\norm{b-B\theta_\phi}_\infty+\zeta_\phi\bigr)
 =\Delta_\phi-\zeta_b-\zeta_\phi .
\]
For the second term, $\norm z\le\norm{H^{-1}}_{\rm op}\norm w\le\norm w/(1-\zeta_G)$ and $\norm w\le\norm{\widehat B_{\widehat j}}/\sqrt\gamma\le L_{B,n}/\sqrt\gamma$, so that the Cauchy--Schwarz inequality gives
\[
 |y^{\mathsf T}(I_p-H)z|
 \le\norm y\,\norm{H-I_p}_{\rm op}\,\norm z
 \le\frac{\zeta_G}{1-\zeta_G}\cdot\frac{\norm{Dh_\phi}_2\,L_{B,n}}{\sqrt\gamma}.
\]
Combining the last three displays proves \eqref{eq:initial-statistical-bound}.

We now treat the case $\widehat b=0$. Then $\widehat v=0$ and $I(a_{\widehat v})=0$. The first bound in \eqref{eq:initial-deviations} gives $m_\F(0)=\norm b_\infty\le\zeta_b$, so that $\Delta_\phi-\zeta_b-\zeta_\phi\le m_\F(0)-\zeta_b\le0$. The right-hand side of \eqref{eq:initial-statistical-bound} is therefore non-positive, and the bound holds.
\end{proof}

\subsection{Deviation bounds for a linear dictionary}
\label{sec:initial-deviation-lemmas}

We now bound the three deviations in \eqref{eq:initial-deviations} for a linear dictionary. The two moment deviations are bounded by the global Rademacher complexity of the test class, as in \cref{sec:sparse-linear}. For the Gram matrix, a bound based on the global Rademacher complexity of the gradient class would be of the order $\sqrt{p/n}$. The next lemma instead uses the matrix Chernoff inequality of \citet{tropp2012}, which controls the error of the empirical Gram matrix relative to $G$, with a logarithmic dependence on the dimension of the dictionary.

\Needspace{6\baselineskip}
\begin{lemma}[Matrix Chernoff bound for the empirical Gram matrix]
\label{lem:initial-gram-chernoff}
Suppose that $G\succ0$, with $\gamma=\lambda_{\min}(G)$, and that $\norm{J_\phi(x)}_{\rm op}\le L_\phi$ for every $x$. Then the empirical Gram matrix $G_n=\Pn[J_\phi(X)^{\mathsf T}J_\phi(X)]$ satisfies, for every $0<t<1$,
\begin{equation}\label{eq:initial-gram-chernoff}
 \Pp\bigl\{(1-t)\,G\preceq G_n\preceq(1+t)\,G\bigr\}
 \ \ge\ 1-2p\exp\Bigl(-\frac{\gamma\,n\,t^2}{3L_\phi^2}\Bigr).
\end{equation}
\end{lemma}

\begin{proof}
Let $\Lambda=L_\phi^2/\gamma$ and, for $1\le i\le n$, let
\[
 W_i=G^{-1/2}J_\phi(X_i)^{\mathsf T}J_\phi(X_i)G^{-1/2}.
\]
These are independent positive semi-definite $p\times p$ matrices, with $\E[W_i]=G^{-1/2}GG^{-1/2}=I_p$ and
\[
 \norm{W_i}_{\rm op}=\norm{J_\phi(X_i)G^{-1/2}}_{\rm op}^2
 \le L_\phi^2\,\norm{G^{-1/2}}_{\rm op}^2=\Lambda .
\]
Let $\bar W=\frac1n\sum_{i=1}^nW_i=G^{-1/2}G_nG^{-1/2}$. The matrix Chernoff inequalities \citep[Theorem~1.1]{tropp2012}, applied to the sum $\sum_iW_i$, whose expectation is $n\,I_p$, give for every $0<t<1$,
\[
 \Pp\{\lambda_{\max}(\bar W)\ge1+t\}
 \le p\Bigl[\frac{e^{t}}{(1+t)^{1+t}}\Bigr]^{n/\Lambda},
 \qquad
 \Pp\{\lambda_{\min}(\bar W)\le1-t\}
 \le p\Bigl[\frac{e^{-t}}{(1-t)^{1-t}}\Bigr]^{n/\Lambda}.
\]
Since $e^{t}/(1+t)^{1+t}\le e^{-t^2/3}$ and $e^{-t}/(1-t)^{1-t}\le e^{-t^2/2}$ for $0<t<1$, each of the two probabilities is at most $p\exp(-nt^2/(3\Lambda))$. On the complement of the two events, $(1-t)I_p\preceq\bar W\preceq(1+t)I_p$, and multiplying by $G^{1/2}$ from both sides gives $(1-t)G\preceq G_n\preceq(1+t)G$.
\end{proof}

Note that the proof uses the bound on the Jacobian only through $\norm{J_\phi(x)G^{-1/2}}_{\rm op}^2\le L_\phi^2/\gamma$. We can now state the deviation bounds that we used in \cref{sec:initial-direction}.

\Needspace{6\baselineskip}
\begin{lemma}[Deviation bounds for a linear dictionary]
\label{lem:initial-linear-deviations}
Suppose that \cref{ass:linear-dictionary}, case (a), holds with the test class \eqref{eq:linear-test-class}, that $G\succ0$ with $\gamma=\lambda_{\min}(G)$, that $\norm{\theta_\phi}\le R_2$, and that $\widehat G=G_n$. There is a universal constant $C$ such that, if $n>3(L_\phi^2/\gamma)\log(4p/\eta)$, then with probability at least $1-\eta$, \eqref{eq:initial-deviations} holds with
\[
 \zeta_b=\zeta_\phi=C\,r_\F,\qquad
 \zeta_G=L_\phi\sqrt{\frac{3\log(4p/\eta)}{\gamma\,n}}<1,
\]
where $r_\F\le\Delta_{n,k}$ is the global complexity of the test class in \eqref{eq:linear-radii}. Moreover, $L_{B,n}\le L_\phi$
almost surely, and $\norm{Dh_\phi}_2\le1$.
\end{lemma}

\begin{proof}
Under \cref{ass:linear-dictionary}, case (a), \cref{ass:statistical} holds for
$\Hc=\{h_\theta:\norm\theta\le R_2\}$ and $\F=\{c\,\psi_\ell:1\le\ell\le k,\ c\in[-1,1]\}$. The zero
function lies in $\Hc$. Since $\norm{\theta_\phi}\le R_2$, so does $h_\phi$. As $\Hc\subseteq\V$, each of
them is a comparator in the sense of \cref{def:graph-approximation}, with $\epsilon=\norm{g}_\V$ and
$\epsilon=\norm{h_\phi-g}_\V$, respectively. The value of $\epsilon$ plays no role below. We apply
\cref{lem:concentration} at confidence level $\eta/4$, once with $h^\circ=0$ and once with $h^\circ=h_\phi$.
With probability at least $1-\eta/2$, \eqref{eq:noise-conc} then holds for both comparators, with the global complexity of the test class computed at level $\eta/4$.

Since $\F$ consists of the multiples $c\psi_\ell$ with $|c|\le1$, for every integrable $W$,
$\sup_{f\in\F}|(\Pn-\E)[Wf(Z)]|=\max_\ell|(\Pn-\E)[W\psi_\ell(Z)]|$. Taking $W=Y$ and $W=Y-h_\phi(X)$,
\eqref{eq:noise-conc} with $h^\circ=0$ and $h^\circ=h_\phi$ gives
\[
 \norm{\widehat b-b}_\infty\le Cr_\F,\qquad
 \norm{(\widehat b-\widehat B\theta_\phi)-(b-B\theta_\phi)}_\infty\le Cr_\F .
\]
Changing $\eta$ to $\eta/4$ increases $\ell_n$ by $\log4\le\ell_n$, so that the global complexity changes by at most a factor $\sqrt2$, which we absorb into $C$.

For the Gram matrix, we apply \cref{lem:initial-gram-chernoff} with $t=\zeta_G$, which is smaller than one by the condition on $n$. The probability that the third bound in \eqref{eq:initial-deviations} fails is at most $2p\exp(-\gamma n\zeta_G^2/(3L_\phi^2))=2p\exp(-\log(4p/\eta))=\eta/2$, and the union bound shows that the three bounds hold jointly with probability at least $1-\eta$.

Finally, $\norm{\widehat B_\ell}\le\Pn[|\psi_\ell|\norm\phi]\le L_\phi$ for every $\ell$, so that
$L_{B,n}\le L_\phi$, and $\norm{Dh_\phi(x)}=\norm{J_\phi(x)\theta_\phi}\le L_\phi R_2\le1$ for every $x$, so that $\norm{Dh_\phi}_2\le1$.

If $\zeta_G\le1/2$, then $\zeta_G/(1-\zeta_G)\le2\zeta_G$, and substituting these bounds in \eqref{eq:initial-statistical-bound} gives
\[
 I(a_{\widehat v})\ge\Delta_\phi-2Cr_\F-2\zeta_G\frac{L_\phi}{\sqrt\gamma}
 =\Delta_\phi-2Cr_\F-2\sqrt3\,\frac{L_\phi^2}{\gamma}\sqrt{\frac{\log(4p/\eta)}{n}},
\]
which implies \eqref{eq:initial-linear-rates} since $r_\F\le\Delta_{n,k}$ and $2\sqrt3\le4$.

\end{proof}

\bookmarksetup{startatroot}
\phantomsection
\addcontentsline{toc}{section}{References}
\bibliographystyle{plainnat}
\bibliography{references}

\end{document}